\documentclass[jair,twoside,11pt,theapa]{article}
\usepackage{jair,  theapa, rawfonts}
\usepackage[utf8]{inputenc} 
\usepackage{hyperref}       
\usepackage{url}            
\usepackage{booktabs}       
\usepackage{amsfonts}       
\usepackage{nicefrac}       
\usepackage{microtype}      
\usepackage{amsmath}
\usepackage[capitalise,nameinlink,noabbrev]{cleveref}
\usepackage{color}
\usepackage{xcolor}
\usepackage{array}
\usepackage{amssymb}
\usepackage{graphicx}
\usepackage{makecell}
\usepackage{multirow}
\usepackage{lscape}
\usepackage{tikz}
\usetikzlibrary{mindmap, shadows, arrows.meta, positioning, shapes.geometric}
\usepackage{smartdiagram}
\usepackage{wrapfig}
\usepackage{subcaption}
\usepackage{tcolorbox}
\usepackage{xcolor}
\usepackage{soul}
\usepackage{stmaryrd}
\usepackage{bbm}
\usepackage{enumitem}

\newtheorem{definition}{Definition}

\usepackage{algorithm}
\usepackage{algpseudocode} 

\usepackage{todonotes}
\usepackage[round]{natbib}
\usepackage[mathscr]{eucal}
\newcommand{\deriv}[2][]{\frac{\partial#1}{\partial#2}}
\def\given{\middle\vert}

\DeclareMathOperator*{\argmin}{arg\,min}

\newcommand{\checkit}{$\bullet$}
\usepackage{fancyhdr}

\usepackage{bm}
\usepackage{enumerate}
\usepackage{tabularx}
\usepackage[framemethod=TikZ]{mdframed}
\newenvironment{enumerate*}%
  {\begin{enumerate}%
    \vskip 0.1in%
    \setlength{\itemsep}{0pt}%
    \setlength{\parskip}{0pt}}%
  {\end{enumerate}%
   \vskip -0.1in}

\makeatletter

\patchcmd{\NAT@citex}
  {\@citea\NAT@hyper@{%
     \NAT@nmfmt{\NAT@nm}%
     \hyper@natlinkbreak{\NAT@aysep\NAT@spacechar}{\@citeb\@extra@b@citeb}%
     \NAT@date}}
  {\@citea\NAT@nmfmt{\NAT@nm}%
   \NAT@aysep\NAT@spacechar\NAT@hyper@{\NAT@date}}{}{}

\patchcmd{\NAT@citex}
  {\@citea\NAT@hyper@{%
     \NAT@nmfmt{\NAT@nm}%
     \hyper@natlinkbreak{\NAT@spacechar\NAT@@open\if*#1*\else#1\NAT@spacechar\fi}%
       {\@citeb\@extra@b@citeb}%
     \NAT@date}}
  {\@citea\NAT@nmfmt{\NAT@nm}%
   \NAT@spacechar\NAT@@open\if*#1*\else#1\NAT@spacechar\fi\NAT@hyper@{\NAT@date}}
  {}{}

\jairheading{1}{1993}{1-15}{6/91}{9/91}
\ShortHeadings{Discovering Temporal Structure: An Overview of Hierarchical RL}
{}
\firstpageno{1}

\begin{document}

\title{Discovering Temporal Structure:\\ An Overview of Hierarchical Reinforcement Learning}

\author{\name Martin Klissarov\thanks{Equal contribution.} \email martin.klissarov@mail.mcgill.ca \\
       \addr Mila, McGill University
       \AND
       \name Akhil Bagaria\footnotemark[1] \thanks{Work done while at Brown University.}
        \email akhilbg@amazon.com \\
       \addr Amazon
       \AND
       \name Ziyan “Ray” Luo\footnotemark[1] \email ziyan.luo@mail.mcgill.ca\\
       \addr Mila, McGill University
       \AND
       \name  George Konidaris\footnotemark[3] \email gdk@cs.brown.edu\\
       \addr Brown University
       \AND
       \name  Doina Precup\thanks{Equal supervision.} \email dprecup@cs.mcgill.ca \\
       \addr Mila, McGill University\\
       \addr Canada CIFAR AI Chair
       \AND
       \name  Marlos C. Machado\footnotemark[3]
       \email machado@ualberta.ca\\
       \addr Amii, University of Alberta\\
       \addr Canada CIFAR AI Chair
   }

\maketitle

\begin{abstract}
Developing agents capable of exploring, planning, and learning in complex environments is a grand challenge in artificial intelligence (AI). 
Hierarchical reinforcement learning (HRL) offers a promising solution to this challenge by discovering and exploiting temporal structure within an environment, via constructs such as options, skills, or goal-conditioned policies. 
The strong appeal of the HRL framework has led to a rich and diverse body of literature, but it is still not clear what constitutes good structure, or the kind of problems in which identifying it may be helpful. 
We identify the broad classes of benefits---and resulting tradeoffs---that HRL may bring to an agent, and use them to introduce and organize methods for discovering temporal structure, in three settings:  learning directly from online experience, learning using offline datasets, and leveraging foundation models. 
Finally, we highlight the remaining challenges of temporal structure discovery and discuss  domains in  which HRL is  particularly well-suited. 
\end{abstract}

\newpage
\tableofcontents
\newpage

\section{Introduction}
\label{Introduction}

Reinforcement learning (RL) is a general computational framework for designing agents that learn to maximize a scalar reward from  experience.
RL agents sense their environment and produce actions at every single timestep, but effective reward maximization in complex environments requires reasoning and learning over many timescales.
Consider how we typically go about our day: we actuate muscles every few milliseconds, while simultaneously making high-level decisions such as choosing presents for a loved one, deciding what to eat for lunch, and planning the next step in our career.
Such abstract decision-making allows us to decide how to act in a complex world  without being overwhelmed by unnecessary detail.

Hierarchical reinforcement learning (HRL) formalizes the idea of flexibly reasoning over different timescales by developing agents that learn, predict, and act in the world at multiple levels of abstraction.
At its core, HRL builds on the temporal structure discovered through interaction with an environment. 
When that temporal structure is defined by a human expert, HRL can dramatically ease the decision-making burden of the agent by improving exploration \citep{Bellemare2020AutonomousNO},  learning \citep{Vinyals2019GrandmasterLI}, and generalization \citep{saycan2022arxiv}. On the other hand, when it is misaligned with the environment, it can hamper learning---for example, resulting in pathologically bad exploration \citep{AAMAS08-jong}. These appeals and drawbacks naturally lead to the question: how can agents autonomously discover useful temporal structure in HRL?

Before designing algorithms that successfully address the discovery problem, 
we are faced with the question of what constitutes a ``good'' temporal structure in the first place.
Is there one type of ``good'' structure that yields higher rewards in all possible environments? 
Are there specific types of problem settings, like multi-task learning \citep{Plappert2018MultiGoalRL} and continual learning \citep{Khetarpal2020TowardsCR}, where we expect HRL to outperform non-hierarchical RL, and others where we do not? 
How can prior knowledge, obtained for instance via the integration of large language models (LLMs), alleviate the difficulty of discovery?
This work presents various perspectives on what constitutes ``good'' temporal structures through the lens of the fundamental problems of RL---specifically,  how HRL can aid \textbf{exploration}, \textbf{credit assignment}, \textbf{transfer}, and \textbf{interpretability}.

\paragraph{Key Contributions.} 
The discovery of useful temporal structures has been a prolific, albeit challenging, topic of research. 
Before we present the various algorithms that tackle this problem, we take a step back and discuss the potential \textbf{benefits} of HRL methods as well as their \textbf{trade-offs} in the context of sequential decision-making. 
It is through the lens of these benefits and trade-offs that we introduce the diverse approaches that have been developed to tackle the fundamental question of discovery.
While recent surveys in HRL \citep{hrl_2021,hrl_2022} present papers based on their technical differences and domains of application, we present the literature based on how each method contributes to these core benefits. 
We then discuss the \textbf{challenges} associated with discovering structure in HRL and the \textbf{domains} that are particularly well-suited for such methods.\looseness=-1

\begin{figure}[!!htbp]
    \centering
\includegraphics[width=.95\linewidth]{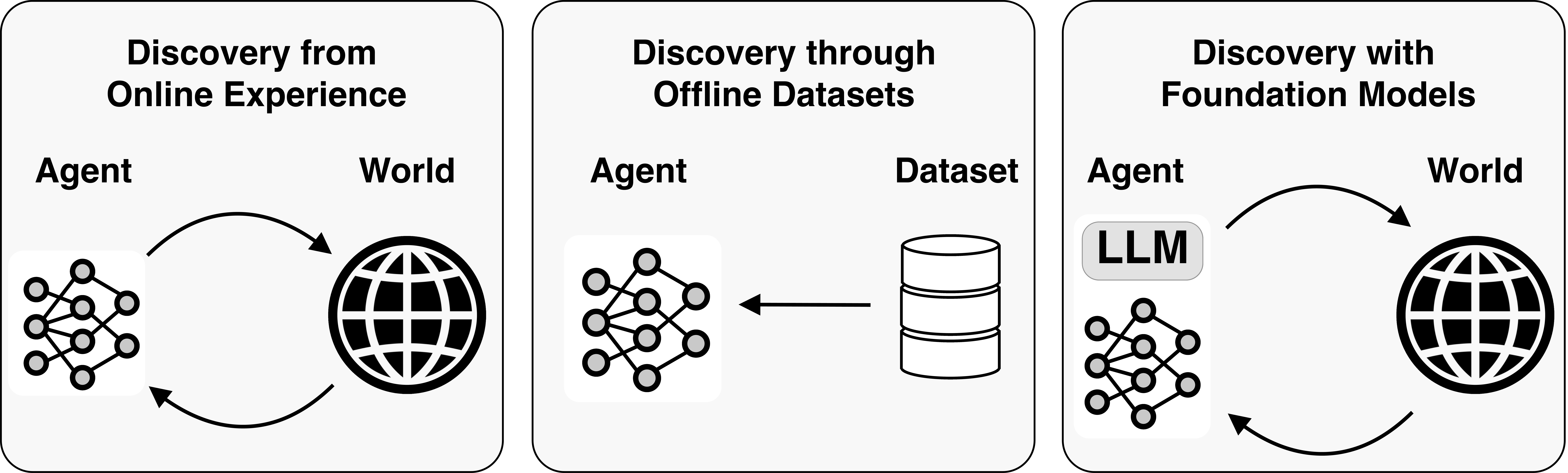}
    \caption{\textbf{Types of methods for temporal structure discovery.} We focus on discovering temporal structure autonomously from data under different assumptions. }
    \label{fig:main}
\end{figure}

\paragraph{Scope.} 
We study temporal structure discovery as a problem faced by the agent. As shown in Figure~\ref{fig:main}, we categorize approaches in terms of the source of data and prior knowledge, presenting works that (1) discover temporal structure from the agent's \textbf{online experience} (Section~\ref{tabularasa}), (2) leverage \textbf{offline datasets}  (Section~\ref{offline}), and (3) build on \textbf{foundation models} such as LLMs (Section~\ref{sec:largemodels}). Most of these algorithms can be instantiated with deep neural networks. For each method within these classes, we discuss the major benefits and limitations that motivate future research. 
We also study how to leverage temporal structure in decision-making (Section~\ref{sec:5}), whether discovered or pre-existing.
\looseness=-1


\section{What is Hierarchical Reinforcement Learning For?}
\label{sec:2}

HRL aims to exploit the temporal structure of sequential decision-making problems. 
Solutions to complex tasks can often be deconstructed into simpler subtasks that are modular and composable. 
\textbf{Modularity} refers to the property that a solution to a subtask can be reused without concern for exactly how it was solved. 
\textbf{Compositionality} means that solutions to subtasks can subsequently be recombined to solve a wide range of more complex problems.

\begin{figure}[!!!ht]
    \centering
    \includegraphics[width=\linewidth]{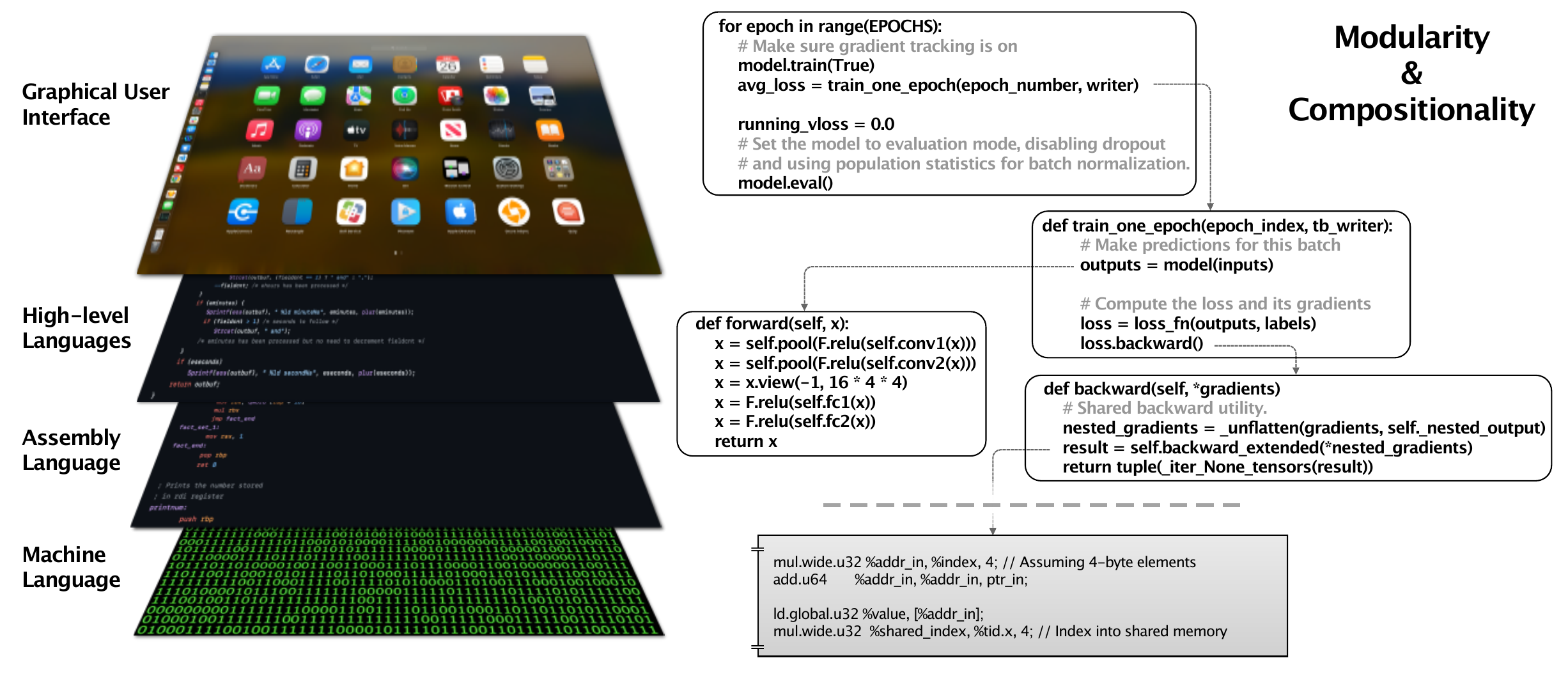}
    \caption{\textbf{(Left)} The hierarchy of software and languages which make computers especially useful, allowing humans to directly interact with a user interface or a high-level language to achieve a diversity of goals. \textbf{(Right)} a code snippet, revealing the usefulness of modular and compositional software structures that researchers use every day. The PyTorch language, which abstracts over the PTX language used within CUDA kernels, allows exploring research ideas efficiently.}
    \label{fig:abstraction}
\end{figure}

To better understand what such a structure might represent in practice, consider a programmer with an abundance of time who cares about solving only a single task. 
In such a scenario, assembly language might be the optimal choice because its precise control over hardware resources maximizes memory efficiency and minimizes execution time. 
However, in practice, programmers often opt for higher-level programming languages and  software libraries because they offer  modules that can be composed to solve common programming subtasks, making  writing   new programs  efficient at the cost of increased execution time.  
In fact, most large software projects would simply be infeasible without such programming languages. This idea is visually represented in Figure~\ref{fig:abstraction}, where abstract interfaces allow us to manipulate machine language efficiently.
Modularity and compositionality are also particularly appealing properties for software expected to undergo changes throughout its life cycle, because modules can be updated and debugged locally, without considering the global complexity of the task they are used to solve.
Developing such a library requires careful consideration of the right code organization to adopt and which guiding principles to follow, balancing execution time, readability, and performance. 
However, once a library is written, the user can focus on the overall program's behaviour without needing to understand the implementation details of each function \citep{wilkes1958preparation}. This is also represented in Figure~\ref{fig:abstraction}, where the modular and compositional nature of PyTorch allows researchers to efficiently explore research ideas.

The temporal structure at the core of HRL is analogous to functions and subroutines in programming languages.\footnote{We provide more details about this analogy in Section \ref{sec:prog}.} Just as a human programmer must write a complex program by breaking it into subtasks, so must RL agents autonomously identify hierarchical structure in a stream of data. Modularity and compositionality properties are therefore good indicators as to the kind of problems in which we might find HRL particularly useful (see Section \ref{sec:8}).
%
%
We now discuss how, by discovering and leveraging such temporal structure, HRL methods can help address fundamental challenges in decision-making.

\subsection{The Benefits of Hierarchical Reinforcement Learning}
\label{sec:benefits} 
Just as modular and compositional codebases can facilitate effective software development, HRL can leverage an environment's structure to improve decision-making. 
This is particularly powerful when an agent is faced with complex, long-horizon tasks that can be broken into  manageable subtasks.
To understand this more precisely, we describe the benefits of HRL through the lens of three fundamental challenges agents face when learning from interaction: how to select the right data to collect (\textit{exploration}), how to efficiently learn from this data (\textit{credit assignment}), and how to transfer knowledge to new situations (\textit{transferability}). Additionally, as agents become increasingly more capable, a new challenge emerges: understanding their decision-making processes (\textit{interpretability}). 
When covering the different families of HRL methods (Sections \ref{tabularasa}, \ref{offline}, and \ref{sec:largemodels}), we will explicitly consider how these benefits are instantiated in practice.\looseness=-1

\textbf{Exploration.} Broadly speaking, RL agents must solve two problems: (a) how to use existing data to learn useful behaviours, and (b) what data to collect in the first place. The latter, known as the \textit{exploration} problem, is both unique and central to RL; the agent must learn how to collect data that improves its understanding of the world even if doing so does not immediately maximize reward in the short term \citep{Amin2021ASO}. 
By exploiting the temporal structure of an environment, an agent can improve exploration in at least three ways. 
First, it can \textit{target states that facilitate progress}, for example by defining subgoals that are more immediately attainable than the overall task and can be organized into a progression resembling a curriculum.
Second, it can \textit{explore over extended time scales} by committing to temporally coherent behaviours, rather than selecting individual actions at each step. 
Finally, agents can \textit{explore at a higher level of abstraction} than individual actions, enabling them to search the solution space more efficiently.
Consider a researcher tackling an important scientific question. 
By learning a high-level programming language, such as PyTorch, and writing modular code, the researcher can iterate faster to investigate many high-level ideas. 
When iterating over ideas, the researcher may seek to achieve some important milestones, 
such as constructing a minimal prototype,
that can reveal new perspectives and provide insights into possible future courses of action.\looseness-1

\textbf{Credit Assignment.}  To improve its decisions over time, an agent must identify the key moments in a sequence of decisions that best explain the observed result. 
RL algorithms typically leverage multi-step error propagation \citep{Sutton1988LearningTP} to learn about temporally distant, or \textit{delayed}, outcomes. 
An agent leveraging the environment's temporal structure could more efficiently identify the origins of an outcome by propagating errors at the abstraction level defined by this structure.
Consider our previous example of a researcher performing a scientific experiment. 
Completing such an experiment consists of a sequence of high-level decisions, such as the choices of data preprocessing or evaluation metrics. 
Each high-level decision is instantiated through a series of keystrokes that make up the final working code. 
By reflecting on the validity of the sequence of high-level decisions, rather than of each individual keystroke, the researcher can better identify which ones were critical for the observed outcome.
By breaking down a task into such segments, it is also easier to identify if a particular segment is completed, narrowing down the search for lower-level mistakes, such as  an errant keystroke that introduced a bug.


\textbf{Transfer.} 
HRL offers a particularly promising way of exploiting structure shared across different situations:
temporally extended skills acquired in one context can be reused and adapted in others~\citep{Thrun1994FindingSI}.
Consider our previous example of an AI researcher conducting experiments and writing a paper for a particular conference. 
The reusable code and writing skills developed while working on one paper could substantially reduce the effort required to produce a follow-up article, improving the agent's ability to conduct high-quality research.
Such transfer can extend beyond behaviours to include representations, models, and other forms of knowledge.
More broadly, transferring skills between distinct tasks is one instance of the broader goal of \textit{generalization}: 
the ability to apply learned knowledge in situations beyond those encountered during learning.
For example, with suitable representations, options can remain useful under similar but previously unseen conditions~\citep{ravindran2003relativized,Konidaris2007BuildingPO}.
When an agent must continually adapt to a non-stationary environment, explicit task boundaries may be absent, yet temporal abstractions can still support generalization across changing contexts (see Section~\ref{sec:continual-rl}).\looseness=-1

By addressing these three fundamental challenges, \textbf{HRL can achieve faster learning}. Beyond them, HRL offers two further benefits of a different character: \textit{planning}, which is mutually reinforcing with the three rather than independent of them, and \textit{interpretability}, which concerns whether humans can understand what or how the agent has learned.

\textbf{Planning.} The three benefits above concern how an agent learns from experience. But experience is costly---real interaction takes time and can carry risk---and the environment may change over time. Both problems can be mitigated by substituting computation for interaction. This is \textit{planning}: producing or improving decisions by computing with a model of the environment~\citep[Chapter 8]{Sutton1998}; for example, by learning from simulated experience or by dynamic programming. 
Temporal abstraction makes this substitution effective over long horizons: models of temporally extended behaviours cover the same horizon in fewer applications, thereby reducing the accumulation of model errors. 
Planning is not a fourth independent benefit but is synergetic with the other three: exploration, credit assignment, and transfer allow an agent to construct plannable abstract models, and planning with those models improves each of them in turn---focusing exploration on the most useful parts of the state space, propagating credit in temporally-extended jumps, and reusing options and their models across similar tasks.

\textbf{Interpretability.} 
While not all HRL algorithms produce interpretable behaviour, those that do can help human observers  better understand an agent's decision-making process.
As agents become increasingly powerful they will  be deployed in real-world situations where the consequences of their actions carry considerable stakes, and  interpreting their decisions will be a key aspect of ensuring their alignment \citep{Amodei2016ConcretePI}.
HRL could provide an interpretable interface for AI alignment via a more abstract decision formulation than the one defined in the environment. 
For instance, in a robot navigation task, low-level actions such as the forces applied at the joints are particularly hard to interpret compared to a sequence of semantically meaningful skills such as ``reach the stairs" and ``descend to the first floor".
Humans may follow the agent's reasoning at that level of abstraction and provide corrective feedback.


\subsection{Trade-offs}
\label{sec:tradeoffs}

\begin{figure}[ht!!!]
    \centering
    \includegraphics[width=.9\linewidth]{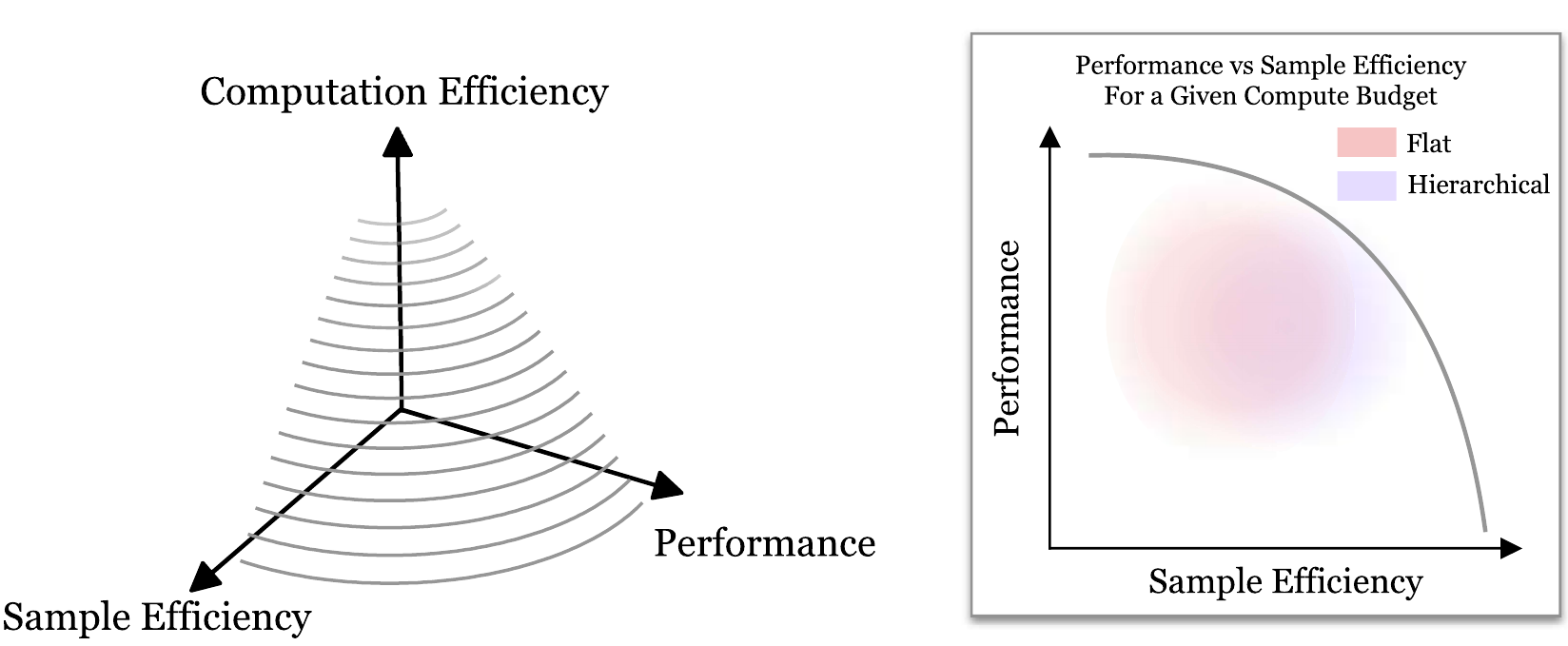}
    \caption{\textbf{(Left)} Agents trade-off  three different objectives: performance (in terms of reward), sample efficiency (the amount of data required), and computational efficiency (amount of computation required). The Pareto frontier between performance and sample efficiency shifts with different compute budgets.  For instance, with unlimited compute (a low computational efficiency), high performance can be achieved at moderate sample efficiency. \textbf{(Right)} A qualitative illustration of a ``flat'' agent and a hierarchical agent under the Pareto frontier of the performance vs. sample efficiency trade-off, given a fixed compute budget. A ``flat'' agent, as opposed to a hierarchical one, does not use temporally-extended actions. HRL agents typically trade some amount of performance for improved sample efficiency.}
    \label{fig:hrl_tradeoff}
\end{figure}

HRL builds on the inductive bias that tasks can be naturally decomposed into simpler subtasks, making it especially effective when such a hierarchical structure is apparent. 
This assumption does not imply that a suitable decomposition is known in advance: the structure may be predefined using domain knowledge, or it may need to be discovered from experience.
When a suitable hierarchy is provided, HRL can exploit it to improve learning or planning efficiency \citep{fruit2017exploration,wen2020efficiency}. However, consistent with the intuition behind the No Free Lunch Theorem~\citep{wolpert1997no}, the benefits of this inductive bias depend on its alignment with the problem. A poorly matched decomposition can slow learning or lead to suboptimal policies \citep{AAMAS08-jong}. Taking the analogy of programming, adding abstractions in a codebase can simultaneously help in seeing a broader picture, but also obscure the important details \citep{victor_2011_ladder}.
When a suitable hierarchy is not provided, the temporal structure itself must be \textit{discovered} by the HRL agent. This discovery incurs a cost in computation and potentially additional environment interactions. The central question is therefore whether, and over what horizon, the benefits of exploiting the learned hierarchy amortize the cost of discovering it. HRL must therefore be assessed in terms of both the suitability of its hierarchy and the amortization of discovery costs, with trade-offs among \textit{performance}, \textit{sample efficiency}, and \textit{computational efficiency}, as illustrated in Figure~\ref{fig:hrl_tradeoff}.

We first consider the trade-off between performance and sample efficiency. This trade-off can be appreciated from a theoretical point of view: the optimal policy can always be represented using primitive actions alone \citep{Bertsekas1995DynamicPA}, and restricting the agent to a predefined set of options may exclude optimal behavior \citep{AAMAS08-jong}.
However, learning the optimal policy for large and realistic environments is often intractable. 
Rather than pursuing perfect solutions, an agent could embrace efficient learning algorithms that find reasonable but often suboptimal policies. 
This is one of the main appeals of HRL: by re-composing existing solutions (i.e., behaviours achieving particular subtasks), an agent may be able to quickly find approximate solutions for a variety of tasks. For example, a pre-trained skill that opens doors allows an agent to bypass learning the complex motor controls for that specific action. However, this very abstraction can be limiting; if a door is stuck and requires an unusual push-and-jiggle motion, the rigid pre-defined skill might fail, preventing the agent from solving an edge case that a more flexible, low-level policy could have.\looseness-1

Another important challenge faced by agents interacting with complex and realistic environments is computational efficiency: the total compute needed to reach a target performance within a given sample budget. The compute includes learning, planning, action selection, and, when needed, discovering temporal abstractions. These computational demands interact with both performance and sample efficiency: an agent may extract more value from limited experience by spending additional computation on learning reusable behaviours or planning with a learned model, whereas reducing computation may compromise performance or increase the need for further interactions. Temporal abstraction can improve this balance by shortening the effective planning horizon. For example, a robot equipped with an LLM may plan over semantically meaningful skills rather than directly search over sequences of continuous motor commands. When these skills match the task, the agent can exploit their internal behaviours without repeatedly reasoning through low-level details. Planning with temporally extended actions can consequently require far fewer iterations to reach a near-optimal solution~\citep{mann2015approximate}, with efficiency gains depending on the structure exploited by the hierarchy~\citep{wen2020efficiency}. These savings, however, must be weighed against the costs of constructing and using the hierarchy. Selecting a skill may require deliberation across multiple levels, with costs depending on neural network size, the number and depth of Monte Carlo Tree Search rollouts~\citep{Coulom2006EfficientSA}, or the length of an LLM's reasoning trace. Executing a skill over multiple timesteps amortizes this cost, but may compromise performance when changing conditions require a different behaviour; more frequent deliberation can improve responsiveness at a higher computational cost. Discovering and refining the hierarchy likewise requires computation and may require additional samples, although existing experience can also be reused. 
Thus, HRL's three-way trade-off concerns how much computation and experience to invest in constructing and using a hierarchy to attain the desired performance, and whether its subsequent benefits amortize that investment. 
 
\paragraph{On the Importance of Prior Knowledge Reuse.} 
A common pitfall in HRL applications is that: \textit{discovering} a problem's hierarchical structure (see Section~\ref{sec:option-discovery}) can require more samples and computation than solving the problem directly using a ``flat'' agent, highlighting the importance of carefully carefully choosing when to apply HRL (see Section \ref{sec:8}).
This cost can be amortized in different ways. For example, it may be offset if the agent is expected to complete \textit{many different tasks within its lifetime}, allowing the learned subtasks to be potentially reused. 
Alternatively, reusing existing knowledge---such as  \textbf{offline datasets} (Section \ref{offline}) and \textbf{foundation models} (Section \ref{sec:largemodels})---can  help mitigate this cost. 
As we will see, HRL offers a natural and particularly promising way to incorporate such prior knowledge.

\section{Formalizing Hierarchical Reinforcement Learning}
\label{sec:3}
We use the notation introduced by  \citet{Sutton1998}: capital letters refer to random variables, whereas lowercase letters refer to their instantiation. Table~\ref{tab:notation_rl_hrl} summarizes the notation used in this section.

\subsection{Reinforcement Learning}
\label{sec:RL}
We consider an agent interacting with an environment where the agent is in state $S_t \in \mathscr{S}$ at timestep $t$, selects an action $A_t \in \mathscr{A}$, and in response the environment emits a scalar reward $R_{t+1}\in \mathbb{R}$ and transitions to a new state, $S_{t+1} \in \mathscr{S}$. This transition happens according to a transition probability distribution, 
\begin{equation}
    p(s'|s,a) = p(S_{t+1} = s' | S_t = s, A_t = a).
\end{equation}

The agent's goal is to find a policy $\pi: \mathscr{S} \to \Delta(\mathscr{A})$, where $\Delta(\mathscr{A})$ is the distribution over $\mathscr{A}$, that maximizes the expected discounted sum of rewards (return): 
\begin{equation}
G_t = \sum_{i=t}^\infty \gamma^{i-t} R_{i+1}, 
\end{equation}
where $\gamma \in [0,1)$ is the discount factor.  This 5-tuple, $\left<\mathscr{S}, \mathscr{A}, R, p, \gamma\right>$, defines a Markov Decision Process (MDP) \citep{Puterman1994MarkovDP}, the most commonly accepted formalism in RL. 

When following a particular policy $\pi$, the value of each state can be represented by the state value function,
\begin{equation}
v_{\pi}(s) = \mathbb{E}_{\pi}\left[ G_t  \given S_t = s\right].
\end{equation}
Similarly, we may consider the value of being in state $s$ and taking action $a$, following policy $\pi$ afterward, represented by the action-value function, or $q$-function,
\begin{equation}
q_{\pi}(s, a) = \mathbb{E}_{\pi}\left[ G_t  \given S_t = s, A_t = a\right].
\end{equation}
This function can be written recursively,  
\begin{align}
q_\pi(s, a) &= \mathbb{E} [ R_{t+1} + \gamma R_{t+2} + \gamma^2 R_{t+3} + \dots \mid S_t = s, A_t = a, \pi ]  \nonumber \\
&= r(s,a) + \gamma \sum_{s'} p(s'|s,a) v_\pi(s')  \nonumber \\
&= r(s,a) + \gamma \sum_{s'} p(s'|s,a) \sum_{a'} \pi(a'|s') q_\pi(s', a').
\end{align}
A similar derivation is possible for the value function, and this set of equations is referred to as the Bellman equations for evaluation \citep{Bellman1957DynamicP}.

The goal of an RL agent is to maximize the rewards it gets from interacting with the environment. In an MDP, there exists at least one optimal policy, defined as,
\begin{equation}
\pi^* = \arg \max_{\pi} q_{\pi}(s, a).
\end{equation}
In most practical settings, where the state or action spaces are large or continuous, the transition dynamics are unknown, or only sampled experience is available, this quantity is impractical to compute exactly, necessitating function approximation methods. 
Such approximations stem from two families of algorithms for learning reward-maximizing policies. The first family of methods, called value-based methods, greedily maximizes an estimated action-value function. Q-Learning \citep{Watkins1992Qlearning} is likely the most used algorithm to estimate the optimal policy, $\pi^*$, whose update rule takes the following form,
\begin{equation}
Q (S_t, A_t) \gets Q (S_t, A_t) + \alpha \left[ R_{t+1} + \gamma \max_{a \in \mathscr{A}} Q(S_{t+1}, a) - Q(S_t, A_t) \right].
\end{equation}
This update has been the basis of the Deep Q-Networks (DQN) algorithm \citep{Mnih2015HumanlevelCT}.\looseness-1

The second family of methods directly maximizes the quantity of interest, that is, the discounted sum of returns. The policy gradient theorem \citep{Sutton1999} provides the gradient of the expected discounted return from an initial state distribution, $d(s_0)$, with respect to a stochastic policy, $\pi_\zeta(\cdot|s)$, parameterized by $\zeta$, 
\begin{equation}
\deriv[J(\zeta)]{\zeta} = \sum_{s} d_\pi^{\gamma}(s) \sum_{a} \deriv[\pi_{\zeta}\left(a | s\right)]{\zeta}q_{\pi}(s, a),
\end{equation}
where  $d_\pi^{\gamma}(s) = \sum_{s_0} d(s_0) \sum_{t=0}^{\infty} \gamma^t \sum_a P_{\pi}(S_t = s | S_0 = s_0)$ is the discounted state occupancy measure, and  $P_{\pi}(S_t = s | S_0 = s_0)$ the probability of reaching state $s$ from $s_0$ in $t$ steps when following policy $\pi$. This update has been the basis of many modern algorithms, including the well-known proximal policy optimization \citep{Schulman2017ProximalPO}.

\subsection{Hierarchical Reinforcement Learning}
\label{hrl_formalism}

Hierarchical reinforcement learning extends the usual RL
formalism by explicitly representing and exploiting the
modular, compositional structure argued for in
Section~\ref{sec:2}. It does so by factoring the
policy $\pi$ into a library of sub-policies, each
governing behaviour over a stretch of time, and a
mechanism that determines how they are invoked. This
idea has a long history in AI, where such sub-policies
are commonly referred to as skills \citep{fikes1971strips, Thrun1994FindingSI}. Several
mathematical formalisms make this idea precise,
including options, goal-conditioned policies, and
others. We mainly adopt the options framework
\citep{Sutton1999BetweenMA, Precup2000TemporalAI} as it
provides a widely adopted and principled
formalism for expressing this structure. In
Section~\ref{sec:related_formalism}, we expand on how
alternative formalisms connect to the option framework.

\subsubsection{Options: A Mathematical Formalism}
\label{sec:options}
\begin{figure}
    \centering
    \includegraphics[width=0.9\linewidth]{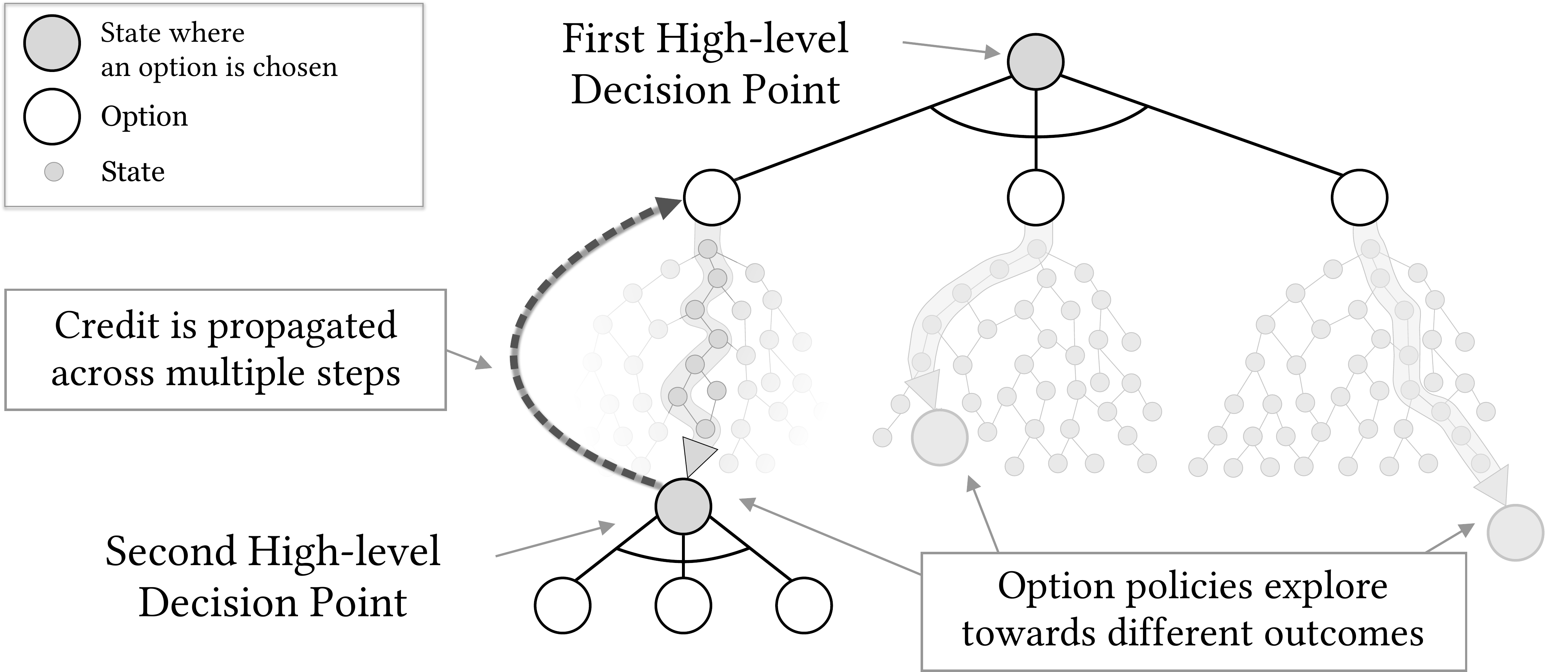}
    \caption{A simplified diagram illustrating the decision process of a hierarchical agent. 
    Large white nodes denote high-level decisions over options, which are assumed to be discrete here for simplicity.
    Large grey nodes represent the state observed by the agent when a high-level decision is made. 
    Grey trajectories represent state transitions during option execution, where the agent follows the option policy until it terminates.
    Different options can last for different durations and once an option terminates, the agent makes its next high-level decision.
    }
    \label{fig:hrl_diagram}
\end{figure}

An HRL agent makes use of a set of options, $\mathscr{O}$\footnote{The option set may be either (i) finite or countably infinite (discrete), or (ii) uncountably infinite, which can be equipped with a continuous parameterization (see Definition~\ref{def:option_id})~\citep{daSilva2012LearningPS}.},
which are defined by three components: a \textit{policy}, an \textit{initiation function}, and a \textit{termination function}.  These components can be implemented through parameterized functions, such as neural networks, or symbolically through code (e.g., see Section~\ref{sec:largemodels} and Section~\ref{sec:prog}). In Figure \ref{fig:hrl_diagram}, we illustrate the temporal structure exhibited by options while interacting with an environment. More formally,


\begin{definition}[Option]
\label{def:option}
An \emph{option} $o \in \mathscr{O}$ is a triple
$
o = \left<\mathscr{I}, \pi, \beta\right>,
$
where:
\begin{itemize}
    \item $\pi: \mathscr{S} \times \mathscr{O} \to \Delta(\mathscr{A})$ is an \emph{option policy},\footnote{ It can also be referred to as the skill's policy, the intra-option policy, or the goal-conditioned policy (see Section \ref{sec:related_formalism} for more details).} which selects an action\footnote{More generally, the option policy may select over both primitive actions and other options, i.e., $\pi: \mathscr{S} \times \mathscr{O} \to \Delta(\mathscr{A} \cup \mathscr{O})$, enabling more general hierarchical structures. We restrict option policies to primitive actions here, following the standard definition by \citet{Sutton1999BetweenMA} for simplicity.} according to the current state and the current option.\footnote{We slightly abuse the notation here with respect to the symbol representing the policy of a non-hierarchical RL agent. The distinction between the two will be clear through context. A separate notational convention concerns the option argument in these functions: it indexes the components associated with an option, rather than referring to the option triple itself. We retain $\mathscr{O}$ as the argument domain for notational convenience here and introduce explicit option identifiers later in Definition~\ref{def:option_id}.} When parameterized by weights $\theta$, we write $\pi_{\theta}(a \mid s, o)$.
    \item $\beta: \mathscr{S} \times \mathscr{O} \to [0,1]$ is the \emph{termination function}, giving the probability that an option $o$ terminates in state $s$. When parameterized, we write $\beta_{\psi}(s, o)$.
    \item $\mathscr{I}: \mathscr{S} \times \mathscr{O} \to \{0,1\}$ is the \emph{initiation set indicator}, specifying whether an option $o$ is permitted to start in state $s$.\footnote{In practical implementations, the binary initiation condition is relaxed into a continuous membership function for differentiable parameterization. In such cases, one may write $\mathscr{I}_\chi(s,o)\in [0,1]$ to represent the degree to which state $s$ belongs to the initiation set of option $o$.}
\end{itemize}
\end{definition}

Additionally, to select among the set of options, an HRL
agent often uses a \textit{high-level policy}
$\mu : \mathscr{S} \to \Delta(\mathscr{O} \cup
\mathscr{A})$, which outputs a probability distribution
over the set of options $\mathscr{O}$ and actions
$\mathscr{A}$ given a state $s$. When this probability
is parameterized, we use the notation
$\mu_{\kappa}(o|s)$, where $\kappa$ represents the high-level policy parameters. 
Similarly to the previous
components, in practice, this policy can also be
instantiated by other means, e.g., a programmatic policy
that directly encodes domain knowledge or follows
predefined rules (see Section \ref{sec:prog}). 


When planning with options, an agent can make use of an
\emph{option model}, comprising a transition model,
$P_{\mathscr{O}} \colon \mathscr{S} \times \mathscr{O}
\times \mathscr{S} \times [0,1] \to [0,1]$, which gives
the discounted probability that option~$o$, initiated in
state~$s$, will terminate in state~$s'$, marginalizing
over all possible durations, and a reward model,
$R_{\mathscr{O}} \colon \mathscr{S} \times \mathscr{O}
\times [0,1] \to \mathbb{R}$, which gives the expected
cumulative discounted reward collected during the option's
execution. Both quantities depend on the discount factor~$\gamma$
and enable planning at the level of options without reasoning
over individual timesteps \citep{Sutton1999BetweenMA}.

Not all of the papers covered in this work will explicitly instantiate each of these components. It is common for research in HRL to only highlight the components for which a significant contribution is made and to make assumptions about the other components. For instance, the termination function can be assumed to output termination after a fixed number of timesteps. Similarly, the option initiation function is often assumed to allow option initiation across the whole state space. We will discuss this further in Section~\ref{sec:option-discovery} and highlight the relevant aspects within the presentation of each method.

Using these terms, we can now define the option value function,
\begin{align}\label{eq:option-value}
q_{\pi}(s, o) &= \sum_a \pi(a \mid s, o) q_u(s, o, a),
\end{align}
where $q_u : \mathscr{S}  \times \mathscr{O}  \times \mathscr{A}  \to \mathbb{R}$ is the value of executing action $a$ in the context of a state-option pair:\looseness-1
\begin{align}
q_u(s, o, a) &= r(s, a) + \gamma \sum_{s'} p(s' \mid s, a) u_{\beta}(o, s') .
\end{align}
The function $u_{\beta} : \mathscr{O} \times \mathscr{S} \to \mathbb{R}$ is called the option-value function upon arrival, that is, the value of executing option $o$ upon entering a state $s'$ is given by:
\begin{align}
\label{eq:arrival}
u_{\beta}(o, s') &= (1 - \beta(s', o)) q_{\pi}(s', o) + \beta(s', o) v_{\mu}(s').
\end{align}
Finally, the function $v_{\mu} : \mathscr{S} \to \mathbb{R}$ is defined as the value function over a set of options,  
\begin{align}\label{eq:value-over-options}
v_{\mu}(s) &= \sum_o \mu(o|s)q_{\pi}(s, o).
\end{align}
We say an option is \textit{executed} when, upon its selection, the agent takes actions according to its policy $\pi(\cdot \mid s, o)$, terminating at each state $s'$ with probability $\beta(s', o)$. How options are translated into action selection in this way, and when the agent reconsiders its choice of options, is called the \textit{execution model} and we discuss it in detail in Section~\ref{sec:5}. Equations~\ref{eq:option-value}--\ref{eq:value-over-options} hold under the general \emph{intra-option} setting, in which the high-level policy and option policies are evaluated at every timestep; a notable special case is \textit{call-and-return}, and we discuss that next.


\paragraph{The Semi-Markov Decision Process (SMDP).}
Building on an MDP, $\left<\mathscr{S}, \mathscr{A}, R, p, \gamma\right>$, an SMDP allows decisions to persist for a random, variable duration, so that both transitions and rewards are defined over temporally-extended intervals \citep{bradtke94smdp}.

Notably, when a high-level policy $\mu(o \mid s)$ selects an option and the agent executes it until termination (the \textit{call-and-return} execution model), the induced decision process over options forms an SMDP \citep{Sutton1999BetweenMA}. In this case, the decision space is the (fixed) option set $\mathscr{O}$, and the transition kernel is given by
\begin{equation}
p(s', \tau \mid s, o),
\end{equation}
where $\tau$ is the (random) duration until termination governed by the option policy $\pi_o$ and option's termination function $\beta$. The corresponding option-level reward aggregates discounted rewards accrued during execution:
\begin{equation}
r(s, o)
=
\mathbb{E}\!\left[
\sum_{k=0}^{\tau-1} \gamma^k R_{t+k+1}
\;\middle|\;
S_t = s,\ O_t = o
\right].
\end{equation}

Because the state at each option-initiation point is a sufficient statistic for future evolution, the SMDP inherits the convergence guarantees of standard dynamic-programming and RL algorithms, now applied at the level of options rather than primitive actions. The practical consequence is that the agent plans over a much shorter effective horizon: a sequence of hundreds of primitive steps may reduce to a handful of option-level decisions, making both learning and planning substantially more tractable. 


In what follows, we will primarily adopt the standard MDP formulation rather than the SMDP formalism for two reasons: (i) the SMDP view relies on the call-and-return assumption, which does not hold when options can be interrupted or when the agent combines multiple option policies to select primitive actions;\footnote{These execution models will be discussed in Section~\ref{sec:5}.} and (ii) an induced MDP representation can be obtained by marginalizing over the option duration $\tau$, which can be sufficient in practice. 
More generally, in hierarchical settings with multiple levels, each level makes decisions whose effective durations are determined by lower-level dynamics and termination mechanisms, so these durations need not be modeled explicitly at that level.
That is, given an MDP and a set of options, an SMDP can be induced (cf. the result ``MDP + options = SMDP” by~\citet{Sutton1999BetweenMA}).

\subsubsection{Option Discovery}
\label{sec:option-discovery}
In the preceding sections, we introduced options (Definition~\ref{def:option}) and studied prediction and control given a fixed, predefined set of options. In general, however, such a set might not be specified a priori and must instead be \emph{discovered} from data. This gives rise to the problem of \emph{option discovery},\footnote{Also termed ``skill discovery,'' encompassing temporal abstraction discovery beyond the option formalism.} namely the algorithmic construction of options.


\begin{definition}[Option Discovery]
\emph{Option discovery} is the problem of constructing a set of options $\mathscr{O}$, i.e., specifying the option triple $\left<\mathscr{I}, \pi, \beta\right>$  for each option $o\in\mathscr{O}$.\looseness=-1
\end{definition}



Option discovery often seeks to construct an option set with desired properties, such as supporting effective task performance (Section~\ref{sec:learn_from_rewards}), exhibiting behavioural diversity (Section~\ref{sec:empowerment_skill_discovery}), enabling faithful trajectory reconstruction (Section~\ref{sec:maxlike}), facilitating exploration, or supporting transfer across tasks (Section~\ref{sec:formalguarantee}).

Constructing a set of options is combinatorially challenging, as it requires jointly specifying a collection of option triples~\citep{sutton2023reward}.
Consequently, practical approaches adopt additional design principles that make the search space more tractable. We first highlight common and largely orthogonal design principles in the literature, and then introduce an algorithm template for HRL with option discovery in Algorithm~\ref{alg:hrl_generic}.

\smallskip
\paragraph{Design Principle 1: One objective per option.}
Although the option definition itself does not require optimizing any objective, many approaches focus on \emph{purposeful} options, where each option is designed to achieve an \textit{associated objective}, analogous to how an RL agent is driven by a reward function \citep{littmanrewardhypothesis}. 
We denote the objective associated with an option $o$ by $\omega_o \in \Omega$, where $\Omega$ is a set of admissible objectives.
In such cases, the option triple (typically the policy) can be derived or learned from this objective.
We refer to  options with associated objectives as \textit{subgoal options}~\citep{bagaria2024skill}, and a large body of work on option discovery can be understood from this perspective \citep[e.g.,][]{mcgovern2001automatic,Precup2000TemporalAI,
colas2022autotelic,sutton2023reward}.

A common and general way to specify such an objective is through an \emph{option reward}.
\begin{definition}[Option Reward]
\label{def:option_rew}
An \emph{option reward function} is a mapping
\[
r^o \colon \mathscr{S} \times \mathscr{A}\to \mathbb{R},
\]
which specifies a reward signal conditioned on an option $o$. When parameterized, we denote the option reward by $r^o_\nu$ with parameters $\nu$.
\end{definition}

This signal is distinct from the environment reward~$r(s,a)$ because it is internal to an 
option on the agent side.
It is also referred to as ``\textit{pseudo-reward}”~\citep{dietterich1998maxq} or ``\textit{subgoal reward}”~\citep{bagaria2024skill} in the literature. The corresponding option policy $\pi(\cdot \mid s, o)$ is typically obtained by (approximately) maximizing the expected return induced by $r^o$. However, not all subgoal options are defined through reward maximization: some approaches derive option behaviour from demonstrations \citep[e.g.,][]{
simateam2024scalinginstructableagentssimulated}, and others specify it through symbolic methods such as code (see Sections~\ref{sec:largemodels}
and~\ref{sec:prog}). What unifies these approaches is that the option's behaviour is purposeful, however the objective is specified.
    
\paragraph{Design Principle 2: Constrained initiation and termination.} 
The search space can be restricted by imposing structure on (a subset of) the option triple $\left<\mathscr{I}, \pi, \beta\right>$, thereby reducing degrees of freedom in the specification. 
Examples include fixing the initiation function to constrain admissible states (e.g., \citet{Bacon2016TheOA} assume $\forall s,\ \mathscr{I}(s)=1$), and restricting the termination function to control when options can end.
A common instance of the latter is to augment termination with a maximum step count $H_o$, introducing a clock that forces termination after a fixed horizon. This can be viewed as allowing $\beta$ to depend on an internal time index, thus rendering the option non-Markov. An alternative that preserves the Markov property is to use a constant per-timestep termination probability (e.g., $\beta = 1/H_o$), yielding the same expected duration while preserving the Markov property. Both formulations can be interpreted as restricting the class of admissible termination functions.

\paragraph{Design Principle 3: Structured option representation.}
The option space can also be restricted by representing options with an identifier $z \in \mathscr{Z}$ whose structure, via a shared parametric form (e.g., $\pi(a \mid s, z)$), enables \textit{generalization} across options (e.g., via geometry or semantic meaning), which resembles the idea of function approximation (see Section~\ref{sec:RL}). 
In earlier sections, we have used $o$ to refer to the identifier naming an option when no ambiguity arises. 
We now make this distinction explicit by separating the notion of an option from how it is denoted.

\begin{definition}[Option Identifier]
\label{def:option_id}
An \emph{option identifier} is an element $z \in \mathscr{Z}$ that uniquely denotes an option, with an associated option
$
o_z = \left<\mathscr{I}_z, \pi_z, \beta_z\right>.
$
\end{definition}

This abstraction isolates the combinatorial object (the option triple) from its parameterization. In the literature, the identifier $z$ is commonly instantiated as follows:\looseness-1

\begin{itemize}
\item \textit{Option index and latent variable.} In the classical options framework, options are elements of a finite, discrete set $\mathscr{O}$, and $z$ is simply an index (often denoted as ``$i$'') selecting $o \in \mathscr{O}$ \citep{Sutton1999BetweenMA}. 
In some methods, $z$ is sometimes treated as a latent variable, sampled from a prior distribution and inferred from interaction via a learned posterior or discriminator \citep{eysenbach2019diversity,kipf2019compile}. This enables discovering options via variational or mutual-information objectives. In particular, this inference enforces only that $z$ is separable from different behaviour. Thus, in either case, $\mathscr{Z}$ can impose a strong prior by restricting the option space to a fixed, typically small set of behaviours.\looseness-1

\item \textit{Continuous representation or embedding.} Some approaches endow $z$ with geometric structure, viewing it as a continuous representation in which nearby identifiers correspond to behaviourally similar options \citep{kim2019variational}. This supports interpolation and generalization across options.

\item \textit{Goal or subgoal.} In goal-conditioned formulations, $z$ corresponds to a goal or subgoal (often presented as ``$g\in \mathscr{G}$''), and options are defined conditionally on $z$, with the policy and sometimes the termination function derived from that goal specification \citep{schaul2015universal, Andrychowicz2017HindsightER}. This imposes a semantic interpretation on $z$. See the discussion on goal-conditioned RL in Section~\ref{sec:related_formalism}.

\item \textit{Program, code, and natural language}. In some approaches, $z$ lies in a space of symbolic descriptions that compositionally specify procedural behaviour, e.g., sequences or trees of primitives \citep{andreas2017modular,Konidaris2018FromST,sun2020program}. Here, $\mathscr{Z}$ carries compositional (syntactic) structure with procedural semantics, in contrast to goal-based formulations that specify desired outcomes. Additionally, 
$z$ can also be expressed in natural language \citep{Jiang2019LanguageAA}, which can exhibit compositional structure, albeit less formally than symbolic representations (see Section~\ref{sec:largemodels}).
In such approaches, options are derived from $z$ through an additional decoding mechanism, and $z$ often affords interpretability (Section~\ref{sec:benefits}).
\end{itemize}

\begin{algorithm}[t]
\caption{Hierarchical Reinforcement Learning with Option Discovery.}
\label{alg:hrl_generic}
\begin{algorithmic}[1]
\Require Environment $Env$, options $\mathscr{O}$,
         high-level policy $\mu$,
         data $\mathscr{D}$,
         option objectives $\Omega$.
\Procedure{OptionDiscovery}{$\mathscr{O}, \mathscr{D}, \Omega$}
    \State $\Omega \leftarrow
      \textsc{OptionSpecification}(\mathscr{O}, \mathscr{D})$
      \Comment{Define or refine option objectives}
    \State $\mathscr{O} \leftarrow
      \textsc{OptionLearning}(\mathscr{O}, \mathscr{D}, \Omega)$
      \Comment{Optimize option components}
    \State \Return $\mathscr{O}, \Omega$
\EndProcedure

\Statex \Comment{\textbf{Main} procedure below}
\For{$t = 0, 1, \dots$}
    \State $\mathscr{D} \leftarrow
      \textsc{Interact}(Env, \mathscr{O}, \mu, \mathscr{D})$
      \Comment{Collect experience}
    \State  $\mathscr{O},\Omega \leftarrow \textsc{OptionDiscovery}(\mathscr{O}, \mathscr{D}, \Omega)$ \Comment{See lines 1--5}
    \State $\mu \leftarrow
      \textsc{PolicyImprovement}(\mu, \mathscr{O}, \mathscr{D})$
      \Comment{Improve the high-level policy}
\EndFor
\end{algorithmic}
\end{algorithm}


Bringing the preceding components together,
Algorithm~\ref{alg:hrl_generic} 
 outlines a generic HRL template with option discovery, with diverse instantiations detailed in Sections~\ref{tabularasa}--\ref{sec:largemodels}. It interacts with an environment~$Env$ and is initialized with a set of options~$\mathscr{O}$, a high-level policy~$\mu$, available data~$\mathscr{D}$ (possibly empty), and an initial set of option objectives~$\Omega$ (possibly empty, described in Section~\ref{sec:option-discovery}). 
Each iteration (lines 7--9) operates as follows.

\begin{itemize}
\item
\textsc{Interact} (line 7) collects data by executing actions according to the options and the high-level policy, appending experiences to~$\mathscr{D}$; in offline settings, this step can be vacuous since the dataset is fixed in advance. This entails specifying how options (and, if applicable, the high-level policy) are used to generate primitive actions, i.e., the execution model (Section~\ref{sec:5}).

\item
\textsc{OptionDiscovery} (line 8, implemented in lines 1--4) maps the current set of options $\mathscr{O}$, data $\mathscr{D}$, and option objectives $\Omega$ to updated options $\mathscr{O}'$ and objectives $\Omega'$. The specification step determines the objectives, and learning step produces the behaviour that achieves them. 

\item
\textsc{OptionSpecification} (line 2) defines or refines the set of objectives, $\Omega$, that options should achieve. 
In many cases where Design Principle 1 in Section~\ref{sec:option-discovery} holds, each option objective determines \textit{what} the associated option should do, for example by specifying an option reward function~$r^o$ (Definition~\ref{def:option_rew}), or a subgoal $g$. 
Methods without this assumption instead often use a single shared objective for all options, in which case this procedure always outputs a fixed objective.

\item
\textsc{OptionLearning} (line 3) optimizes the option components---policies~$\pi(\cdot|s,o)$, termination functions~$\beta$, and initiation functions~$\mathscr{I}$---given those objectives $\Omega$. 


\item
\textsc{PolicyImprovement} (line 9) updates the
high-level policy~$\mu$ to better select among the
current options, if a high-level policy is applicable.
\end{itemize}

To illustrate how these operations interact concretely,
consider the example of a robot that should
acquire an option for grasping an object.
\textsc{OptionSpecification} would identify grasping as
a useful subgoal and define a corresponding option
reward~$r^o$ that provides a positive signal when the
object is held. \textsc{OptionLearning} would then train
the option policy~$\pi(\cdot|s,o)$ to maximize~$r^o$
and derive appropriate termination and initiation
conditions---perhaps to initiate the option when the
object is in the vicinity of the robot, and terminating
when it has been successfully grasped.
\textsc{PolicyImprovement} would update~$\mu$ to learn
when selecting the grasping option is useful for the
task at hand.

The template is deliberately general; different methods vary in how they instantiate each operation and in design choices. First, $\mathscr{D}$ denotes the data available to the agent and may consist of newly collected online experience, a growing replay buffer, a fixed offline dataset, or any combination thereof; the template does not prescribe how data is obtained. 
Second, the ordering and frequency
of the operations may differ: some methods interleave
all four operations at every iteration, while others
separate discovery into a distinct phase---calling
\textsc{OptionSpecification} \emph{before} deploying the
resulting options objectives, possibly without further modification.
More generally, these correspond to different update ratios, where one specification step may be followed by multiple learning steps.
As we will see in
Sections~\ref{tabularasa}--\ref{sec:largemodels}, both
paradigms are common in the literature. Third, not all
methods explicitly invoke every operation. Some specify options with
fixed objectives and only learn the option components (see Algorithm~\ref{alg:oc_inst} and~\ref{alg:offline_skill_latents});
others expand the option set without ever revising
existing options (see Section~\ref{sec:relationship-bottleneck}); others still do not learn a high-level
policy at all, relying instead on programmatic or
language-based planners (see Section~\ref{sec:prog}) or generalized policy improvement (GPI)-style updates (see Section~\ref{sec:option_composition_gpi}).
What the template makes explicit is that, regardless of the specific method, 
\textbf{HRL with option discovery can be understood as a combination of these operations.}

\subsubsection{Related Terminologies and Formalisms}
\label{sec:related_formalism}

The previous section uses the language of options to formalize temporal abstractions. 
As the field of HRL is rich and diverse, some researchers may feel misrepresented by such a formalism. 
Therefore, we adopt the options framework as main formalism throughout, but use ``options”, ``skills”, and ``goal-conditioned policies” following the terminology used in the original work, and occasionally refer more broadly to temporal abstractions.
We believe that, despite differences in terminology, formalism, and representation, the underlying ideas are closely aligned.
We now highlight the differences among various related formalisms and illustrate their connections to options.


\paragraph{Skills.} The skill terminology has largely been used \textit{informally} in the HRL literature to refer to temporally-extended behaviours. 
Skills can be formalized using the options framework \citep[e.g.,][]{konidaris2009skill}, but they can also be formalized using other formalisms such as macro-actions \citep{fikes1971strips}, feudal hierarchies \citep{dayan1993feudal}, partial policies \citep{Thrun1994FindingSI}, MAXQ \citep{dietterich1998maxq}, and HAMs \citep{parr1997reinforcement}. \citet{Barto2003RecentAI} compare those different formalisms.
Moreover, in recent work, this term is also used to denote an option identifier (Definition~\ref{def:option_id}), such as a latent variable or embedding (see Section~\ref{sec:empowerment_skill_discovery} and Section~\ref{offline} for details).

\paragraph{Feudal RL.} In Feudal RL \citep{dayan1993feudal}, decision-making is divided across multiple levels of the hierarchy, where higher-level ``managers'' set subgoals for lower-level ``workers'' who are rewarded by their managers for achieving these subgoals. 
Managers operate under information hiding, relying on an abstracted state representation that omits lower-level details handled by the workers. See Section~\ref{sec:learn_from_rewards} for details.

\paragraph{HAMs.} Hierarchies of Abstract Machines
\citep{parr1997reinforcement} constrain the agent's
behaviour through partial programs specified as
finite-state machines (with a call hierarchy). The key idea is that some decisions are fixed by the program while others are left as \emph{choice points} where the agent must learn what
to do. For example, a HAM might fix that after picking
up an object the agent must carry it to a door, but
leave the agent to learn \emph{which} object to pick up.
This reduces the effective policy space: the product of
a HAM and an MDP yields a reduced SMDP whose optimal
policy is the best policy consistent with the
constraints imposed by the machine
\citep{parr1997reinforcement}. A sequence of actions
between consecutive choice points implicitly defines an
option, but unlike the options framework, HAMs
additionally constrain the order in which
temporally-extended behaviours can be invoked through
the machine's control flow. This structure must be
specified by the designer.

\paragraph{MAXQ.} The MAXQ framework
\citep{dietterich1998maxq} decomposes the task into a
hierarchy of subtasks organized as a directed acyclic
graph. Its key insight is a recursive decomposition of
the action-value function:
$q(s, i) = v_o(s) + c(s, i)$, where $v_i(s)$ is the
value of completing subtask~$M_i$ and $c(s, i)$ is the
\textit{completion function}---the expected reward
after~$M_i$ terminates under the parent policy in the
hierarchy. This
decomposition enables state abstraction within each
subtask, since $v_i(s)$ need only depend on state
variables relevant to~$M_i$. 
In a nutshell, the designer specifies a hierarchy of subtasks, which induces the recursive value-function decomposition. The value functions and policies for the subtasks are then learned within this fixed hierarchy. Once learned, each subtask can be viewed as exhibiting option-like behaviour during execution.


\paragraph{General Value Functions (GVFs).} 

Value-based RL can be viewed as learning a single piece of predictive knowledge. By contrast, \textit{general value functions} (GVFs) $v_{\pi}^c(s)$ extend this notion by allowing arbitrary \textit{cumulants} $c \in \mathcal{C}$, thereby representing a rich collection of predictions about the agent’s observation stream~\citep{sutton2011horde}. Formally, a GVF is defined as
\begin{equation}
v_{\pi}^c(s) = \mathbb{E}_\pi \Big[ \sum_{t=0}^\infty \Big( \prod_{k=0}^{t-1} \gamma(s_k) \Big) c(s_t, a_t, s_{t+1}) \,\Big|\, s_0 = s \Big],    
\end{equation}
where the cumulant $c$ specifies the signal of interest and the discount $\gamma : \mathscr{S} \to [0,1]$ may depend on the state, enabling termination-like semantics.

This contrasts with procedural knowledge, which specifies how to act, as embodied by options, skills, or behaviours discussed earlier. Procedural knowledge corresponds to policies (e.g., $\pi$, or option policies $\pi_o$) and their execution semantics (e.g., initiation and termination), whereas predictive knowledge corresponds to value functions or models that estimate consequences of behaviour. GVFs thus occupy the predictive side: they do not directly prescribe behaviour, but instead encode structured predictions that can inform or support procedural components. 
In this sense, GVFs and options are complementary rather than disjoint abstractions: options specify temporally extended behaviour, while GVFs encode predictions about the consequences of that behaviour. Moreover, prior work has shown that options themselves can be specified or characterized through GVF-based predictive knowledge, including predictions about subgoals, termination events, or expected outcomes~\citep{sutton2023reward}.
\looseness-1

\paragraph{Goal-conditioned RL (GCRL).} 
A goal can be formally defined using a triple: $(g, r^g, \gamma_g)$, where $g:  \mathscr S \rightarrow \mathbb{R}^d$ is a goal vector that can, for example, be used to condition a policy, $\pi(a|s,g)$, or a value function, $v(s|g)$, $r^g: \mathscr S\rightarrow\mathbb{R}$, is a goal reward function that maps each state to real-valued number, and, finally, $\gamma_g: \mathscr S \times \mathscr A \times \mathscr S\rightarrow [0, 1]$ describes the goal's continuation function, and hence the timescale for achieving that goal \citep{kaelbling1993learning,schaul2015universal}.
Many works in \textit{goal-conditioned} RL (GCRL) consider the state space to be the set of goals an agent should reach (e.g., \citealp{Andrychowicz2017HindsightER}), although the framework itself is more general and can admit goal-spaces that are distinct from the environment's state-space \citep{schaul2015universal}. 

It is natural to view GCRL through the lens of GVFs. In particular, each goal $g$ specifies a goal reward $r^g$ and continuation function $\gamma_g$, thereby inducing a goal-contingent prediction about future outcomes. Thus, GCRL can be understood as learning a collection of related GVFs indexed by goals, where each goal defines a distinct predictive question.

A direct consequence of this view is that one could, in principle, learn a separate approximator for each goal-induced GVF, as in the \textit{Horde} architecture \citep{sutton2011horde}. However, since goals typically exhibit structure (as described in Section~\ref{sec:option-discovery}), this motivates sharing representations across goals. Universal Value Function Approximators (UVFAs) \citep{schaul2015universal} exploit this structure by representing the family of goal-indexed GVFs with a single function approximator, e.g., $V_{\phi}: \mathscr S \times \mathscr G \rightarrow \mathbb{R}$, alongside a goal-conditioned policy $\pi_{\theta}: \mathscr S \times \mathscr G \rightarrow \Delta(\mathscr{A})$ that generalizes behaviour across goals. 
Although some of the works from the GCRL literature are present in this work (e.g., Section~\ref{sec:curriculum} and \ref{sec:subgoalgen}), we refer the reader to \citet{Liu2022GoalConditionedRL} for an in-depth review. 


\subsubsection{Beyond Architectural Choices}

In the previous section, we noted that options, skills, goal-conditioned policies, and related constructs reflect the same underlying principle, but exhibit nuanced differences in their level of formalization and usage.
We now clarify this statement by examining how these constructs manifest across different architectural instantiations.

HRL is often associated with a particular architecture.
A prominent example is a \emph{modular hierarchical architecture}, where a high-level policy selects among options, each parameterized separately. Another widely used instantiation is the \emph{goal-conditioned policy}, where a single policy network is conditioned on a goal variable (potentially specified by a foundation model such as an LLM; see Section~\ref{sec:largemodels}). 
An HRL system may even consist solely of a single, monolithic neural network, where options or goals are not explicitly represented but implicitly encoded in the parameters~\citep{kobayashi2025emergent}.\looseness-1



Despite these architectural differences, all these instantiations share a common principle: they introduce variables (e.g., options or goals) that parameterize temporally-extended behaviour. This commonality indicates that hierarchy arises from \textit{how behaviour is defined and learned}, rather than how the agent is structured. Thus, we argue that \textbf{HRL is fundamentally defined through the algorithm, not the agent architecture.}

An HRL algorithm could empower the agent's exploration by directing behaviour through sub-policies that operate at different timescales. It can also facilitate more effective credit assignment by decomposing a long, continuous stream of experience into meaningful subtasks. Additionally, HRL can better prepare an agent for future challenges by promoting the learning of reusable behaviours, which can be explicitly or implicitly defined. These essentially represent the core benefits of HRL, as outlined earlier in this work in Section \ref{sec:benefits}. 

\section{Discovery from Online Experience}
\addtocontents{toc}{\protect\setcounter{tocdepth}{2}}
\label{tabularasa}

In this section, we present work in option discovery that takes place in the \textit{online} setting: the agent seeks to construct useful options by simply interacting with the environment. This setting has received significant attention because it holds the promise of scalability \citep{SuttonBitterLesson}---a long-lived agent that can learn new, useful options simply via interaction and can potentially keep increasing its competence in the world, bootstrapping new skills with previously discovered ones \citep{ring95continual,schmidhuber10formal}.

\newcommand{\dotone}{{\large\ding{108}}}
\newcommand{\dottwo}{{\large\ding{108}\!\ding{108}}}

\begin{table}[h]
    \centering
    \caption{A summary of HRL methods shown in Section \ref{tabularasa}, \ref{offline}, and \ref{sec:largemodels} that discover temporally abstract behaviours, \textbf{highlighting the main benefits} elaborated in Section \ref{sec:benefits}. Each method links to the corresponding section. A single black dot (\checkit) indicates that a class of methods generally contributes to addressing a specific challenge, while a double black dot (\checkit\checkit) signifies that the class of methods is explicitly designed to tackle that challenge.}
    \renewcommand{\arraystretch}{1.5}
    \resizebox{\linewidth}{!}{%
    \begin{tabular}{@{}>{\centering\arraybackslash}p{3.5cm} >{\centering\arraybackslash}p{4.5cm} >{\centering\arraybackslash}p{3cm} c c c@{}}
        \toprule
        \multirow{2}{*}{\centering \textbf{Categories}} &
        \multirow{2}{*}{\centering \textbf{Methods}} &
        \multirow{2}{*}{\centering \textbf{Reference}} &
        \multicolumn{3}{c}{\textbf{Focus of Methodologies on Different Benefits}} \\
        \cmidrule(lr){4-6}
        & & & Credit Assignment & Exploration & Transferability \\
        \midrule

        \multirow{9}{*}{\parbox{3.5cm}{\centering \textbf{Discovery from \\ Online Experience}}}
                                      & \makecell{Bottleneck \\ Discovery}               &  \ref{sec:od/graph-based} & \checkit & \checkit & \checkit \\ \cmidrule(lr){2-6}
                                      & \makecell{Spectral \\ Methods}                   &  \ref{sec:od/spectral} & \checkit & \checkit\checkit & \checkit \\ \cmidrule(lr){2-6}
                                      & \makecell{Skill \\ Chaining}                     &  \ref{sec:skillchaining} & \checkit\checkit &  & \checkit \\ \cmidrule(lr){2-6}
                                      & \makecell{Empowerment \\ Maximization}           &  \ref{sec:empowerment_skill_discovery} & \checkit & \checkit\checkit & \checkit \\ \cmidrule(lr){2-6}
                                      & \makecell{Via \\ Environment Reward}             &  \ref{sec:learn_from_rewards} & \checkit\checkit & & \checkit \\ \cmidrule(lr){2-6}
                                      & \makecell{Curriculum Learning}  &  \ref{sec:curriculum} & \checkit & \checkit\checkit & \checkit \\ \cmidrule(lr){2-6}
                                      & \makecell{Meta Learning}                         &  \ref{sec:metalearning} & \checkit & \checkit & \checkit\checkit \\ \cmidrule(lr){2-6}
                                      & \makecell{Intrinsic Motivation}                  &  \ref{sec:other_im_od} & & \checkit\checkit & \checkit \\ \cmidrule(lr){2-6}
                                      & \makecell{Directly Optimizing the \\ Benefits of HRL} &  \ref{sec:formalguarantee} & \checkit & \checkit & \checkit \\
        \midrule

        \multirow{2}{*}{\parbox{3.5cm}{\centering \textbf{Discovery through \\ Offline Datasets}}}
                                      & \makecell{Variational \\ Inference}              &  \ref{sec:maxlike} & \checkit & \checkit & \checkit \\ \cmidrule(lr){2-6}
                                      & \makecell{Hindsight Sub-goal \\ Relabeling}      &  \ref{sec:subgoalgen} & \checkit\checkit & & \\
        \midrule

        \multirow{4}{*}{\parbox{3.5cm}{\centering \textbf{Discovery with \\ Foundation Models}}}
                                      & \makecell{Embedding \\ Similarity}               &  \ref{sec:embsim} & & \checkit\checkit & \checkit \\ \cmidrule(lr){2-6}
                                      & \makecell{Providing \\ Feedback}                 &  \ref{sec:provfeedback} & \checkit & \checkit\checkit & \checkit \\ \cmidrule(lr){2-6}
                                      & \makecell{Reward as \\ Code}                     &  \ref{sec:rewardascode} & & & \checkit\checkit \\ \cmidrule(lr){2-6}
                                      & \makecell{Directly Modeling \\ the Policy}       &  \ref{sec:directmodelingpolicy} & & \checkit & \checkit\checkit \\
        \bottomrule
    \end{tabular}%
    }
    \label{tab:hrl_benefits}
\end{table}
We broadly categorize this literature based on the proxy objectives and principles used for option discovery. For each family of methods, we first describe the core intuition and key methodological patterns. Then, we discuss how each category contributes to the core benefits of HRL (as outlined in Section~\ref{sec:benefits}). Finally, we discuss some limitations of each category and highlight opportunities for research.

Before presenting the methods in detail, we direct the reader's attention to Table~\ref{tab:hrl_benefits}, which provides an overview of all the discovery methods discussed in this work. For each method, we highlight which benefits have been mostly studied by researchers from the field, where a single black dot (\checkit) indicates that a class of methods generally contributes to addressing a specific challenge, while a double black dot (\checkit\checkit) signifies that the class of methods is explicitly designed to tackle that challenge.

\subsection{Bottleneck Discovery}\label{sec:od/graph-based}

Many challenging problems in RL have \textit{bottlenecks}: small regions of states that an agent must pass through to reach a larger, potentially more interesting region of the state space. For example, in the \textit{Two Rooms} task \citep{Sutton1999BetweenMA, Solway14Optimal}, the agent must go through a doorway state to access the goal in the other room. Another example is a player in a video game who must pick up a key to unlock a door that leads to other levels. In these examples, the doorway and the key act as bottlenecks---reaching those states grants the agent access to an entirely new region of interesting states.

Algorithm~\ref{alg:bottleneck} shows how bottleneck discovery instantiates
the generic template of Algorithm~\ref{alg:hrl_generic}.
\textsc{OptionSpecification} analyzes the dataset~$\mathscr{D}$ to identify
a set of bottleneck states~$B$ and, for each bottleneck~$b \in B$, defines a
subgoal reward~$r^o_b$ that fires upon reaching~$b$. The resulting set of
option objectives $\Omega = \{(b, r^o_b)\}_{b \in B}$ is passed to
\textsc{OptionLearning}, which trains an option policy~$\pi_b$ to
maximize~$r^o_b$ for each bottleneck, derives the corresponding termination
and initiation conditions, and adds the complete option to the option
set~$\mathscr{O}$. Bottleneck
methods typically specify $\beta_b$ and $\mathscr{I}_b$ directly based on the subgoal; for other discovery methods (e.g., the option-critic;
Section~\ref{sec:learn_from_rewards} or skill chaining; Section~\ref{sec:skillchaining}), they are optimized incrementally; but for simplicity, we choose to put them in the same subroutine either way. \textsc{PolicyImprovement} then
updates the high-level policy~$\mu$ to select among the available options,
and is unchanged from the generic template.

\begin{algorithm}[t]
\caption{Bottleneck Discovery: Instantiation of Algorithm~\ref{alg:hrl_generic}.}
\label{alg:bottleneck}
\begin{algorithmic}[1]
\Procedure{OptionSpecification}{$\mathscr{O}, \mathscr{D}$}
    \State $B \leftarrow \textsc{IdentifyBottlenecks}(\mathscr{D})$
        \Comment{e.g., diverse density, min-cut, betweenness}
    \State $\Omega \leftarrow \varnothing$
    \For{$b \in B$}
        \State $r^o_b(s) \leftarrow \mathbbm{1}(s = b)$
            \Comment{Example subgoal reward}
        \State $\Omega \leftarrow \Omega \cup \{(b, r^o_b)\}$
    \EndFor
    \State \Return $\Omega$
\EndProcedure
\Statex
\Procedure{OptionLearning}{$\mathscr{O}, \mathscr{D}, \Omega$}
    \For{$\omega = (b, r^o_b) \in \Omega$}
        
        \State $\beta_b(s) \leftarrow \mathbbm{1}(s = b)$
            \Comment{Example termination condition}
        \State $\mathscr{I}_b(s) \leftarrow 1$
            \Comment{Example initiation condition}
        \State Train option policy $\pi_b(\cdot|s)$ to maximize $r^o_b$.
        \State $\mathscr{O} \leftarrow \mathscr{O} \cup \{(\mathscr{I}_b, \pi_b, \beta_b)\}$
    \EndFor
    \State \Return $\mathscr{O}$
\EndProcedure
\end{algorithmic}
\end{algorithm}

The specific choices of~$r_b^o, \beta_b$ and~$\mathscr{I}_b$ in
Algorithm~\ref{alg:bottleneck} are representative defaults. For example,
some methods restrict initiation to states within the partition leading to the
bottleneck~\citep{menache2002q}, and implementations often augment the
termination condition with a maximum step count to avoid indefinite
execution. What matters most across bottleneck methods is the
\textsc{IdentifyBottlenecks} subroutine: the remainder of this section describes the major approaches.

Most algorithms for finding bottlenecks begin with a graph-based view of the MDP: states are treated as nodes and an edge exists between two states $(s, s')$ when the agent can transition from $s$ to $s'$ in a single timestep:
\begin{equation}
    G=(\mathscr{S}, E), \quad e_{s, s'} \in E \iff \sum_{a\in\mathscr{A}}p(s'|s, a) > 0.
\end{equation}
This graph is constructed from transitions stored in the dataset $\mathscr{D}$, which may have been collected by calls to the \textsc{Interact} sub-routine or provided in advance. In this graph, bottlenecks have been identified using the following approaches.

\subsubsection*{Diverse Density}

An early approach to option discovery used the concept of diverse density, which measures how much more likely a state is to lie on a successful trajectory than an unsuccessful one. \citet{mcgovern2001automatic} formulate bottlenecks as states with high diverse density and propose a simple algorithm to identify them. Concretely, consider that the agent has a set of successful trajectories $\mathcal{T}^+$ (sequences of states leading to a goal state) and a set of unsuccessful trajectories $\mathcal{T}^-$ (sequences that did not reach the goal state). The diverse density score $\text{DD}(s)$ captures the probability that a given state $s$ occurs in successful trajectories and does not occur in unsuccessful trajectories:
\begin{equation}
    \text{DD}(s) = \prod_{\tau \in \mathcal{T}^+} P(s \in \tau) 
    \prod_{\tau \in \mathcal{T}^-} \Big(1-P(s \in \tau) \Big),
\end{equation}
where the probability that a state occurs in a trajectory can be computed in tabular domains using visitation counts:
\begin{equation}
    P(s \in \tau) = \frac{\text{Number of times } s \text{ appears in } \tau}{|\tau|}.
\end{equation}
In practice, this product is computed in log-space to avoid numerical underflow. States with a diverse density greater than a threshold are chosen as bottlenecks, and subgoal options are created to reach them. A drawback is that trajectories must be classified as positive or negative depending on whether they were on the path to a goal state. \citet{Stolle2002LearningOI} address this shortcoming by defining diverse density over a \textit{family of tasks}: bottleneck states are those that are repeatedly visited while solving many goal-reaching tasks.\looseness=-1
    
\subsubsection*{Graph Partitioning}

Under the graph view of MDPs, bottlenecks can be interpreted as ``accumulation'' nodes---states through which many paths coincide. These states tend to separate loosely connected subgraphs that are otherwise densely connected internally. To see why, \citet{menache2002q} cast the problem of going from a start state $s$ to a goal state $g$ as a Max-Flow problem \citep{ahuja1993network}: the agent should maximize the flow of probability along paths originating at $s$ and terminating at $g$. By the Max-Flow Min-Cut theorem \citep{ford1962flows}, this is equivalent to finding the lowest-capacity edges whose removal completely separates $s$ from $g$.\looseness-1

\begin{figure}
    \centering
    \begin{subfigure}[b]{0.45\textwidth}
        \centering
        \includegraphics[width=\textwidth,page=1]{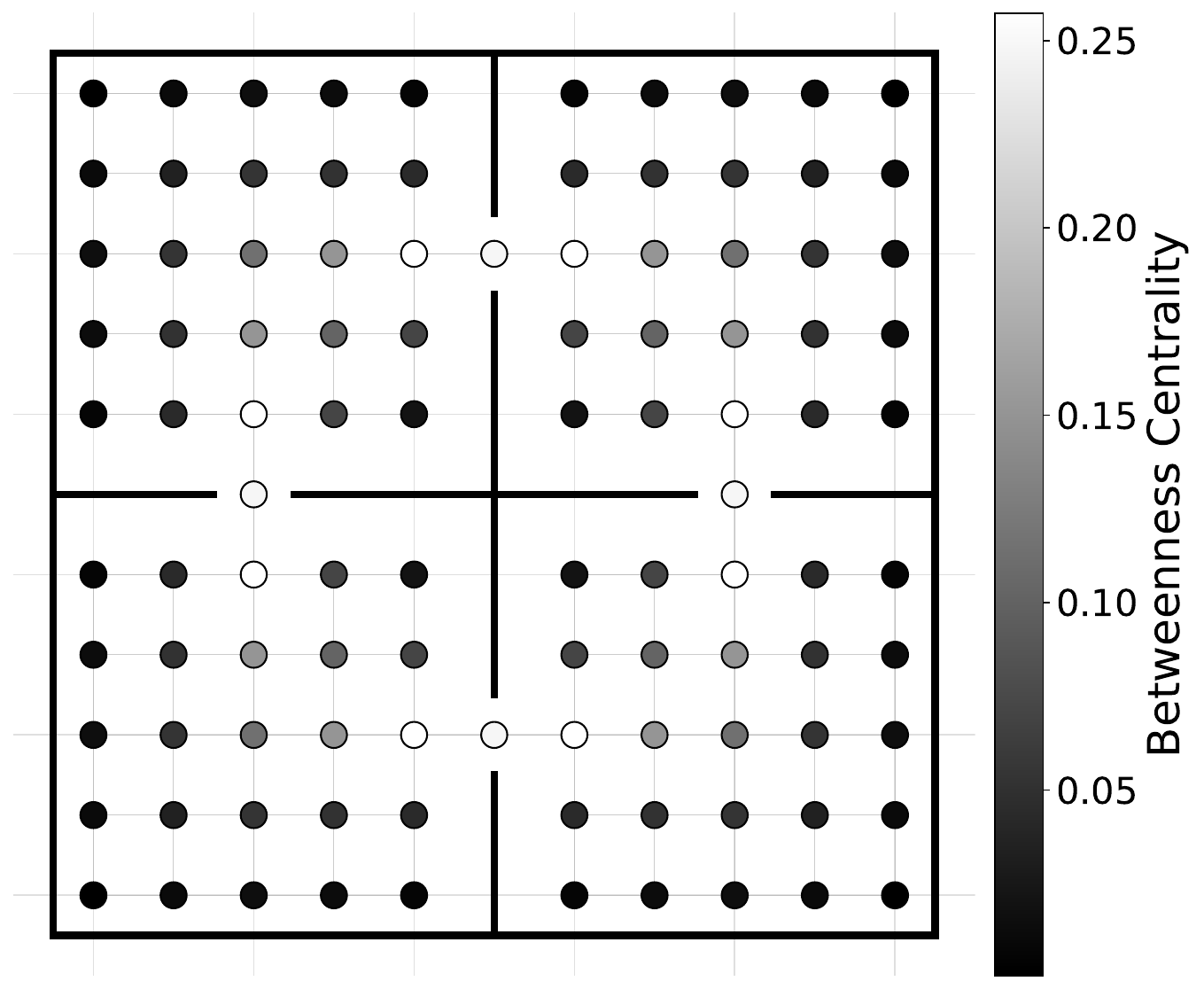}
        \vspace{0.5cm}
        \caption{Betweenness Centrality}
        \label{fig:pdf1}
    \end{subfigure}
    \hfill
    \begin{subfigure}[b]{0.45\textwidth}
        \centering
        \includegraphics[width=0.88\textwidth,page=1]{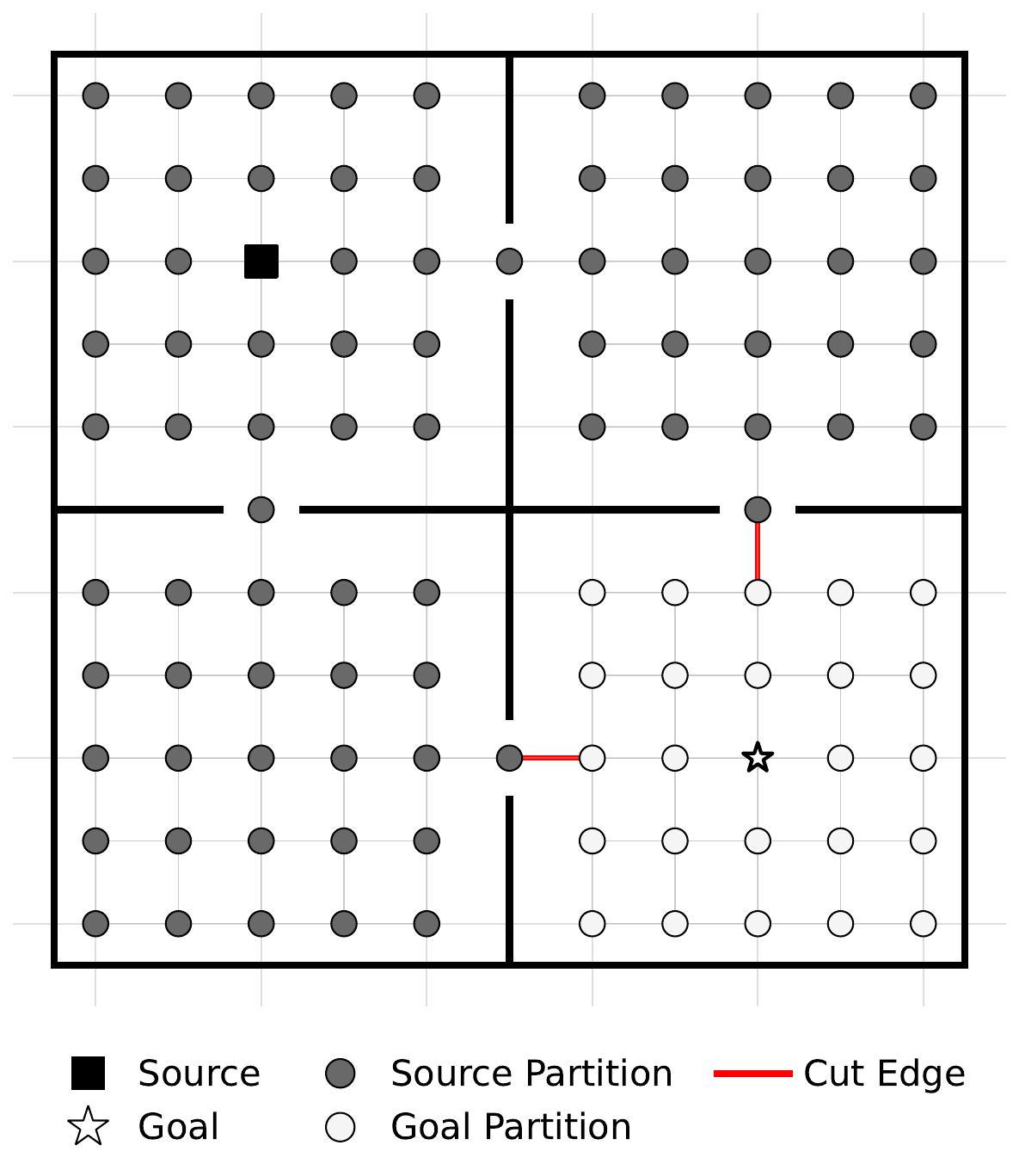}
        \caption{Q-Cut}
        \label{fig:pdf2}
    \end{subfigure}
    \caption{\textbf{Bottleneck Discovery in Four Rooms.} Skill discovery using (a) betweenness centrality, a measure of the likelihood that a state lies on the shortest path between any two other states, and (b) Q-cuts, which finds the edge that solves the Min-Cut problem on the transition graph. Both classes of methods attempt to identify bottleneck states and use them as option subgoals.}
    \label{fig:both-pdfs}
\end{figure}

Specifically, the Q-Cut algorithm \citep{menache2002q} finds bottlenecks by solving the Min-Cut problem. The nodes of the graph are partitioned into two disjoint sets $U$ and $\bar{U} = \mathscr{S}\setminus U$, such that the source state belongs to $U$ and the goal state belongs to $\bar{U}$. The cut-value is defined as the total transition probability along edges crossing the partition:
\begin{align}\label{eq:q-cut}
    \text{Cut}(U, \bar{U}) = \sum_{\substack{(i,j)\in E:\\ i\in U,\; j\in \bar{U}}} \sum_{a \in \mathscr{A}} p(j|i,a).
\end{align}
The min-cut is the partition that minimizes this quantity:
\begin{align}
    U^* &= \argmin_{U\subset\mathscr{S}:\; s\in U,\; g\notin U} \text{Cut}(U, \mathscr{S}\setminus U).
\end{align}
Although there are exponentially many valid partitions, the Min-Cut problem can be solved in polynomial time \citep{ford1962flows}. The bottlenecks are defined as the destination nodes of the min-cut edges: $B=\{j \mid (i, j) \in E,\; i \in U^*,\; j \notin U^*\}$.

A drawback of Q-Cut is that it requires a global graph over the entire state space, which is not scalable. To address this, L-Cut \citep{csimcsek2005identifying} constructs local graphs from states visited within a single episode. Rather than searching for individual bottleneck states, \citet{mannor2004clustering} suggest identifying clusters of states and connecting them using options; this approach has been extended using more sophisticated clustering techniques \citep{metzen12online,srinivas2016option,campos2020edl,Bacon2013OnTB}. Notably, \citet{evans2023creating}'s use of graph modularity \citep{newman2004finding} as the clustering metric allows them to efficiently learn multi-level hierarchies, where each level operates at a different timescale.\looseness-1

\subsubsection*{Graph Centrality}

In graph theory, centrality measures the importance of each node within a graph. The search for useful subgoals in an MDP can be viewed as analogous to identifying central nodes. Centrality measures are theoretically well-understood and several efficient algorithms exist for computing them, making them attractive for option discovery. Although many centrality measures exist, \citet{simsek2008betweenness} advocate for \textit{betweenness} centrality because of its ability to find bottlenecks in large graphs. Betweenness quantifies how often a node appears on shortest paths between other nodes. Specifically, the betweenness score $b(v)$ for a state $v$ is:\looseness-1
\begin{equation}\label{eq:betweenness}
    b(v) = \sum_{\substack{s, t \in \mathscr{S} \\ s \neq t,\; s\neq v,\; t\neq v}} \frac{\sigma_{st}(v)}{\sigma_{st}}\, w_{st},
\end{equation}
where $\sigma_{st}$ is the number of shortest paths from $s$ to $t$, $\sigma_{st}(v)$ is the number of those paths passing through $v$, and $w_{st}$ is a weight assigned to the pair $(s,t)$. When $w_{st}=1$ for all pairs, Equation~\ref{eq:betweenness} recovers the standard betweenness centrality measure on graphs. To tailor this measure to MDPs, \citet{simsek2008betweenness} set $w_{st}$ to the expected reward obtained while going from $s$ to $t$.

\subsubsection{Relationship to the Algorithmic Template}
\label{sec:relationship-bottleneck}

It is worth noting how bottleneck methods relate to the phasing distinction raised in Section~\ref{sec:option-discovery}---that is, whether \textsc{OptionSpecification} is interleaved with data collection or run once as a separate, upfront phase. 
In principle, bottleneck discovery can be interleaved with interaction, as illustrated by L-Cut \citep{csimcsek2005identifying}, which constructs local graphs from each episode and incrementally updates the set of bottlenecks as the agent explores; this effectively interleaves \textsc{OptionSpecification}, \textsc{Interact}, and \textsc{PolicyImprovement}. However, many methods instead adopt a modular pipeline, where they first collect data, construct the transition graph, identify bottlenecks, and define subgoal options before deploying them. In such cases, \textsc{OptionSpecification} runs once, which may limit adaptability when new regions of the state space are discovered.\looseness-1


\subsubsection{Benefits and Opportunities}


\paragraph{Exploration.} If an agent can reliably reach bottleneck states, it can perform more effective exploration, because states that were once hard to reach become accessible even under a random policy (\citealp{Simsek2004UsingRN} refer to these states as \textit{access states}). For example, picking up a key makes it easy for a video game player to visit previously unseen rooms. When bottleneck discovery is performed incrementally, the agent progressively expands the frontier of its experience in the environment.

\paragraph{Credit Assignment.} Methods like those of \citet{mcgovern2001automatic} and \citet{simsek2008betweenness} require the agent to solve the problem several times before option discovery can begin; in such cases, exploration is clearly not the main benefit. However, once the agent identifies bottlenecks, it can perform rapid credit assignment for three reasons: (a) value can propagate in large, multi-step jumps from option termination states to option initiation states, (b) value from rewarding events only needs to propagate along trajectories passing through the bottleneck, greatly reducing the state-action pairs whose values need updating, and (c) in long-horizon problems, the difference in value between actions---the \textit{action-gap} \citep{farahmand2011action,bellemare2016actiongap}---tends to vanish \citep{lehnert2018lhp}; partitioning the state space along bottlenecks creates shorter-horizon subtasks with larger action-gaps, easing credit assignment \citep{lehnert2018lhp}.

\paragraph{Transfer.} Bottlenecks are useful for transfer because they are largely task-agnostic---they capture structure in the transition function rather than the reward function, so the same bottlenecks are often useful across a family of tasks. For example, in the \textit{Two Rooms} task, the ability to reliably reach the doorway enables the agent to reach the goal regardless of where it is placed in the second room \citep{mcgovern2001automatic}. \citet{Solway14Optimal} offer formal support for this view; we discuss their analysis in Section~\ref{sec:formalguarantee}.

\paragraph{Opportunities for Research.}

\begin{itemize}
    \item \textit{Scalability.} Most methods for finding bottlenecks apply to discrete, explicitly constructed graphs; as a result, they struggle to scale to large, continuous MDPs. In terms of Algorithm~\ref{alg:hrl_generic}, the bottleneck is in \textsc{OptionSpecification}, which requires building and analyzing the full transition graph. A scalable approach includes spectral methods (Section~\ref{sec:od/spectral}), which can compute bottleneck-relevant, continuous properties of the underlying graph without explicitly representing it.
    \item \textit{Performance guarantees.} With some notable exceptions \citep[e.g.,][who relate bottleneck-aligned hierarchies to learning efficiency across tasks]{Solway14Optimal}, it is generally not well understood how targeting bottlenecks contributes to high-level objectives such as reward maximization or faster planning. Methods outlined in Section~\ref{sec:formalguarantee} address this question for option discovery in general, but given the number of bottleneck-specific algorithms, it would be valuable to identify whether there is a high-level objective that is directly optimized (at least approximately) by this family of proxy objectives.
\end{itemize}

\subsection{Spectral Methods}\label{sec:od/spectral}

Many option discovery methods are based on the idea of leveraging the state space's topology, be it to discover options that identify key states that connect different partitions of the environment \citep{csimcsek2005identifying}, that connect states that are far from each other when looking at the diffusion properties of the environment \citep[e.g.,][]{machado2016learning,machado2017laplacian}, or that easily allow the agent to traverse the environment in a reusable manner \citep{liu2017eigenoption,Klissarov2023DeepLO}. 
They are termed spectral methods because they extract information about the state space, such as connectivity or diffusion, through the eigenvectors of a matrix representation of the environment.

The different algorithms in this group leverage the different ways of representing the environment as a matrix and the different types of information one can extract from such matrices. Originally, heavily inspired by results from the graph theory literature, these methods were based on the graph Laplacian and its eigenfunctions,\footnote{
Eigenfunctions can be seen as a generalization of eigenvectors to continuous state spaces.} which can approximate any function on the graph \citep{chung1997spectral}. The normalized graph Laplacian, $\mathcal{L}$, for example, is defined as 
\begin{equation}\label{eqn:laplacian}
    \mathcal{L} = \mathbf{D}^{-1/2} (\mathbf{D} - \mathbf{A}) \mathbf{D}^{-1/2},
\end{equation}
where $\mathbf{A}$ is the graph's adjacency matrix obtained by modeling each state in the environment as a node. The adjacency matrix reflects the degree of connectivity between two states. The matrix $\mathbf{D}$ is a diagonal matrix whose entries are the row sums of $\mathbf{A}$. In the reinforcement learning literature, the eigenvectors of the graph Laplacian are also known as proto-value functions \citep[PVFs;][]{mahadevan2005proto,mahadevan2007proto}. 

Importantly, when considering the eigendecomposition, $\mathcal{L}\mathbf{e} = \lambda \mathbf{e}$, the eigenvector of the graph Laplacian associated to the second smallest eigenvalue captures the number of connected components in a graph~\citep{shi2000normalized}, allowing one to easily identify bottleneck  states~\citep{csimcsek2005identifying}, as discussed in Section \ref{sec:od/graph-based}. The eigenvectors of the graph Laplacian, in general, capture different time scales of diffusion, which can be used to discover options that promote exploration, such as eigenoptions~\citep{machado2017laplacian,machado2018eigenoption,machado2019efficient}, covering options \citep{pmlr-v97-jinnai19b,jinnai2020Exploration}, and covering eigenoptions~\citep{machado2023temporal}.

Eigenoptions, for example, are defined such that each option, $o_i$, is associated with the corresponding eigenvector, $\textbf{e}_i$, of the graph Laplacian. Their policy is defined as the policy that maximizes the intrinsic reward that incentivizes the agent to navigate alongside the direction pointed by $\textbf{e}_i$, which, in the linear function approximation (and tabular) case, is formalized as
\begin{equation}
\label{eq:lfa-eigen-reward}
    r^{\textbf{e}_i}(s,s') = \textbf{e}_i^\top \big(\boldsymbol{\phi}(s') - \boldsymbol{\phi}(s)\big),
\end{equation}
where $\boldsymbol{\phi}(s)$ denotes the feature representation of state $s$. Originally, an option $o_i$ was defined to terminate in state $s$ if $q_\pi^{\textbf{e}_i}(s,a) \leq 0$ for all $a \in \mathscr{A}$, where $q_\pi^{\textbf{e}_i}$ is defined w.r.t. $r^{\textbf{e}_i}(\cdot, \cdot)$. All other states in the environment were defined to be in the initiation set.

\begin{figure}[t]
    \centering
    \includegraphics[width=\linewidth]{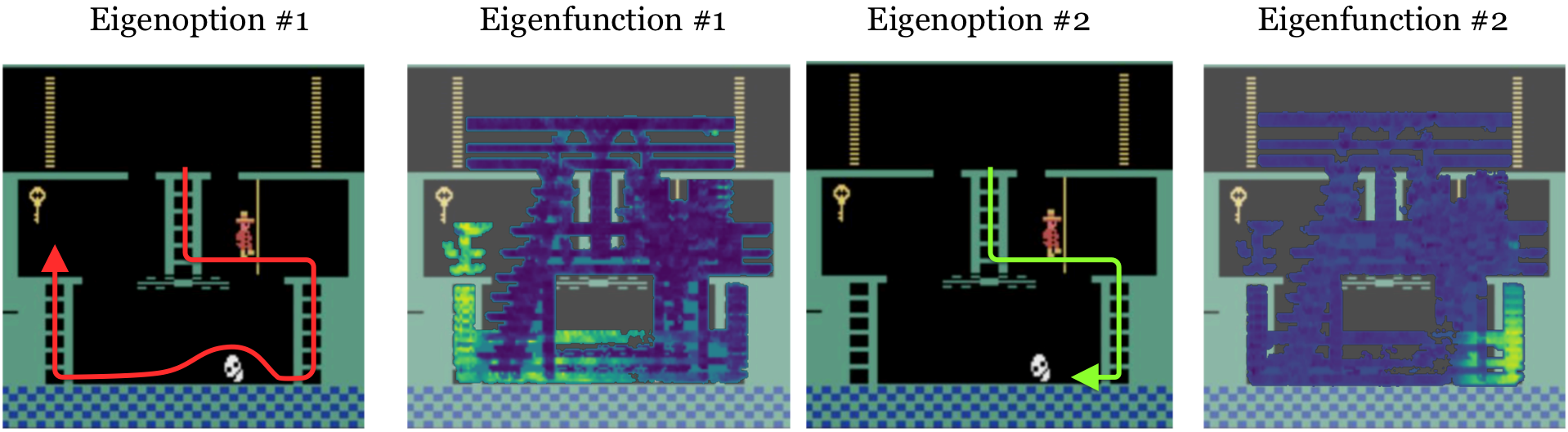}
    \caption{Visualization of the first and second eigenfunctions on Montezuma's Revenge, an Atari 2600 game, discovered by the algorithm proposed by \citet{Klissarov2023DeepLO}. The arrows depict what the eigenoption induced by these eigenfunctions could end up being.}
    \label{fig:eigenoptions}
\end{figure}

Naturally, explicitly representing an environment through its underlying graph is not scalable. Existing methods leverage approximations of the eigenfunctions of the graph Laplacian that can be obtained through neural networks trained with stochastic gradient descent~\citep{wu2019laplacian,wang2021towards,Gomez2023ProperLR}. The underlying idea behind these methods is to learn a representation that captures the properties of the approximated eigenvectors such that observations that happen ``close in time'' are close in representation space and that different eigenfunctions are indeed orthogonal to each other.  The current state-of-the-art method for doing so is called the \emph{augmented Lagrangian Laplacian objective}~\citep[ALLO;][]{Gomez2023ProperLR}. It consists of the following max-min objective for approximating $d$ eigenfunctions:
\begin{align}
    \max_{\boldsymbol{\omega}}\min_{\mathbf{u}\in\mathbb{R}^{d \times |\mathscr{S}|}} \vspace{2pt}
    &
    \sum_{i=1}^{d} \langle \mathbf{u}_i,\mathcal{L} \mathbf{u}_i \rangle 
    + \sum_{j=1}^{d}\sum_{k=1}^{j}\omega_{jk}\big(\langle\mathbf{u}_{j}, \llbracket\mathbf{u}_{k}\rrbracket\rangle - \delta_{jk}\big)
    + b \sum_{j=1}^{d}\sum_{k=1}^{j}\big(\langle\mathbf{u}_{j}, \llbracket\mathbf{u}_{k}\rrbracket\rangle - \delta_{jk}\big)^2\,,
\end{align}
where $\mathcal{L}$ denotes the graph Laplacian again, $\llbracket \cdot \rrbracket$ the stop gradient operator, $\delta_{jk}$ the Kronecker delta, $b$ is a scalar hyperparameter, and $\boldsymbol{\omega} =[\omega_{1,1},\omega_{2,1},\omega_{2,2},\cdots,\omega_{d,1},\cdots,\omega_{d,d}]\in\mathbb{R}^{d(d+1)/2}$ is a vector containing all of the dual variables of the objective. Note that the optimal dual variables, $\boldsymbol{\omega}^*$, are proportional to the smallest
eigenvalues of $\mathcal{L}$. These approximations have now been used to learn options that are effective in various domains, including continuous control tasks~\citep{jinnai2020Exploration,shehmar2026alps,chandrasekar2026lk}, 3D navigation tasks, and Atari 2600 games  (result shown in Figure~\ref{fig:eigenoptions})~\citep{Klissarov2023DeepLO}. An issue these methods had to circumvent was that most of these approximation objectives assume the ability to sample uniformly the entire state space. This is currently addressed by iteratively increasing the region covered by the agent~\citep[e.g.,][]{machado2023temporal}; some methods even do so explicitly in the objective they minimize~\citep{erraqabi2022tact}. \looseness=-1

The process to compute the intrinsic reward maximized by the option is slightly different when using neural network estimates of the eigenfunctions of the graph Laplacian. Instead of first computing the eigenvectors, one usually directly estimates the components of the eigenfunction associated with a particular state, $s$. Formally, 
\begin{equation}
\label{eq:eigenfunc-reward}
    r^{f_{e_i}}(s,s') = f_{e_i}(s') - f_{e_i}(s),
\end{equation}
where we used $f_{e_i}(s)$ to denote the value of $i$-th eigenfunction of the graph Laplacian associated with state $s$. In this setting, stochastic option terminations are more common in practice due to the difficulties generalization introduces to accurately estimating action-value functions without interference \citep[e.g.,][]{Klissarov2023DeepLO}.

Many other mathematical objects are somewhat equivalent to the eigenvectors of the graph Laplacian and have also been used for option discovery. Slow Feature Analysis~\citep[SFA;][]{Wiskott02Slow,Sprekeler11Relation}, for example, are a key component of Continual
Curiosity-driven Skill Acquisition~\citep[CCSA;][]{Kompella17Continual}. The eigenvectors of the successor representation~\citep[SR;][]{dayan93successor} have also been shown to be equivalent to the eigenvectors of the graph Laplacian~\citep{machado2018eigenoption}.

The equivalence between the eigenvectors of the SR and of the graph Laplacian is particularly important due to the predictive nature of the SR and the ease with which one can learn it incrementally. In fact, the SR now has a quite prominent role in the option literature, being used in the discovery of options for both faster credit assignment~\citep{ramesh2019successor} and exploration~\citep{machado2018eigenoption,machado2019efficient}. 

The successor representation is defined as 
\begin{equation}
    \boldsymbol{\Psi}_{\pi}(s, s') = \mathbb{E}_{\pi, p} \left[ \sum_{t=0}^{\infty} \gamma^t \mathbbm{1}_{\{S_t = s'\}} \mid S_0 = s \right],
\end{equation}
where $\mathbbm{1}$ denotes the indicator function. The SR was originally introduced through an intuition that is very similar to the one outlined above: one should capture the environment's dynamics by assigning similar values to temporally close states, thus creating a representation of the underlying structure. 
It can also be estimated with temporal-difference learning methods~\citep{Sutton1988LearningTP}, which, as we mentioned above, allows us to learn it incrementally:
\begin{equation}
\Psi(S_t, j) \leftarrow \Psi(S_t, j) + \eta \Big(\mathbbm{1}_{\{S_t = j\}} + \gamma \Psi(S_{t+1}, j) - \Psi(S_t, j) \Big),
\end{equation}
where we used $\Psi(\cdot,\cdot)$ to denote a sample-based approximation of $\boldsymbol{\Psi}_{\pi}$.

Importantly, beyond the discovery methods mentioned above; as a representation, which was its original purpose, the SR can also be used to combine options without additional learning~\citep{barreto2021optionkeyboard}, and recent results in neuroscience and cognitive sciences suggest the SR can model activations in the hippocampus~\citep{stachenfeld2017hippocampus,cothi2020neurobiological,gomez2025representations} and explain some human behaviour~\citep{momennejad2017successor}. These results have led \cite{machado2023temporal} to propose that the successor representation should be seen as the ``natural substrate for the discovery and use of temporal abstractions'' in RL.

In terms of scalability, again, there have been many proposals on how to scale the SR to function approximation settings ranging from specific neural network architectures~\citep{kulkarni2016deep,machado2018eigenoption,chua2024learning} to ideas such as successor features~\citep{barreto2017successor} and successor measures~\citep{touati21successor,farebrother2023protovalue}.  Successor features, for example, can be seen as a projection of the SR onto the space realizable by the representation, $\boldsymbol{\phi}$. In matrix form, if we use  $\mathbf{\Phi} \in \mathbb{R}^{|\mathscr{S}|\times d}$ to denote the matrix encoding the $d$-dimensional feature representation of each state, successor features are defined as $\mathbf{\Psi}_\pi = \sum_{t=0}^\infty (\gamma \mathbf{P}_\pi)^t \mathbf{\Phi} = (I - \gamma \mathbf{P}_\pi)^{-1} \mathbf{\Phi}$.




\begin{algorithm}[t]
\caption{Spectral Option Discovery: Instantiation of 
Algorithm~\ref{alg:hrl_generic}.}
\label{alg:spectral_inst}
\begin{algorithmic}[1]
\Procedure{OptionSpecification}{$\mathscr{O}, \mathscr{D}$}
    \State Update or define the graph representation from $\mathscr{D}$.
    \Statex \Comment{E.g., eigenvectors or eigenfunctions of the graph Laplacian $\mathcal{L}$  in Eq.~\ref{eqn:laplacian}}
    \State Update pseudo-rewards, $\Omega=\{r^{e_i}\}$.
    \Statex\Comment{E.g., $r^{\mathbf{e}_i}(s,s')$ in Eq.~\ref{eq:lfa-eigen-reward} or $r^{f_{e_i}}(s,s')$ in Eq.~\ref{eq:eigenfunc-reward}}
    \State \Return $\Omega$
\EndProcedure
\Statex
\Procedure{OptionLearning}{$\mathscr{O}, \mathscr{D}, \Omega$}
    \State For each intrinsic reward $r^{e_i}$ in $\Omega$, learn an option policy $\pi(\cdot\mid s, o=i)$.
    \Statex\Comment{Example initiation and termination: }
    \Statex\Comment{$\forall s,\ \mathscr{I}(s)=1$, terminate with probability $\frac{1}{H_o}$ for each step}
    \State \Return $\mathscr{O}$
\EndProcedure
\end{algorithmic}
\end{algorithm}

\subsubsection{Relationship to the Algorithmic Template}
Algorithm~\ref{alg:spectral_inst} provides a general instantiation of Algorithm~\ref{alg:hrl_generic} for spectral methods.
The algorithm has two key steps. 
First, it constructs or updates a graph using the experiences from the dataset $\mathscr{D}$, computes or updates graph representation (e.g., Laplacian eigenvectors), and converts each component into a pseudo-reward that measures progress along that direction in the state space. 
Second, for each such pseudo-reward, it learns a policy that maximizes it, yielding a set of option policies, each corresponding to a distinct spectral mode of the environment’s structure. For initiation and termination, we show an example from~\citet{Klissarov2023DeepLO}, which corresponds to the discussion in Design Principle 2 of Section~\ref{sec:option-discovery}.\looseness-1

\subsubsection{Benefits and Opportunities}

\paragraph{Exploration.}
The eigenoptions line of work~\citep{machado2017laplacian} has popularized the the use of temporal abstraction for exploration, alongside studies that highlighted its benefits, including \citet{brunskill2014pac} and \citet{fruit2017exploration}. Eigenoptions can significantly decrease the diffusion time\footnote{The diffusion time encodes the expected number of ``decisions'' required to navigate between two states randomly chosen in an environment~\citep{machado2017laplacian}.} in an environment, and this afforded exploration can lead to faster learning. \citet{machado2018eigenoption} further extends previous work to the function approximation case by estimating the successor representation and then performing a singular value decomposition on it. \citet{pmlr-v97-jinnai19b} introduce covering options, arguing that rather than constructing an option for every eigenvector of the graph Laplacian, a single option constructed based on the second eigenvector is sufficient.
This is because that single option minimizes the cover time of the underlying MDP, which loosely refers to how long it takes for a random high-level policy to visit all states. Leveraging direct approximations of the eigenfunctions of the graph Laplacian, \citet{jinnai2020Exploration} extended covering options to the function approximation case, and  \citet{Klissarov2023DeepLO} did the same for covering eigenoptions~\citep{machado2023temporal}, demonstrating strong exploration properties in a variety of reinforcement learning problems. More recently, \cite{theophilou2025novel} have shown how to augment the Laplacian representation with state-visitation counts, obtaining options that lead to even better exploration in the tabular case, while also being able to better deal with asymmetric environments. \looseness=-1


\paragraph{Transferability.} Options are often thought to be important in continual (lifelong) learning settings (See also Section~\ref{sec:continual-rl}) where skills can be reused in an ever-changing world. The benefit of Laplacian-based options in such settings has been demonstrated both in simpler tabular problems in which the goal location changes regularly~\citep{liu2017eigenoption} and in more complex, high-dimensional settings in which not only the goal location would change but also the topology of the environment~\citep{Klissarov2023DeepLO}.

\paragraph{Opportunities for Research.}
\begin{itemize}
    \item \textit{Improving Representations.} \citet{machado2023temporal} have proposed the perspective that spectral methods consist of a phase in which a representation is first learned (e.g., PVFs, SR), followed by a phase in which options are then derived from such a representation. This process can even be done in a cycle, which  \citet{machado2023temporal} called Representation-driven Option Discovery (ROD) cycle. Thus, better representation learning methods are an exciting research frontier for this line of work in which the learned representation informs the option discovery process. This can be investigated from the SR perspective~\citep[e.g.,][]{touati21successor,carvalho2022composing,farebrother2023protovalue}, or from the perspective of directly estimating the spectral decomposition of the SR~\citep[e.g.,][]{Pfau2018SpectralIN,wang2021towards,Wang2022ReachabilityAwareLR,Gomez2023ProperLR}, including non-symmetric settings~\citep{Wang2023OptimalGR}.
    \item \textit{Planning.} Another promising line of work involves further exploring the recent success of Laplacian-based methods in planning and credit assignment in general, as these options are often used in a reward-agnostic way~\citep[e.g.,][]{sutton2023reward}. Validating these results beyond the tabular case and extending existing results to partially-observable settings are also intriguing lines of work.
    \item \textit{Reward-Aware Representations.} The representations discussed in this section rely on the topology of the environment without considering the underlying reward function. There is an interesting question of whether one should define proximity not only in terms of when observations take place but also in terms of the reward associated with them. Interestingly, the linear MDP formalism~\citep{todorov2006linearly,todorov2009efficient} gives rise to representations akin to the SR, but that are reward-aware. In this context, \cite{tse2025reward} has shown that options derived from the eigenvectors of such a reward-aware representation, termed the default representation~\citep{piray2021linear}, exhibit qualitatively different exploratory behaviour when faced with regions of negative reward in the state space. More recently, \cite{tse2026drogo} has proposed an objective that allows us to use neural networks to estimate the principal eigenvector of the default representation.
\end{itemize}

\subsection{Sequentially Composable Options}\label{sec:skillchaining}

Options are said to be \textit{sequentially executable} when each option terminates in a region where another option can successfully achieve its own subgoal. Sequentially composable options are more useful for high-level planning \citep{Konidaris2018FromST} and can result in highly robust solutions \citep{tedrake2010lqr}. While most methods attempt to sequentially compose discovered options \textit{post hoc}, some methods explicitly incorporate sequential composition into the option discovery objective. A prominent family of such methods is that of Skill Chaining \citep{konidaris2009skill,bagaria2020option}.

In the language of Algorithm~\ref{alg:hrl_generic}, skill chaining is distinctive in that \textsc{OptionSpecification} depends on the output of \textsc{OptionLearning} from a previous iteration: the initiation set learned for the most recent option becomes the subgoal for the next option in the chain. This creates a tight interleaving between specification and learning that contrasts with bottleneck methods (Section~\ref{sec:od/graph-based}), where \textsc{OptionSpecification} can run as a single preprocessing phase. In skill chaining, each iteration of the outer loop in Algorithm~\ref{alg:hrl_generic} adds at most one new option to the chain, and the loop terminates when the start state is covered by some option's initiation set.

\begin{figure}[h!]
    \centering
    \includegraphics[width=\linewidth]{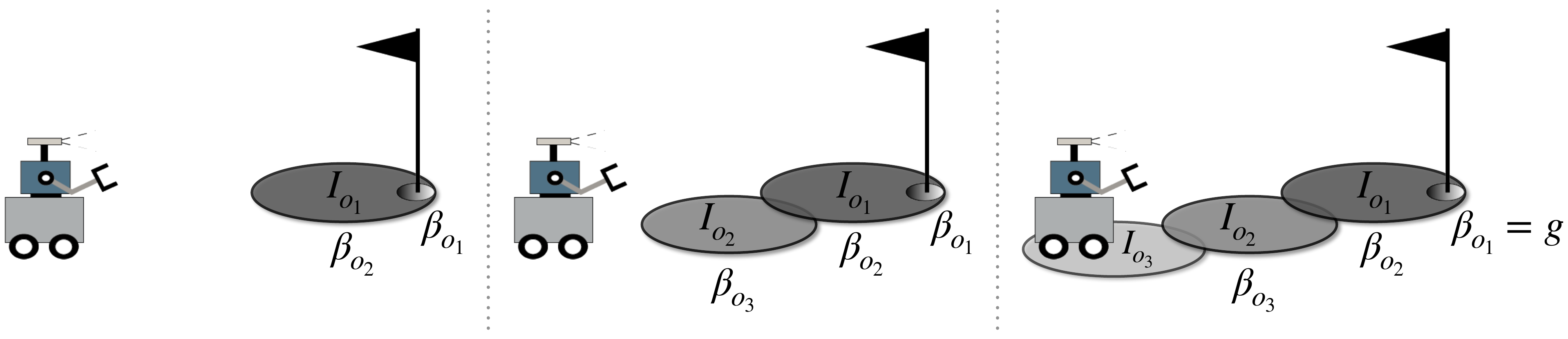}
    \caption{\textbf{Sequentially Composable Options.} The skill chaining algorithm incrementally learns options backwards from the goal, such that the subgoal of each option is the initiation region of another option. First the agent finds the states from which it can reliably reach the goal \textit{(left)}, then it finds the states from where it can reach the first region \textit{(middle)}, and so on, until there is a high probability of success from the environment's start state \textit{(right)}.}
    \label{fig:skill-chaining-illustration}
\end{figure}

Figure~\ref{fig:skill-chaining-illustration} illustrates the algorithm. Given a target region of states $g\subset \mathscr{S}$ (for example, the task goal, shown as a flag in Figure~\ref{fig:skill-chaining-illustration}), skill chaining discovers subgoal options that can be sequenced together so that each option execution roughly brings the agent closer to $g$. This is done by learning options backward from the goal: first, the agent learns option $o_1$ with termination condition $\beta_{o_1}(s)=\mathbbm{1}(s \in g)$ and subgoal reward $r^{o_1}(s)=\mathbbm{1}(s \in g)$. Option learning entails training two functions: (a) the option policy $\pi(\cdot|s,o_1)$, which aims to maximize $r^{o_1}$, and (b) the initiation function $\mathscr{I}(s,o_1)$, which represents the probability that $\pi(\cdot|s,o_1)$ can reliably reach $g$ from state $s$. Shortly after, the agent creates another option $o_2$ whose subgoal is the initiation region of the previous option---this is because the agent can reach the goal with high probability from states inside the first option's initiation region. This process continues until the start state $s_0\sim\rho_0$, where $\rho_0$ denotes the environment's initial state distribution, is inside the initiation region of some option, at which point the agent can simply execute its learned options in sequence to achieve the goal $g$. Sequential composability is explicitly enforced by setting the subgoal of each new option $o_i$ to be the states in which the previous option $o_{i-1}$ has a high initiation probability, i.e., $\mathscr{I}(s,o_{i-1})$ is greater than some pre-specified threshold $c \in [0,1]$.

In skill chaining, the initiation set of an option has special meaning: it represents the states from which option execution is likely to achieve its subgoal. Learning the initiation function is usually framed as a binary classification problem: states along successful option trajectories (those that achieve the option's subgoal) are considered as positive examples $\mathbf{s}^+=\{s^+_1, \cdots, s^+_n\}$ and states along unsuccessful trajectories are considered as negative examples $\mathbf{s}^-=\{s^-_1, \cdots, s^-_m\}$. Then, a probabilistic classifier (with parameters $\chi$) is fit on these training examples using the binary cross-entropy loss. Now, when a new state $s$ is encountered during learning, $\mathscr{I}(s,o)$ represents the probability that the agent can reach option $o$'s subgoal in a single execution of $\pi(\cdot|s,o)$. While this classification approach is simple to implement, some of its drawbacks include: (a) the classifier struggles to adapt to changing option policies, and (b) the agent must wait until the end of option execution to update its initiation function. To address this issue, \citet{bagaria2023effectively} frame the initiation function as a general value function \citep{sutton2011horde}: the agent uses each experience tuple $(s, a, r, s')$ to update its prediction of whether an option execution will achieve its subgoal. Letting $g_o(s)$ denote the option's subgoal function---a binary function indicating whether state $s$ satisfies option $o$'s subgoal---this is done using the following temporal difference (TD) error and stochastic gradient descent update rule:
\begin{align}
    \delta_{\mathscr{I}}(s,o) &= g_o(s') + \big(1 - g_o(s')\big)\,\mathscr{I}_\chi(s',o) - \mathscr{I}_\chi(s,o),\\
    \chi &= \chi - \alpha \delta_{\mathscr{I}} \nabla_\chi \mathscr{I}_\chi(s,o),
\end{align}
where $\alpha\in\mathbb{R}^+$ is a step size parameter and $\chi$ are the initiation function parameters. However, at a given state $s$, an option's initiation probability $\mathscr{I}_{\chi}(s, o)$ can be low either because the option policy is unlikely to successfully reach its subgoal from state $s$ or because the agent does not have enough data to confidently estimate $\mathscr{I}_{\chi}(s, o)$. As a result, the skill chaining agent additionally estimates its uncertainty $\mathcal{U}(s, o)$ about its initiation function's predictions: when deciding whether an option is executable from a state $s$, it is optimistic with respect to that uncertainty, but when targeting another option's initiation region, it is pessimistic with respect to it \citep{bagaria2021robustly}.

\subsubsection{Relationship to the Algorithmic Template}

\begin{algorithm}[h!]
\caption{Skill Chaining: Instantiation of Algorithm~\ref{alg:hrl_generic}.}
\label{alg:skillchaining}
\begin{algorithmic}[1]
\Require Target region $g \subset \mathscr{S}$, option horizon $H_o$,
         initiation thresholds $c_1, c_2, c \in [0,1]$.
\State Initialize $\mathscr{O} \leftarrow \{o_1\}$ with
       $g_{o_1}(s) \leftarrow \mathbbm{1}(s \in g)$,
       $\mathscr{D} \leftarrow \varnothing$.
\Statex
\Procedure{Interact}{$E, \mathscr{O}, \mu, \mathscr{D}$}
    \State Sample option $o \in \mathscr{O}$ from:
           $$\frac{\mu(o|s)\mathscr{I}^+(s,o)}
                 {\sum_{o'\in\mathscr{O}} 
                  \mu(o'|s)\mathscr{I}^+(s,o')},$$
    \Statex \hspace{\algorithmicindent} where
           $\mathscr{I}^+(s,o) = 
            \texttt{clip}(\mathscr{I}(s,o) + 
            \mathcal{U}(s,o),\; 0,\; 1)$.
    \While{option $o$ does not terminate}
        \State Execute $a \sim \pi(\cdot|s,o)$,
               observe $r, s'$; append to $\mathscr{D}$.
    \EndWhile
    \State \Return $\mathscr{D}$
\EndProcedure
\Statex
\Procedure{OptionSpecification}{$\mathscr{O}, \mathscr{D}$}
    \If{$\mathbb{E}_{s_0 \sim \rho_0}[\mathscr{I}(s_0,o)] < c_1
        \;\&\; \mathbb{E}_{s_0 \sim \rho_0}[\mathcal{U}(s_0,o)] < c_2,
        \;\forall o \in \mathscr{O}$}
        \State Let $o_k$ be the last option added to the chain.
        \State $g_{o'}(s) \leftarrow 
               \mathbbm{1}(\mathscr{I}(s, o_k) > c)$
               \Comment{New subgoal function}
        \State $r^{o'}(s) \leftarrow g_{o'}(s)$
               \Comment{New option reward}
        \State \Return $\Omega = \{(g_{o'}, r^{o'})\}$
        \Else
            \State \Return $\Omega = \varnothing$
    \EndIf
\EndProcedure
\Statex
\Procedure{OptionLearning}{$\mathscr{O}, \mathscr{D}, \Omega$}
    \If{$\Omega \neq \varnothing$}
        \State Create new option $o'$ from $\Omega$; \;
               $\mathscr{O} \leftarrow \mathscr{O} \cup \{o'\}$.
    \EndIf
    \State For all $o \in \mathscr{O}$, set
           $\beta_o(s) \leftarrow g_o(s) + 
           (1 - g_o(s)) \cdot \tfrac{1}{H_o}$.
           \Comment{Example termination condition}
    \State Sample transitions $(s, a, r, s', o)$ 
           from $\mathscr{D}$.
    \State Update option policy $\pi(\cdot|s,o)$ 
           to maximize $r^o$.
    \State Update initiation via TD using $g_o$:
           $\delta_{\mathscr{I}}(s,o) = 
            g_o(s') + (1 - g_o(s'))\,\mathscr{I}(s',o) - \mathscr{I}(s,o)$.
    \State Update uncertainty estimate $\mathcal{U}(s,o)$.
            \Comment{E.g., using count-based methods}
    \State \Return $\mathscr{O}$
\EndProcedure
\end{algorithmic}
\end{algorithm}

Algorithm~\ref{alg:skillchaining} presents skill chaining as an
instantiation of Algorithm~\ref{alg:hrl_generic}. \textsc{Interact} is
defined explicitly because the high-level policy~$\mu$ is modulated by 
each option's initiation probability, so that options are only selected 
in states where they are likely to succeed. \textsc{OptionLearning} 
updates all existing options in~$\mathscr{O}$: for each option, it sets 
the termination condition from the subgoal function~$g_o$, trains the 
option policy against~$r^o$, and learns the initiation function via 
temporal difference updates whose cumulant is the subgoal 
function~$g_o$---this is how the learned initiation function remains 
grounded in the option's subgoal.\footnote{In continuous state spaces, 
the hard threshold in the subgoal function 
$g_{o'}(s) = \mathbbm{1}(\mathscr{I}(s, o_k) > c)$ is replaced by a 
soft sigmoid $\sigma(\mathscr{I}(s, o_k) - c)$; we use the indicator 
for clarity.} When \textsc{OptionSpecification} returns a non-empty 
$\Omega$, \textsc{OptionLearning} also creates the corresponding new 
option and adds it to~$\mathscr{O}$. Unlike bottleneck methods 
where~$\mathscr{I}$ is fixed by definition, here the initiation 
function is genuinely \textit{learned}. \textsc{PolicyImprovement} 
updates~$\mu$ using extrinsic reward; since this is standard, it is 
omitted from Algorithm~\ref{alg:skillchaining}. Finally, 
\textsc{OptionSpecification} checks whether the current chain covers 
the start state; if not, it returns a new option objective whose 
subgoal function is derived from the initiation set of the last option 
in the chain. Note that this decision depends on the initiation 
functions learned by \textsc{OptionLearning} in previous iterations, 
creating a feedback loop between the two subroutines across 
iterations of Algorithm~\ref{alg:hrl_generic}.

\subsubsection{Benefits and Opportunities}

\paragraph{Planning.} Each option execution drives the agent to a small, predictable region of the state space. Since those states are constructed to be inside the initiation region of another option, they can be sequentially composed. In practice, each option's initiation and termination region is parameterized using probabilistic classifiers, so there is a \textit{probability} that two options can be executed in sequence, which eventually permits computation of the probabilistic feasibility of entire plans \citep{Konidaris2018FromST}. \citet{bagaria2023effectively} used graph-search to find recursively optimal solutions and \citet{bagaria2021robustly} provided a dynamic programming algorithm to approximate hierarchically optimal ways of planning with subgoal options.

\paragraph{Credit Assignment.} Skill chaining has demonstrated more sample-efficient credit assignment in goal-reaching tasks than non-hierarchical RL, which can be attributed to the following reasons. (1) \textit{Jumpy transitions}: Skill chaining methods usually use the entire $T$-step option transition $(s_t, o, \sum r_{t:t+T}, s_{t+T})$ to update the high-level policy $\mu(o|s)$. Much like $n$-step returns and TD($\lambda$) in non-hierarchical RL, this has the effect of rapidly propagating credit among state-action pairs. (2) \textit{Focused next-state distribution}: not only does each option execute for multiple timesteps, but it also guides the agent to states that are closer to the goal. In other words, options in the skill chain modify the agent's state distribution to make states closer to the goal more likely. Since these states are usually the ones with non-zero values, bootstrapping-based value learning (e.g., TD) progresses more rapidly.

\paragraph{Exploration.} Since skill discovery proceeds backward from the goal, the algorithm requires either an exploration policy or a set of demonstration trajectories \citep{konidaris2010constructing,kang2022deep} that achieve the task goal. This advocates for a view of skill chaining as producing options that are good for exploitation, which can be combined with options that are good for exploration. Deep skill graphs (DSG) \citep{bagaria2021skill,bagaria2025imdsg} overcome this limitation: the agent finds intrinsically motivating states and learns skill chains that connect them to each other; the resulting chains form a graph abstraction of the environment, which is useful for planning. Furthermore, the graph building process has a \textit{Voronoi bias} \citep{LaValle1998RapidlyexploringRT,lindemann2004incrementally}, meaning that it tends to grow towards parts of the state space where the agent has the least experience.

\paragraph{Opportunities for Research.}
\begin{itemize}
    \item \textit{Goal-reaching options.} To learn the initiation set of each option in the chain, its subgoal must be described using a binary function: either the subgoal is achieved in the current state, or it is not. Such a subgoal description is not universal, as it cannot be naturally used to describe continuing tasks like maximizing velocity or repeating periodic motions. If the initiation cumulant \citep{bagaria2023effectively} can be formulated for general reward functions, then skill chaining can be applied to non-goal-reaching tasks as well.
    \item \textit{Controlling all state variables at the same time.} If we think of states being composed of different state variables, which is a property known as \textit{factoredness} \citep{boutilier2000factored}, then skill chaining drives the value of all variables to a certain range of values. In more complex environments, it may be unnecessary, or even impossible, to control all state variables at the same time. Future work could create a version of skill chaining that leverages the factoredness of the state space and only controls a subset of all the factors at any given time.
\end{itemize}

\paragraph{Additional connections to control theory and motion planning.} \citet{lozano1984automatic}, \citet{mason1985mechanics}, and \citet{burridge99sequential} popularized the view of policies as \textit{funnels}: these policies drive a large set of ordinary states to a small set of desired states. Policies can be sequentially composed to reach some target set of states by placing the end (narrow part) of each funnel inside the beginning (broad part) of some other funnel. \citet{tedrake2010lqr,ames2019bounded} provided a way to compute these initiation regions for complex, dynamical systems using convex optimization and built robust controllers for fixed-wing UAVs \citep{tedrake2010lqr}. Meanwhile, \citet{konidaris2009skill} applied this idea to model-free RL; the broad part of the funnel corresponds to an option's initiation set, while the narrow part corresponds to the option's subgoal. \citet{bagaria2020option} then upgraded the skill chaining algorithm with deep learning so that it could be applied to higher-dimensional systems. Variants of deep skill chaining have been used in robotic surgery \citep{huang2023value}, manipulation \citep{lee2021adversarial,vats2023efficient}, multi-agent RL \citep{xie2022deep}, and task and motion planning \citep{mishra2023generative}.

\subsection{Empowerment Maximization}
\label{sec:empowerment_skill_discovery}

Empowerment-based methods discover diverse skills by maximizing an agent's control over its environment. At its core, empowerment quantifies how much influence an agent has over its future observations---an agent is more empowered when it can reliably cause a wider variety of outcomes \citep{klyubin2005empowered,salge2014empowerment}. For example, having access to a car empowers you to reach many different locations; learning to swim empowers you to survive in water, and so on. Empowerment can also be seen as a way to maximize social influence in multi-agent settings \citep{jaques2019social}, or to seek agreement between future states and the agent's internal representations \citep{hafner2020apd}.

Formally, empowerment is defined as the mutual information between an agent's actions and its future states. Mutual information $I(X; Y) = \sum_{x,y} p(x,y) \log \frac{p(x,y)}{p(x)p(y)}$ measures how much information one random variable provides about another, equaling zero when the variables are independent and increasing as they become more statistically dependent. 

\begin{figure}
    \centering
    \includegraphics[width=0.9\linewidth]{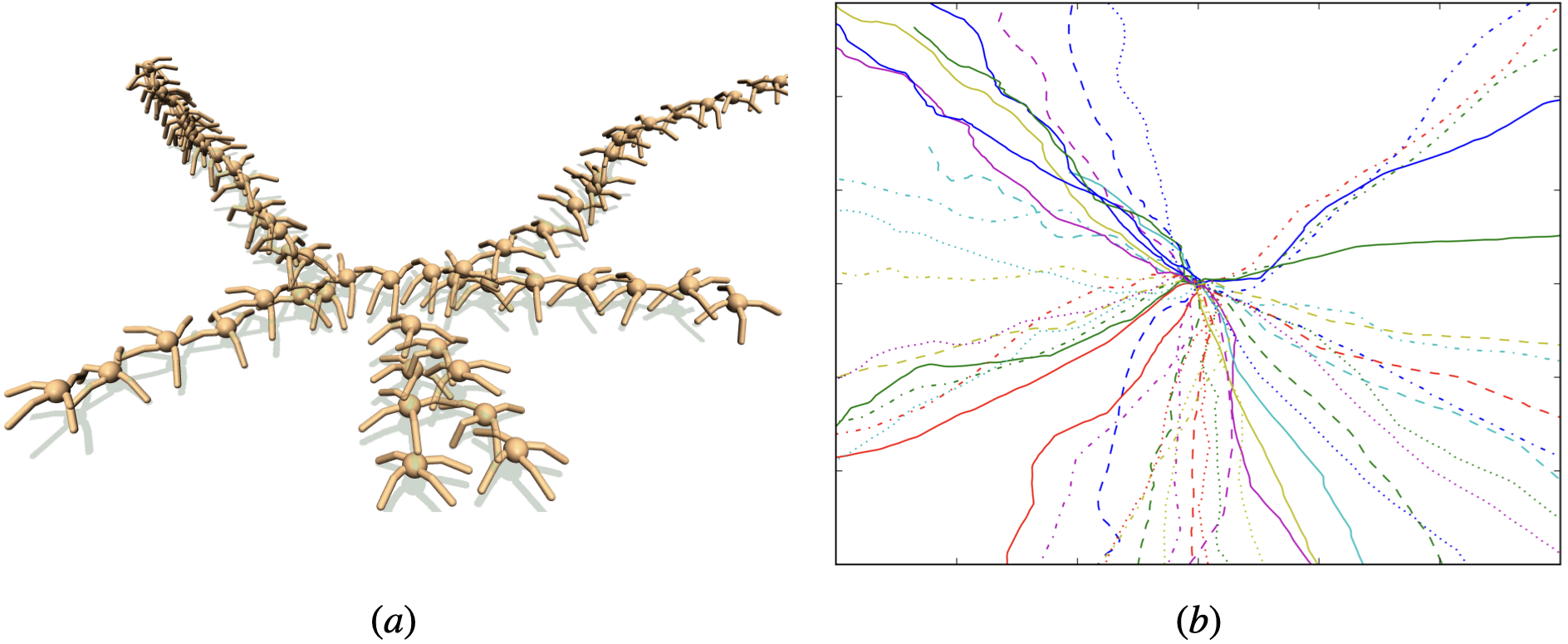}
    \caption{Empowerment-based skill-discovery methods learn skills that generate trajectories that are maximally different from one another, with the constraint that, having observed a trajectory, it should be clear which skill generated it. $(a)$ Trajectories generated by $6$ distinct skills in \textsc{MuJoCo Ant}; $(b)$ $(x, y)$ location of the center of mass of the \textsc{Ant} plotted after executing skills learned by the DADS algorithm. Figure from \citet{sharma2020dads}, used with permission.}
    \label{fig:empowerment-illustration-dads}
\end{figure}

Now, consider an agent that executes a sequence of $n$ actions $\mathbf{a} = (a_t, a_{t+1}, \ldots, a_{t+n-1})$ starting from state $s_t$, resulting in state $s_{t+n}$. The $n$-step empowerment at state $s_t$ is:
\begin{align}
\mathcal{E}_n(s_t) = \max_{p(\mathbf{a})} I(\mathbf{a}; s_{t+n} | s_t),
\label{eq:empowerment_definition}
\end{align}
where $p(\mathbf{a})$ is the probability distribution over action sequences that the agent can choose. This captures the maximum amount of information that action sequences can provide about future states, optimized over all possible action distributions $p(\mathbf{a})$. Expanding the mutual information reveals its intuitive meaning:
\begin{align}
I(\mathbf{a}; s_{t+n} | s_t) &= \mathcal{H}(s_{t+n} | s_t) - \mathcal{H}(s_{t+n} | s_t, \mathbf{a}),
\label{eq:empowerment_expanded}
\end{align}
where the first term represents uncertainty about future states given only the current state, while the second represents remaining uncertainty after choosing actions. Empowerment measures how much this uncertainty can be reduced through deliberate action choice.

As the mutual information is intractable, \citet{mohamed2015variational} propose to estimate it through variational inference. Specifically, the authors estimate $p(\mathbf{a} | s_{t+n})$ using the variational approximation $q_{\phi}(\mathbf{a}|s_{t+n})$ and leverage the non-negativity of the KL divergence to obtain:\footnote{Note that we are departing from our usual notation to denote $p$ and $q$ using the conventional notation in variational inference.}
\begin{align}
I(\mathbf{a}; s_{t+n} | s_t) &= \mathcal{H}(\mathbf{a} | s_t) - \mathcal{H}(\mathbf{a} | s_t, s_{t+n}) \\ 
&= \mathcal{H}(\mathbf{a} | s_t) + \mathbb{E}[\log p(\mathbf{a}|s_t, s_{t+n})]\\
&\ge \mathcal{H}(\mathbf{a}|s_t) + \mathbb{E}[\log q_{\phi}(\mathbf{a}|s_t, s_{t+n})] \quad \text{(Variational Bound).}\label{eq:empowerment_md_rezende}
\end{align}
This variational bound (Equation~\ref{eq:empowerment_md_rezende}) provides a practical way to compute empowerment in high-dimensional continuous spaces using neural networks (with parameters $\phi$). However, this objective finds open-loop action sequences $\mathbf{a}$, and we want to discover \textit{skills} (closed-loop policies). \citet{gregor2017variational} addressed this concern by introducing Variational Intrinsic Control (VIC), which replaces fixed action sequences with parameterized skills $\pi_\theta(a|s,z)$ conditioned on skill variables $z$. In the language of options,  $z$ is the option identifier (see Definition~\ref{def:option_id}), and each $z$ specifies a distinct option whose policy is $\pi_\theta(\cdot|s,z)$. VIC maximizes the mutual information between skills and final states reached from skill execution:
\begin{align}
J_{\text{VIC}} = I(z; s_{t+n} | s_t),
\label{eq:vic_objective}
\end{align}
where $s_t$ is the initial state and $s_{t+n}$ is the final state after executing skill $z$ for $n$ timesteps. We can expand this as:
\begin{align}
I(z; s_{t+n} | s_t) = \mathcal{H}(z | s_t) - \mathcal{H}(z | s_{t+n}, s_t).
\end{align}
Similar to \citet{mohamed2015variational}, VIC uses the variational lower bound:
\begin{align}
I(z; s_{t+n} | s_t) &\geq \mathcal{H}(z | s_t) + \mathbb{E} [\log q_\phi(z | s_{t+n}, s_t)].
\label{eq:vic_bound}
\end{align}
This variational lower bound can be optimized by training two neural networks: a policy $\pi_\theta(a|s,z)$ that executes skills, and a discriminator $q_\phi(z|s_{t+n}, s_t)$ that predicts which skill was used based on the final state.

\citet{eysenbach2019diversity} simplified VIC's approach in their method \textit{Diversity is All You Need} (DIAYN). While VIC maximizes mutual information between skills and final states, DIAYN instead focuses on making skills distinguishable from the states they visit throughout execution---note the shift from trajectory endpoints to all visited states. DIAYN builds on maximum entropy RL, which augments the standard RL objective with an entropy bonus $\mathcal{H}(A|S)$ to encourage exploration. DIAYN learns skills by maximizing:
\begin{align}
J_{\text{DIAYN}} = I(s; z) + \mathcal{H}(a | s) - I(a; z | s),
\label{eq:diayn_objective}
\end{align}
which has an intuitive interpretation: skills should be distinguishable from the states they visit ($I(s; z)$), actions should be diverse ($\mathcal{H}(a | s)$), 
but skills should differ primarily through state visitation rather than action choices ($-I(a; z | s)$).
This objective can further be simplified by expanding the mutual information in terms of the conditional entropies, and then applying the variational approximation similar to \citet{mohamed2015variational}:
\begin{align}
    J_{\text{DIAYN}} &= \Big(\mathcal{H}(z) - \mathcal{H}(z|s)\Big)
    + \mathcal{H}(a|s) - \Big(\mathcal{H}(a|s) - \mathcal{H}(a|s,z)\Big)\\
    &= \mathcal{H}(z) - \mathcal{H}(z|s) + \mathcal{H}(a|s,z)\\
    &= \mathcal{H}(a|s,z) + \mathbb{E}[\log p(z|s)] - \mathbb{E}[\log p(z)]\\
    &\geq \mathcal{H}(a|s,z) + \mathbb{E}[\log q_\phi(z|s)] - \mathbb{E}[\log p(z)].
\end{align}

The final step implies the use of a discriminator $q_\phi(z|s)$ to variationally approximate $p(z|s)$; the result is an intra-skill pseudo reward function for each skill $z$:
\begin{equation}\label{eq:diayn_reward}
r^z(s) = \log q_\phi(z|s) - \log p(z),
\end{equation}
assuming that the $\mathcal{H}(a|s,z)$ term is maximized using a maximum entropy RL formulation \citep{Haarnoja2017ReinforcementLW}. This leads to a practical algorithm: sample skill $z \sim p(z)$, execute policy $\pi_\theta(a|s,z)$, train discriminator $q_\phi(z|s)$ to predict skills from states, and update the policy using pseudo-reward in Equation~\ref{eq:diayn_reward}.

While DIAYN successfully learns diverse behaviours, \citet{sharma2020dads} observed that it can discover skills with unpredictable effects, making them difficult to sequentially compose downstream. Their method, \textit{Dynamics-Aware Discovery of Skills} (DADS), addresses this by explicitly encouraging predictable skill dynamics while maintaining diversity. Their formulation captures two desirable properties simultaneously: different skills should lead to different future states (diversity), and given the current state and skill, the future state should be predictable. Expanding the mutual information:
\begin{align}
I(z; s_{t+n} | s_t) &= \mathcal{H}(s_{t+n} | s_t) - \mathcal{H}(s_{t+n} | s_t, z) = \mathbb{E} \left[ \log \frac{p(s_{t+n}|s_t,z)}{p(s_{t+n}|s_t)} \right].
\end{align}
DADS uses variational approximation with two learned models: a skill dynamics model $q_\psi(s_{t+n}|s_t,z)$ and a marginal dynamics model $q_\xi(s_{t+n}|s_t)$. The skill reward function becomes
\begin{equation}
r^z(s_t,s_{t+n}) = \log q_\psi(s_{t+n}|s_t,z) - \log q_\xi(s_{t+n}|s_t),    
\end{equation}
encouraging skills that make state transitions more predictable than the marginal dynamics. Resulting skills have focused effects, and examples of learned skills are visualized in Figure~\ref{fig:empowerment-illustration-dads}. These methods share the key insight that useful skills should be distinguishable from their effects on the environment---whether through final states (VIC), visited states (DIAYN), or state transitions (DADS), each approach learns discriminators that identify which skill was executed from observations. This creates an intrinsic reward signal driving the discovery of diverse, meaningful behaviours without external supervision.

In terms of option components, most empowerment methods use universal initiation ($\mathscr{I}(s) = 1$ for all skills) and fixed-horizon termination: each skill executes for exactly $n$ steps, after which a new skill is sampled. The skill variable $z$ is sampled once at initiation and held fixed throughout execution, so the conditioned policy $\pi_\theta(a|s,z)$ implements a single closed-loop option for each value of $z$.

Several extensions address limitations of the basic DIAYN framework. One line of work improves the discriminator: VALOR \citep{achiam2018variational} replaces DIAYN's per-timestep decomposition $\log q_\phi(z|\tau) = \sum_{i} \log q_\phi(z|s_{t+i})$ with an LSTM architecture that captures temporal dependencies across the full trajectory, while DISDAIN \citep{strouse2022disdain} augments the discriminator with a novelty bonus to counteract its pessimism in unfamiliar states. A second line improves the optimization of the mutual information objective itself; for example, CIC \citep{laskin2022cic} uses contrastive learning for tighter estimation. A third line revisits \textit{what} skills should be distinguishable by; for example, Relative VIC \citep{baumli2021relative} rewards skills for causing characteristic \textit{changes} in state rather than visiting specific states, encouraging transferable behaviours that are defined by their effects rather than their specific endpoints.

\subsubsection{Relationship to the Algorithmic Template}

\begin{algorithm}[t]
\caption{Empowerment-Based Skill Discovery: Instantiation of 
Algorithm~\ref{alg:hrl_generic}.}
\label{alg:empowerment}
\begin{algorithmic}[1]
\Require Skill distribution $p(z)$ (e.g., uniform), skill horizon $H_o$.
\State Initialize policy $\pi_\theta$, discriminator $q_\phi$, 
       dataset $\mathscr{D} \leftarrow \varnothing$.
\Statex
\Procedure{Interact}{$E, \mathscr{O}, \mu, \mathscr{D}$}
    \State Sample skill $z \sim \mu$ where $\mu = p(z)$.
    \State Execute $\pi_\theta(\cdot|s,z)$ for $H_o$ steps; 
           append $(s_t, a_t, r_t, s_{t+1}, z)$ tuples 
           to $\mathscr{D}$.
    \State \Return $\mathscr{D}$
\EndProcedure
\Statex
\Procedure{OptionSpecification}{$\mathscr{O}, \mathscr{D}$}
    \State Update $q_\phi$ to predict $z$ from labeled 
           transitions $(s, a, r, s', z)$ sampled 
           from $\mathscr{D}$.
    \State \Return $\Omega = q_\phi$
    \Comment{Implicitly defines $r^z$ for all $z$}
\EndProcedure
\Statex
\Procedure{OptionLearning}{$\mathscr{O}, \mathscr{D}, \Omega$}
    \State Compute $r^z$ from $\Omega = q_\phi$ using Eq.~\ref{eq:diayn_reward}.
    \State Update $\pi_\theta(\cdot|s,z)$ to maximize $r^z$ 
           using transitions sampled from $\mathscr{D}$.\\
           \Comment{$\pi_\theta$ implicitly updates all option policies in $\mathscr{O}$}
    \State All options have $\mathscr{I}(s,o)=1, \forall s \in \mathscr{S}$.
    \State All options terminate after $n$ steps or $\beta(s,o)=1/H_o$.
    \State \Return $\mathscr{O}$
\EndProcedure
\end{algorithmic}
\end{algorithm}

Algorithm~\ref{alg:empowerment} highlights the shared structure across 
empowerment methods. \textsc{Interact} samples a skill~$z \sim p(z)$ 
and stores labeled transitions~$(s, a, r, s', z)$ 
in~$\mathscr{D}$, so that subsequent subroutines can identify which 
skill generated each transition. As in previous sections, the primary 
variation lies in how \textsc{OptionSpecification} defines option 
objectives. Here, rather than identifying discrete subgoals, 
\textsc{OptionSpecification} trains a discriminator~$q_\phi$ to 
predict which skill~$z$ was executed from observed transitions; the 
trained discriminator is itself the option objective 
($\Omega = q_\phi$), from which the option reward~$r^z$ can be 
derived for any skill on demand 
(Equations~\ref{eq:vic_bound}--\ref{eq:diayn_reward}). A major 
component that varies across VIC, DIAYN, and DADS is what the 
discriminator conditions on: final states, visited states, or state 
transitions, respectively. The option reward is not an indicator 
function over states but a learned pseudo-reward that evolves during 
training---a form of non-stationarity shared with other methods such 
as skill chaining (Section~\ref{sec:skillchaining}) and spectral 
methods (Section~\ref{sec:od/spectral}), though arising here from 
the co-training of~$q_\phi$ with the policy. 
\textsc{OptionLearning} trains a single policy~$\pi_\theta(a|s,z)$ 
that implements all options simultaneously, parameterized by~$z$---this 
contrasts with methods like bottleneck discovery where each option 
typically has a separate policy. Since updating~$\pi_\theta$ 
simultaneously updates the behaviour of every option indexed by~$z$, 
the returned policy implicitly represents the entire option 
set~$\mathscr{O}$. Both the discriminator and the policy are updated 
using labeled transitions sampled from the shared 
dataset~$\mathscr{D}$, which grows as the agent interacts with the 
environment.

The phasing of operations is also distinctive. 
\textsc{OptionSpecification} and \textsc{OptionLearning} are tightly 
interleaved, with both the discriminator and the policy updated at 
every iteration. \textsc{PolicyImprovement}, by contrast, is often 
deferred to a separate downstream phase in which a high-level 
policy~$\mu$ selects among the discovered skills for a specific task. 
Some methods partially bridge this gap---for example, DADS learns a 
dynamics model during skill discovery that can be used for planning at 
test time~\citep{sharma2020dads}---but a separation between skill 
discovery and skill use is common in this family.

\subsubsection{Benefits and Opportunities}
\paragraph{Exploration.} Empowerment serves as an intrinsic motivation that encourages agents to seek states where they have the most control over future outcomes \citep{klyubin2005empowered}. Several recent empowerment methods seek to address the exploration problem in RL by explicitly optimizing for skill diversity \citep{eysenbach2019diversity} and state-space coverage \citep{campos2020edl}; by learning a diverse set of skills, agents not only explore effectively in a single-task setting \citep{Massari2021ExperimentalET}, but also adapt quickly to new tasks, reducing sample complexity \citep{sharma2020dads,baumli2021relative,hansen2020fast}. One difficulty in obtaining better exploration with empowerment-based methods is that the learned skills tend to be localized, i.e., only cover a small area. Lipschitz-constrained Skill Discovery (LSD) \citep{park2022lipschitzconstrained} replaces mutual information estimation with a Lipschitz-constrained objective, ensuring that learned skills correspond to large, meaningful state transitions rather than minor variations. Exploration is further enhanced by paying attention to the parts of the environment that are within the agent's control \citep{csd}---this is done by using a controllability-aware distance function that assigns higher values to harder-to-control state transitions, leading to more complex skill acquisition, such as object manipulation, without direct external supervision.

\paragraph{Credit Assignment.} While the primary motivation of empowerment-driven techniques is that of exploration, recent methods seek to improve the process of learning policies over discovered skills. For example, \citet{leibfried2019unified} derive Bellman operators that combine empowerment- and reward-maximization; \citet{sharma2020dads,sharma2020emergent} advocate for learning option models during the skill-discovery phase so that the learned options can be composed via a planner at test-time. In fact, the simplification of empowerment-based objectives to state-reaching \citep{pitis2020mega} can be viewed as a compromise between learning more expressive skills and creating stationary objectives that ease online policy learning for utilizing discovered skills.

\paragraph{Transfer.} Most empowerment-based skill-discovery algorithms learn skills that are distinguishable via specific states encountered in sampled trajectories (for example, the last state of the trajectory). This approach leads to skills that are tied to specific states encountered during skill-learning; in other words, learned skills do not transfer to unseen, related parts of the state space. Relative VIC is a promising approach for learning transferable skills because it rewards skill policies for causing characteristic \textit{changes} in state, rather than targeting specific states themselves. Some algorithms use successor features to enable transfer \citep{zahavy2021discovering}, but more research is needed on learning skills that simultaneously maximize empowerment and enable reuse across different portions of the state space.

\paragraph{Opportunities for Research.}

\begin{itemize}
    \item \textit{Optimization challenges.} Despite significant progress in online mutual information estimation, empowerment remains challenging to estimate and optimize \citep{achiam2018variational}. 
    To address this, \citet{park2023metra} introduce a Wasserstein variant of the mutual information objective, where the KL divergence in MI is replaced with the Wasserstein distance. Finding such ways to ease the estimation and optimization of the empowerment objective is a key area of current research. 
    
    \item \textit{Connections to causal learning.} \citet{pittphilsci23268} hypothesizes that if an agent learns an accurate causal model of the world, it will necessarily increase its empowerment, and, conversely, increasing empowerment will lead to a more accurate (albeit implicit) causal model of the world \citep{salge2014empowerment}. This could enable model-based planning for complex, long-horizon problems, fully unleashing the power of HRL.\looseness=-1
    \item \textit{Possible signatures in human learning.} There is mounting evidence in developmental cognitive science that the drive to learn causal models of the world is behind many of the exploratory capabilities of children \citep{gopnik2012reconstructing}. For example, \citet{rovee1979economics} show that infants as young as $3$ months old vary their actions to observe their causal effects on their environment; \citet{du2023what} show that children playing some video games can be thought of as maximizing their empowerment. \citet{pittphilsci23268} hypothesizes that empowerment maximization in RL could become the new dominant paradigm (after Bayesian approaches that struggle to scale to large hypothesis spaces) for explaining exploration in humans and other animals.
\end{itemize}

\paragraph{Connections to goal-based exploration.} The empowerment framework connects naturally to goal-conditioned RL, and understanding this connection clarifies the role of the variational distribution in shaping skill discovery. When the variational distribution in Equation~\ref{eq:vic_bound} is Gaussian and fixed, empowerment objectives reduce to goal-based exploration in RL \citep{choi2021variational}, by which we mean methods that propose random target states and use a goal-conditioned policy \citep{schaul2015universal} to reach them, for example, in hindsight experience replay \citep{Andrychowicz2017HindsightER} and Go-Explore \citep{Ecoffet2020FirstRT}. In fact, it is possible to think of goal-based exploration and variational empowerment as lying on a spectrum: the more expressive the variational distribution, the more powerful, albeit non-stationary, the associated representation learning problem \citep{choi2021variational}. Furthermore, \citet{farley2019discern} advocate for taking a mutual information maximization approach to goal-conditioned reward functions, \citet{pitis2020mega} argue that empowerment maximization is roughly equivalent to maximizing the size of the set of goals that can be achieved by the agent's policy, and \citet{levy2023hierarchicalempowermenttractableempowermentbased} find that goal-conditioning can make the empowerment objective significantly easier to compute and optimize. These findings further blur the lines between goal-based exploration in RL and empowerment maximization.

\subsection{Via Environment Rewards} \label{sec:learn_from_rewards}
Much of the work on learning options has focused on discovering intrinsic reward functions, which are then used to learn option policies. There are, however, two important lines of work that instead aim to learn behaviour directly through the rewards given by the environment. 

\subsubsection*{Feudal Methods}
The first set of approaches builds on \textit{feudal reinforcement learning }\citep{dayan1993feudal}. In this framework, the agent is decomposed hierarchically into managers and workers: managers set subgoals for workers to achieve, and workers use non-hierarchical RL to achieve those subgoals. In this way, goal-setting is decoupled from goal-achievement; each level in the hierarchy communicates to the level below
it what must be achieved, but does not specify how to do
so. The manager maximizes the reward coming from the environment to define the goals that the worker should achieve. Feudal RL was extended to deep RL through FeUdal Networks (FuNs) \citep{vezhnevets2017feudal}. FuN learns a two-level hierarchy in which the higher-level manager outputs a goal vector $g_t$ at time $t$ that specifies the direction in which the lower-level worker should modify the agent's current state. Specifically, a linear transformation, $\phi$, then maps the last $c$ goals outputted by the manager into an embedding vector $w_t$, 
\begin{equation}
\label{eq:fun-goal-emb}
w_t = \phi\big(\sum_{i=t-c}^t g_i \big).
\end{equation}
The worker's policy is then defined through this embedding vector and a matrix of learnable parameters $U_t$, that is $\pi^{\text{worker}} = \text{SoftMax}(U_tw_t)$.

The worker policy is trained through the standard policy gradient update rule, where the rewards are defined as,
\begin{equation}
\label{eq:fun-worker}
    r^{\text{worker}}(s_t,g_{t-i},s_{t-i}) = \frac{1}{c} \sum_{i=1}^c d_{cos} (s_t - s_{t-i},g_{t-i}),
\end{equation}
where $d_{cos}(x,y)$ is the cosine similarity measure between vector $x$ and $y$. The manager policy is learned with the task reward function; however, the authors propose the following update rule, which they term directional policy gradient:
\begin{equation}
    \nabla \mu(g_t|s_t) = \nabla d_{cos}(s_{t+c}-s_t,g_t) \big(Q^{\text{manager}}(s_t, g_t)-V^{\text{manager}}(s_t)\big).
\end{equation}
This update closely puts an emphasis on the direction in which a goal vector points to, and whether that direction was achieved by transitioning from $s_t$ to $s_{t+c}$.

As with most option discovery methods, the high-level policy is trained at the same time as the low-level policy; as the low-level policy changes during learning, data from a high-level action taken in the past may not yield the same low-level behaviour in the future \citep{nachum2018data}. This non-stationarity is addressed using relabeling tricks and off-policy learning in the HIRO algorithm \citep{nachum2018data}. In their work, as well as following literature, the worker's intrinsic reward is defined as,
\begin{equation}
\label{eq:hiro-worker}
    r^{\text{worker}}(s_t,g_t,s_{t+1}) = - ||s_t + g_t - s_{t+1}||_2.
\end{equation}
This definition forgoes the explicit use of the cosine similarity; however, it maintains the idea that a goal vector would represent a delta between state transitions. Later, \citet{levy2018learning} present the Hierarchical Actor-Critic (HAC) algorithm, which improves upon HIRO by removing the need for dense reward functions by instead using hindsight experience replay \citep{Andrychowicz2017HindsightER}. In a separate direction, \citet{hafner2022deep} instantiate the feudal architecture within a model-based algorithm called Director, which shows strong performance across a wide range of environments. Their approach additionally provides interpretability as the world model can decode goals into images.

\begin{algorithm}[t]
\caption{Feudal Methods: Instantiation of 
Algorithm~\ref{alg:hrl_generic}.}
\label{alg:feudal_inst}
\begin{algorithmic}[1]
\Procedure{OptionSpecification}{$\mathscr{O}, \mathscr{D}$}
    \State Collect additional goals $G$ from $\mathscr{D}$.
    \ForAll{$g \in G$}
        \State Define the option reward $r^{\text{worker}}(\cdot;\, g)$.
               \Comment{E.g., Eq.~\ref{eq:fun-worker} and~\ref{eq:hiro-worker}}.
        \State $\Omega \leftarrow \Omega \cup \{r^{\text{worker}}(\cdot;\, g)\}$
    \EndFor
    \State \Return $\Omega$
\EndProcedure
\Statex
\Procedure{OptionLearning}{$\mathscr{O}, \mathscr{D}, \Omega$}
    \State Update worker policy $\pi_{\theta}(a\mid s,g)$ to maximize $r^{\text{worker}}$.
    \State Construct options $\mathscr{O}$:
    \Statex \hspace{\algorithmicindent} $\pi_o(a\mid s)=\pi_{\theta}(a\mid s,g),\quad \forall s, \ \mathscr{I}(s)=1$
    \Statex\Comment{Goal-induced termination, or terminate after fixed $H_o$ timesteps}
    \State \Return $\mathscr{O}$
\EndProcedure
\end{algorithmic}
\end{algorithm}

\subsubsection*{Option-Critic}

The second set of approaches is based on the \textit{option-critic} \citep{Bacon2016TheOA}. In this work, the authors derive both the intra-option policy gradient theorem as well as the termination gradient theorem, which provide the update rules for learning option policies and termination functions, respectively. The intra-option policy gradient theorem leads to the following update,
\begin{align}
    \frac{\partial  q_{\pi}(s, o)}{\partial \theta} = \sum_{s,o} d^{\gamma}_{\pi,\mu,\beta}(s,o) \sum_a \frac{\partial  \pi_{\theta}(a| s, o)}{\partial \theta} q_{u}(s,o,a), 
    \label{intra_option}
\end{align}
where, $d^{\gamma}_{\pi,\mu,\beta}(s,o)=\sum_t \gamma^t P_{\pi,\mu,\beta}(S_t=s,O_t=o)$, is the $\gamma$-discounted occupancy measure over state-option pairs.  In the policy gradient theorem \citep{Sutton1999}, the gradient of the flat policy is weighted by the action-value function, leading to an increased probability for actions whose future discounted return is higher. In the case of the intra-option policy gradient, the quantity modulating the action probabilities is the state-action-option value function; therefore, a strict generalization of the policy gradient theorem. The termination gradient theorem used to learn the termination function is derived from the option value of option $o$ upon arrival in state $s$ (see Equation \ref{eq:arrival}). The update rule takes the following form,
\begin{align}
    \frac{\partial  u_{\beta}(s', o)}{\partial \psi} = \sum_{s,o} d^{\gamma}_{\pi,\mu,\beta}(s,o)  \frac{\partial  \beta_{\psi}(s', o)}{\partial \psi} A(s', o),
    \label{intra_option_2}
\end{align}
where $A(s', o) = q_{\pi}(s', o) - v_{\mu}(s')$ is the advantage function over options, representing how advantageous it is to be in state $s'$ with option $o$ with respect to the value of state $s'$ averaged over all options. Usually, the advantage function is implemented as a heuristic for reducing the variance of the estimator, but in this case, it comes naturally from the derivation of the theorem. Later, \citet{Bacon2018Temporal} unified these different objectives and derivation through the following objective,
\begin{align}
    J_\alpha(\omega) = \sum_{s,o} \alpha(s,o) Q_\omega(s,o) = \mathbb{E}_{\alpha,\omega} \left[ \sum_{t=0}^{\infty} \gamma^t r(S_t, A_t) \right],
\end{align}
where $\alpha:Dist( \mathscr{S} \times \mathscr{O})$ is a distribution of an initial state-option pair, and where $\omega$ defines all the parameters within the options framework, including the termination, option policies, and high-level policy. The authors show how, by assuming independence between the parameters of these components, the previous update rules can be recovered.

\begin{algorithm}[t]
\caption{Option-Critic: Instantiation of 
Algorithm~\ref{alg:hrl_generic}.}
\label{alg:oc_inst}
\begin{algorithmic}[1]
\Procedure{OptionSpecification}{$\mathscr{O}, \mathscr{D}$}
    \State Define the shared objective $\Omega$ as the environment reward $r$.
    \State \Return $\Omega$
    \Comment{No option-specific objectives}
\EndProcedure
\Statex
\Procedure{OptionLearning}{$\mathscr{O}, \mathscr{D}, \Omega$}
    \State Update intra-option policies $\pi_{\theta}(a\mid s,o)$ using Eq.~\ref{intra_option}.
    \State Update termination functions $\beta_{\psi}(s,o)$ using Eq.~\ref{intra_option_2}.
    \State Construct options $\mathscr{O}=\{(\mathscr{I},\pi_\theta,\beta_\psi)\}$ with $\forall s, \ \mathscr{I}(s)=1$.
    \State \Return $\mathscr{O}$
\EndProcedure
\end{algorithmic}
\end{algorithm}

The line of work surrounding the option-critic has received significantly more attention than we can present in detail in this section. Some of the contributions include learning safe policies \citep{Jain2018SafeOL},  using multiple discount factors \citep{Harutyunyan2019PerDecisionOD}, learning option termination in an off-policy manner \citep{harutyunyan2019termination}, extending the theorems to multiple levels of hierarchy \citep{Riemer2018LearningAO}, theoretical derivations that take parameter sharing between options into consideration \citep{Riemer2019OnTR}, and learning initiation functions~\citep{khetarpal2020options}. 

\subsubsection{Relationship to the Algorithmic Template}
Algorithm~\ref{alg:feudal_inst} is the common instantiation of Algorithm~\ref{alg:hrl_generic} for feudal methods. Feudal RL~\citep{dayan1993feudal} instantiates this template by defining a goal space and corresponding option reward in \textsc{OptionSpecification}, while \textsc{OptionLearning} trains a worker policy to achieve these goals; the manager operates as a high-level policy that selects goals based on environment reward. FuNs~\citep{vezhnevets2017feudal} follow the same structure, but parameterize goals as directions in a learned representation space and define $r^{\text{worker}}$ via cosine similarity (Equation~\ref{eq:fun-worker}), with the worker policy conditioned on an aggregated goal embedding (Equation~\ref{eq:fun-goal-emb}). HIRO~\citep{nachum2018data} retains this hierarchy with option reward in Equation~\ref{eq:hiro-worker}, and improves data efficiency via off-policy updates with correction in \textsc{PolicyImprovement}.
HAC~\citep{levy2018learning} adapts the template by replacing dense intrinsic rewards in FuN with sparse, goal-reaching rewards and leveraging hindsight relabeling.

Algorithm~\ref{alg:oc_inst} instantiates the vanilla option-critic algorithm~\citep{Bacon2016TheOA}. Different variants adopt distinct instantiations. 
For example, in line 7, \citet{harutyunyan2019termination} replace the standard advantage-based update with a termination critic that optimizes $\beta$ via
\begin{equation}
    \min \; \mathcal{H}(S_f \mid o), \quad S_f \sim P_\beta(\cdot \mid o),
\end{equation}
where $S_f$ denotes the random variable of the termination state of option $o$, and $H(\cdot)$ is the Shannon entropy.
For line 8, as another example, \citet{khetarpal2020options} replace the hard initiation set with a differentiable interest function $I_\chi(s,o)$, and update it according to
\begin{equation}
\chi \leftarrow \chi + \alpha\, \beta_\psi(s')\,
\frac{\partial \pi_{I_\chi}(o' \mid s')}{\partial \chi}\,
q(s',o'),
\end{equation}
where $\pi_{I_\chi}(o \mid s)$ is defined as
\begin{equation}
\pi_{I_\chi}(o \mid s)
=
\frac{I_\chi(s,o)\,\mu(o \mid s)}{\sum_{o'} I_\chi(s,o')\,\mu(o' \mid s)},
\end{equation}
so that interest is increased for options that, upon termination, are selected in states where they have high option-value.

\subsubsection{Benefits and Opportunities}
\paragraph{Credit assignment.} \citet{vezhnevets2017feudal} show strong performance of their feudal method on a set of Atari 2600 games from the Arcade Learning Environment \citep{Bellemare2012TheAL} and 3D navigation challenges \citep{Beattie2016DeepMindL}. These domains are long-horizon and require the agent to propagate credit across multiple steps. The authors report significantly better results than a baseline not leveraging such a hierarchy. 

\paragraph{Transfer.} \citet{Bacon2016TheOA} present experiments where the learned options improve the ability to generalize across changes in the Four rooms environment \citep{Sutton1999BetweenMA} compared to non-hierarchical RL algorithms. Such changes included modifying the goal location and the agent's starting location. This benefit is later reinforced by multiple works 
\citep{Zhang2019DACTD,khetarpal2020options,Kamat2020DiversityEnrichedO,Klissarov2021FlexibleOL} showcasing the transferability of options learned through the option-critic method in more complex environments such as locomotion control \citep{mujoco} and 3D navigation \citep{chevalier-boisvert2023minigrid}. In these transfer experiments, the agent usually first learns to perform a task before some component of the task is changed.

\paragraph{Interpretability.} 
A particular highlight of the option-critic line of work is that interpretability naturally emerges by learning options directly from environmental rewards.  For example, \citet{Bacon2016TheOA} report experiments where the termination function would highlight bottleneck states, which are often seen as key in learning temporal abstraction \citep{Stolle2002LearningOI}. 
Findings on interpretability are similarly reported across different domains \citep{ klissarov2017learnings,Harb2017WhenWI,Zhang2019DACTD}.

\paragraph{Opportunities for Research.}
\begin{itemize}
    \item \textit{Avoiding option degeneracy.} An important practical obstacle when learning options through the update rules proposed by the option-critic is that it may lead to degenerate solutions.
    Options tend to reduce to actions where each of the options' duration is only one timestep long. Another observed phenomenon is that only one option ends up being executed throughout all episodes. In both cases, the essence of temporal abstraction is lost. 
\citet{Bacon2016TheOA} originally noticed such behaviour and imputed it to the theoretical fact that using primitive actions is enough to solve any MDP. 
To avoid such undesirable behaviour, the authors add a penalty term $c_\text{delib}$ to the termination gradient's advantage function: $A(s', o) + c_\text{delib}$. This term essentially discourages the termination to prefer switching, unless the advantage in doing so is greater than the value of $c_\text{delib}$. A theoretical derivation was later done to justify the use of such a term, which was coined as the deliberation cost \citep{Harb2017WhenWI}. This cost is introduced as a hyperparameter, which raises the question of what value we should choose for a specific environment. Discovering more general solutions to option degeneration remains an open area of research.
\item \textit{Reliance on the environment reward.} The strength of the methods we presented in this section is that they do not require a human-defined objective for learning the hierarchy. 
As such, these methods heavily rely on an informative environment reward. For example, in feudal methods, if the high-level policy is poorly trained due a sparse environmental rewards, it might output goals that fail to drive the learning progress of the lower-level policy. 
To address the exploration challenge, recent methods like HAC-Explore incorporate a novelty-based intrinsic rewards \citep{McClinton2021HACEA} or demonstrations \citep{gupta2019relay} to solve longer-horizon tasks. 
\end{itemize}

\subsection{Curriculum and Hindsight Learning}
\label{sec:curriculum}

Within a complex environment, there exists a diversity of subgoals that are interesting for an agent. Some of these subgoals might be easily achievable, whereas others would simply be impossible to complete for an agent's current capabilities. How could, then, such an agent learn to achieve difficult goals? An effective strategy would be to try to achieve a curriculum of goals, where the complexity of each attempted goal increases continuously with the agent's capabilities. The idea of curriculum learning has a long history in AI that goes beyond the RL setting \citep{Kaplan2003MaximizingLP,schmidhuber2004optimal,Bengio2009CurriculumL,Schmidhuber2011PowerPlayTA}. A central question then becomes, how should one prioritize which goal or task should be attempted at any given time? Taking the HRL perspective, we can rephrase this question as: how should the high-level policy select the next goal? 
This question can be formalized through the following objective function:
\begin{align}
\max_{\mu}  \mathbb{E}_{g \sim \mu} \left[ \mathbb{E}_{\pi} [r^g] \right]
\label{eq:curriculum}
\end{align}
where the goal-conditioned policy, $\pi'$, used in the expectation, $\mathbb{E}_{\pi'}[\cdot]$, depends on the choice of the goal selection distribution $\mu$. Here, $\mu$ can be interpreted as a high-level policy over goals. 
Specifically, $\pi'$ is obtained by starting with an initial policy $\pi_0$ and applying $N$ iterative updates. For each iteration $k=1, \dots, N$, a goal $g_k \sim \mu$ is sampled, and the policy is updated via $\pi_k = U_{g_k}(\pi_{k-1})$, where the update rule $U_{g_k}$ aims to maximize the reward $r^{g_k}$ associated with goal $g_k$, i.e.  through policy gradient updates. The final policy used in Equation \ref{eq:curriculum} is $\pi' = \pi_N$. The optimization will thus find the distribution $\mu$ which, when used to update $\pi$, would lead to the best performance as measured across all goals $\mathscr{G}$.

The objective of Equation~\ref{eq:curriculum} is also referred to as the global learning progress (LP), and is, as such, intractable. Researchers have thus approximated this objective through local measures of $\text{LP}_{\text{local}}$ \citep{stout2010competence,Baranes2013ActiveLO,Forestier2017IntrinsicallyMG,Colas2018CURIOUSIM}, which can be defined as,
\begin{align}
    \text{LP}_{\text{local},g} = V_{\pi_t,g} - V_{\pi_{t-i},g},
\label{eq:learning_progress}
\end{align}
where $V_{\pi_t,g}$ is the estimate of the performance of the updated policy after $t$ iterations on goal $g$ and $V_{\pi_{t-i}}$ is that of the policy at iteration $t-i$. These values are usually obtained through Monte Carlo estimates by rolling out policies over multiple episodes, thus possibly covering a subset of the possible goals within the goal space. 

In such methods, the high-level policy $\mu$ is often optimized through multi-arm bandit algorithms rather than through RL. In other words, $\mu$ maximizes the following one-step reward: $\max_{\mu}  = \mathbb{E}[ r^{\mu} ] = \mathbb{E}[ \text{LP}_{\text{local},g} ] $, which can then be defined as
\begin{equation}
\mu_t(g) = \frac{\exp(|E_t(g)|/e)}{\sum_{g' \in \mathscr{G}}^{}\exp(|E_t(g')|/e)},
\label{eq:bandit-hlp-upd}
\end{equation} 
where $e$ is the temperature and $E_t$ is an exponential moving average of the rate of change in performance on goal $g$,
\begin{equation}
E_{t+1}(g) = (1-\alpha)E_t(g) + \alpha \text{LP}_{\text{local},g}.
\label{eq:exp_avg_lp}
\end{equation} 
The global learning process objective can be approximated through other means, which we discuss in the following sub-section on the benefits and opportunities.

\begin{algorithm}[t]
\caption{Curriculum Learning via Learning Progress: Instantiation of 
Algorithm~\ref{alg:hrl_generic}.}
\label{alg:curriculum_lp}
\begin{algorithmic}[1]
\Require Goal space $\mathscr{G}$, temperature $e$, smoothing parameter $\alpha$.
\State Initialize goal-conditioned policy $\pi_\theta$, 
       high-level policy $\mu$, 
       progress estimates $E(g) \leftarrow 0$ for all $g \in \mathscr{G}$,
       dataset $\mathscr{D} \leftarrow \varnothing$.
\Statex
\Procedure{OptionSpecification}{$\mathscr{O}, \mathscr{D}$}
    \State Collect additional goals $G$ from $\mathscr{D}$.
    \ForAll{$g \in G$}
    \State Define the option reward
           $r^g$.
    \Statex \Comment{$r^g$ examples: distance to a target outcome, success indicator, etc.}
    \State $\Omega \leftarrow \Omega \cup \{r^g\}$
\EndFor
    \State \Return $\Omega$
    \Comment{Defines what each goal-conditioned option should achieve}
\EndProcedure
\Statex
\Procedure{OptionLearning}{$\mathscr{O}, \mathscr{D}, \Omega$}
    \State Update $\pi_\theta(\cdot \mid s, g)$ to maximize $r^g$
           using transitions sampled from $\mathscr{D}$.
    \State Construct options $\mathscr{O}$ from learned goal-conditioned behaviours.
    \Statex \Comment{$\forall s,\ \mathscr{I}(s)=1$; $\beta(s)=\mathbbm{1}\{s \text{ satisfies } g\}$}
    \State \Return $\mathscr{O}$
\EndProcedure
\Statex
\Procedure{PolicyImprovement}{$\mu, \mathscr{O}, \mathscr{D}$}
    \State Estimate a learning progress score $\text{LP}_g$ for goals $g \in \mathscr{G}$
           from $\mathscr{D}$.
    \Comment{Eq.~\ref{eq:learning_progress}}
        \State Update exponential moving average $E(g)$ using Eq.~\ref{eq:exp_avg_lp}.
    \State Update goal selection distribution:
           $
           \mu(g) \propto \exp\big(|E(g)| / e\big).
           $ \Comment{Eq.~\ref{eq:bandit-hlp-upd}}
    \State \Return $\mu$
           
\EndProcedure
\end{algorithmic}
\end{algorithm}

In addition to covering methods that produce curricula by explicitly generating goals according to a certain distribution, we also discuss \textit{implicit curricula}. In these methods, certain properties of the learning algorithm itself create a curriculum-like effect. A prominent example of an implicit curriculum is hindsight experience replay (HER) \citep{Andrychowicz2017HindsightER}, which stores experience generated by seeking a certain goal $g$ in a buffer called an experience replay, and relabels such experience with a variety of other goals $g'$. We present HER in Algorithm \ref{alg:her_inst}, where we highlight the operations that differ from a standard experience replay. 
In particular, hindsight relabeling is governed by a goal sampling strategy $\mathbb{S}$, while the high-level policy $\mu$ specifies the distribution over original goals used during interaction.
In its common form, HER relabels failed trajectories using the achieved final state as the goal, yielding goals that naturally progress from easy to increasingly challenging.









\begin{algorithm}[t]
\caption{Hindsight Experience Replay (HER): Instantiation of 
Algorithm~\ref{alg:hrl_generic}.}
\label{alg:her_inst}
\begin{algorithmic}[1]
\Require Goal relabeling strategy $\mathbb{S}$.
\State Initialize goal-conditioned policy $\pi_\theta$, 
       dataset $\mathscr{D} \leftarrow \varnothing$.
\Statex
\Procedure{Interact}{$E, \mathscr{O}, \mu, \mathscr{D}$}
    \State Sample initial state $s_0$ and goal $g \sim \mu$.
    \State Generate trajectory for an episode $(s_0, a_0, \dots, s_T)$ using $\pi_\theta(\cdot|s,g)$.
    \For{$t = 0$ to $T-1$}
        \State Compute reward $r_t = r^g(s_t, a_t)$.
        \State Store $(s_t, a_t, r_t, s_{t+1}, g)$ in $\mathscr{D}$.
        
        \Statex \Comment{HER-specific operations below}
        
        \State Sample (relabel) additional goals $\mathscr{G}' \sim \mathbb{S}$.
        \ForAll{$g' \in \mathscr{G}'$}
            \State Compute hindsight reward $r'_t = r^{g'}(s_t, a_t)$.
            \State Store $(s_t, a_t, r'_t, s_{t+1}, g')$ in $\mathscr{D}$.
        \EndFor
    \EndFor
    \State \Return $\mathscr{D}$
\EndProcedure
\Statex
\Procedure{OptionSpecification}{$\mathscr{O}, \mathscr{D}$}
    \State Sample a batch $B \subseteq \mathscr{D}.$
    \ForAll{$(s_t, a_t, r_t, s_{t+1}, g_t) \in B$}
        \State Add option reward $r^{g_t}$ to $\Omega$. \Comment{E.g., success indicator based on achieved goal}
    \EndFor
    \State \Return $\Omega$
\EndProcedure
\Statex
\Procedure{OptionLearning}{$\mathscr{O}, \mathscr{D}, \Omega$}
    \State Update $\pi_\theta(\cdot|s,g)$ using transitions related to $g$ from $\mathscr{D}$ to maximize $r^g\in \Omega$.
    \State Construct options $\mathscr{O}$ from learned goal-conditioned behaviours.
    \Statex \Comment{$\forall s,\ \mathscr{I}(s)=1$; $\beta(s)=\mathbbm{1}\{s \text{ satisfies } g\}$}
    \State \Return $\mathscr{O}$
\EndProcedure

\end{algorithmic}
\end{algorithm}

Research on learning from a curriculum of goals has received much more attention than we can cover here, producing a diversity of approximations to the global learning progress \citep{Forestier2016ModularAC,Matiisen2017TeacherStudentCL,Kova2020GRIMGEPLP,Akakzia2021GroundingLT}. For an in-depth review please see the surveys by \citet{colas2022autotelic} and \citet{Portelas2020AutomaticCL}.\looseness=-1

\subsubsection{Relationship to the Algorithmic Template}
\label{sec:template-curriculum}
Algorithms~\ref{alg:curriculum_lp} and~\ref{alg:her_inst} instantiate Algorithm~\ref{alg:hrl_generic} for representative methods of explicit curricula (learning-progress-based) and implicit curricula (HER), respectively.
These two instantiations highlight a key distinction in how curricula are realized. In explicit curriculum methods (Algorithm~\ref{alg:curriculum_lp}), the \textsc{PolicyImprovement} for the high-level policy $\mu$ is the main focus.
The high-level goal-selection policy is directly optimized using a learning progress score. 
In contrast, HER (Algorithm~\ref{alg:her_inst}) does not optimize $\mu$; instead, it relies on the relabeling procedure in \textsc{Interact} to induce an implicit curriculum via the strategy $\mathbb{S}$, reshaping the training data distribution without modifying the goal selection mechanism.
Conceptually, goal relabeling (line 8) and hindsight reward computation (line 10) can be viewed as part of \textsc{OptionSpecification}, as they determine which goals (option objectives) are used to train the goal-conditioned policies (option policies) in \textsc{OptionLearning}; however, in Algorithm~\ref{alg:her_inst} we place them within \textsc{Interact} to reflect the implementation.
For both algorithms, \textsc{OptionSpecification} typically runs only once and $\Omega$ is then fixed.

\subsubsection{Benefits and Opportunities}

\paragraph{Exploration.} One of the main benefits of leveraging curriculum learning to achieve goals is that the agent will be continuously pushed to the limits of its capacity. By doing so, it might discover new locations in an environment or learn completely new behaviour from the combination of previously achieved goals. A family of methods approximates the global learning progress, specifically with the intent of seeking intermediate difficulty. \citet{pmlr-v80-florensa18a} propose using a Goal Generative Adversarial Network (Goal GAN) to automatically generate a curriculum of tasks for reinforcement learning agents.
The method focuses on generating Goals of Intermediate Difficulty (GOID), defined as goals where the agent's current policy, $\pi$, achieves an expected performance $v_{\pi}$ within a specific range:
\begin{equation} 
\text{GOID}_{i} := \{g : v_{\min} \le v_{\pi} \le v_{\max}\}. 
\end{equation}
Here, $v_{\min}$ and $v_{\max}$ represent the minimum and maximum desired performance, ensuring goals are neither too easy nor too hard for the current policy $\pi_i$. The generator in Goal GAN is trained to output goals within the $\text{GOID}$ set, whereas the discriminator is trained to distinguish between goals that are within the set from those that are not.
\citet{Racanire2019AutomatedCT} introduce a setter-solver paradigm with three criteria represented through values in $[0,1]$: validity, feasibility, and coverage. Goals are sampled according to the distribution defined by these criteria, allowing for a balanced selection.
Their findings highlight that these criteria, along with conditioning on the current version of the environment, are crucial for an effective learning curriculum.
\citet{sukhbaatar2018intrinsic} instead rely on asymmetric self-play to generate a curriculum of explorative goals in reversible or resettable environments, leading to improved performance on a diverse set of tasks. \citet{campero2021learning} train a goal-generating teacher to guide a goal-conditioned student policy by proposing goals that are neither too hard nor too easy, as measured by the number of timesteps to reach the goal,
\begin{align}
r^\mu = 
\begin{cases} 
      +a & \text{if } t^+ \ge t^* \\
      -b & \text{if } t^+ < t^*, \end{cases}
\end{align}
where $a, b$ are hyperparameters that quantify the bonus and penalty, $t^{\pi}$ represents the time it took the policy to reach the goal, and $t^*$ is a hyperparameter.

Additionally, HER-based approaches \citep{Andrychowicz2017HindsightER,fang2018dher,yang2021mher} have demonstrated promising results in improving exploration compared to curiosity-driven methods. Curriculum-guided HER \citep{fang2019curriculum} introduces an explicit curriculum that transitions from curiosity-driven selection early on to goal-proximity focus in later stages, mimicking human-like exploration. Complementing HER, CER \citep{liu2019competitive} enhance exploration by introducing a competitive dynamic between two agents learning the same task, where one agent is penalized for revisiting states explored by the other.
Many more works show how curriculum learning can help in hard-to-explore environments \citep{Colas2018CURIOUSIM,Zhang2020AutomaticCL,pitis2020mega,Colas2020LanguageConditionedGG}.

\paragraph{Transfer.} Learning successfully through curricula produces a whole set of behaviours that were previously not seen.  
 \citet{oel2021xland} leverage population-based training to quantify progress on goal completion within large, open-ended, and procedurally-generated environments and tasks. Through a continuum of task difficulty, the authors show that the resulting goal-conditioned agent can generalize zero-shot to new situations. 

\paragraph{Opportunities for Research.} 

\begin{itemize}
    \item \textit{Refining the measure of progress.} One of the main challenges in deriving a curriculum of goals is accurately measuring how difficult a chosen goal is for a learning algorithm at a certain point in time. Different heuristics can work well for certain environments, for example, the number of timesteps required for reaching a goal \citep{campero2021learning}, or might involve a combination of heuristics \citep{oel2021xland}. However, such formalizations might not be generally applicable. Ideas from unsupervised environment design, where the environment evolves as well as the agent's parameters, could be particularly promising \citep{Dennis2020EmergentCA,Jiang2021ReplayGuidedAE,ParkerHolder2022EvolvingCW,Samvelyan2023MAESTROOE}. Another important desideratum is that a curriculum should continuously increase the difficulty of the goals, but should also generate interesting goals. Finding a formalization that would encode a general measure of interestingness and difficulty is still an open question. However, for many tasks of interest, such as tasks where human prior knowledge would be relevant, leveraging foundational models offers a particularly promising way to define such metrics for curriculum learning, as covered in Section \ref{sec:largemodels}. One notable work is that of 
\cite{Zhang2023OMNIOV}, which investigates whether an  LLM's common sense can be a good measure of interestingness in open-ended environments. 
\end{itemize}



\subsection{Meta Learning}
\label{sec:metalearning}

RL algorithms, such as Q-learning, learn policies; meta-RL algorithms, in contrast, aim to learn the RL algorithm itself, or parts of it, to subsequently learn a policy. This creates a bi-level optimization: the algorithm for learning the RL algorithm itself is called the \textit{outer-loop}, while the learned algorithm (which learns a policy) is called the \textit{inner-loop} \citep{Schmidhuber1987EvolutionaryPI,Thrun1998LearningTL,beck2025tutorial}. The appeal of meta-RL approaches is that if the environment demands certain properties from the RL agent (for example, transferability), then such properties will automatically be learned from data, without the explicit need for careful human ingenuity and design in every part of the training process \citep{silver21reward}.

Typically, a meta-RL algorithm consists of an inner and an outer loop. Within each of these loops, a set of parameters is being maximized. Concretely, let $\omega_{out}$ represent the parameters learned by the outer loop, and $\omega_{in}$ the parameters learned by the inner loop.  These parameters in practice represent a particular subset of the option parameters presented in Section \ref{sec:3}. 
For example, \citet{veeriah2021discovery} optimize the parameters of the option policies and the high-level policy in the inner loop, while the outer loop updates the termination functions and option reward functions (subgoals). 
In contrast, \citet{frans2018meta} can be viewed as placing the high-level policy in the inner loop, where it is adapted to solve individual tasks, while the low-level policies are treated as shared primitives that are meta-learned across tasks in the outer loop.

\subsubsection*{Meta-Gradients}
A common instantiation of meta-RL algorithms is through the use of meta-gradients. In the inner loop, the agent updates the inner parameters, 
\begin{equation}
\omega_\text{in}' \leftarrow \omega_\text{in} + \alpha_\text{in} \nabla_{\omega_\text{in}} J_\text{in}(\omega_\text{in}),
\end{equation}
where $J_\text{in}$ is an arbitrary objective that depends on $\omega_\text{in}$. To obtain the meta-gradients, we assume that the outer parameters depend on the inner parameters. Data is then collected with the updated inner parameters in order to proceed to the following update,
\begin{align}
    \omega_\text{out}' &\leftarrow \omega_\text{out}' + \alpha_\text{in} \nabla_{\omega_\text{out}} J_\text{out}(\omega_\text{in}'(\omega_\text{out})), \\ 
\omega_\text{out}' &\leftarrow \omega_\text{out}' + \alpha_\text{out} \nabla_{\omega_\text{in}'} J_\text{out}(\omega_\text{in}'(\omega_\text{out})) \nabla_{\omega_\text{out}} \omega_\text{in}'(\omega_\text{out}),
\end{align}
where $\nabla_{\omega_\text{out}} \omega_\text{in}'(\omega_\text{out})$ encodes how the outer loop parameters affected the updated inner loop parameters. The objectives $J_\text{in}$ and $J_\text{out}$ may differ in various ways, such as defining different distributions over tasks. 

\citet{veeriah2021discovery} leverage meta-gradients \citep{finn2017model, xu2018meta} to learn options in high-dimensional navigation environments. In the inner loop, they update the option policies parameters, $\theta$, and the high-level policy parameters, $\kappa$:
\begin{align}
    \theta' &\leftarrow \theta + \alpha_\theta (G_t - q_{\pi}(s_t,o_t)) \cdot \nabla_{\theta} [\log \pi_\theta(a_t | s_t, o_t) -  q_{\pi}(s_t,o_t)] \label{eq:1}, \\
    \kappa' &\leftarrow \kappa + \alpha_{\kappa} (G_{t}^{\mu} - v_{\mu}(s_{t})) \cdot \nabla_{\kappa} [\log \mu_\kappa(o_t | s_t) -  v_{\mu}(s_t)], \label{eq:2}
\end{align}
where $G_t$ is the option policy return (see Section \ref{sec:options}), and $G_{t}^{\mu}$ is an $n$-step return for the high-level policy defined as
$G_t^{\mu} = \sum_{j=1}^{n} \gamma^{j} r_{t+j} - \gamma^n c + \gamma^{n+1} V_{\mu}(s_{t+n})$ where $c$ is a switching cost added on option terminations, similar to \citet{Harb2017WhenWI}. The outer loop is instantiated through these updates to the parameters $\nu$ of the option reward function $r^o_\nu$, and the parameters $\psi$ of the termination function $\beta_\psi^o$,
\begin{align}
\psi &\leftarrow \psi + \alpha_{\psi} (G_t^{\mu} - v_{\mu}(s_t)) \nabla_{\psi} \log \pi_{\theta'(\psi,\nu)}(a_t | s_t, o_t), \\
\nu &\leftarrow \nu + \alpha_{\nu} (G_t^{\mu} - v_{\mu}(s_t)) \nabla_{\nu} \log \pi_{\theta'(\psi,\nu)}(a_t | s_t, o_t). 
\end{align}
The outer loop updates the option-reward and termination meta-parameters using a new trajectory generated by interacting with the environment using the most recent inner-loop parameters, $\theta'(\psi, \nu)$ and $\kappa'(\psi, \nu)$, which depend on the outer loop parameters. The update in the outer loop assesses the impact of updates to the high-level policy, $\mu_{\kappa}$, and option policies, $\pi_{\theta}$, and it may involve a different distribution of tasks than the one used in the inner loop, as is common in meta learning \citep{finn2017model}.

\subsubsection*{Black-box Meta Reinforcement Learning}

\begin{figure}
    \centering
    \includegraphics[width=0.95\linewidth]{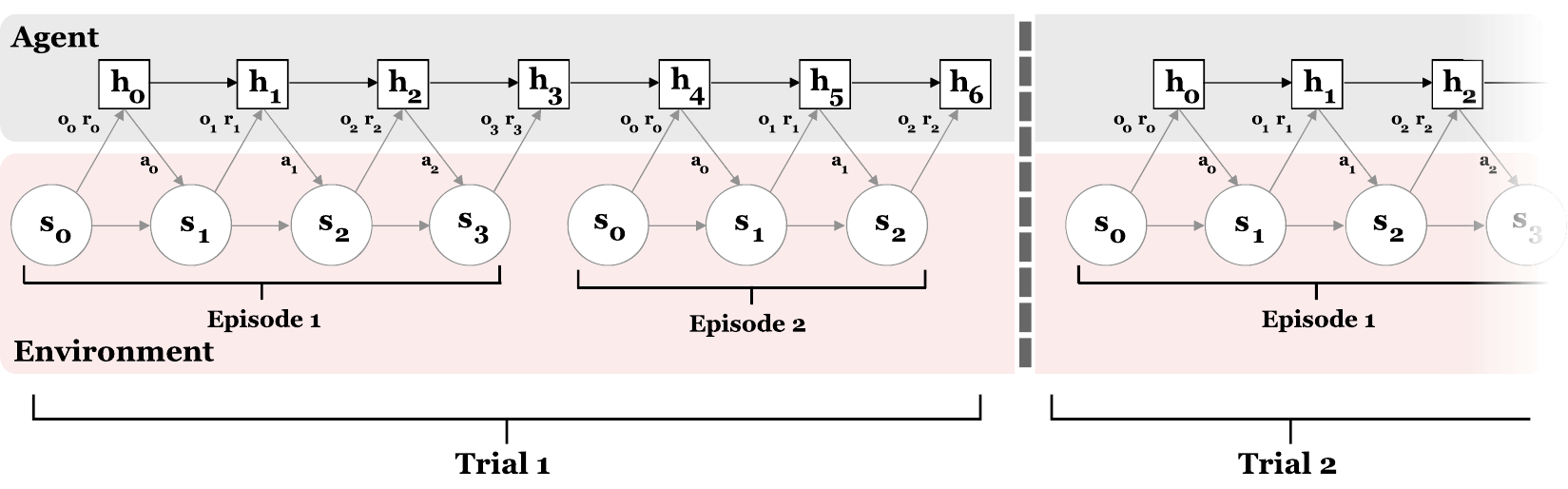}
    \caption{Black-box meta reinforcement learning. Trials consist of multiple episodes during which the hidden state, $h_i$, of the agent is unrolled. The hidden state is only reset between trials. Figure reproduced from the work of \cite{duan2016rl2fastreinforcementlearning}.}
    \label{fig:meta}
\end{figure}

In black-box meta RL \citep{wang2017learningreinforcementlearn,duan2016rl2fastreinforcementlearning}, an agent interacts with a sequence of different tasks drawn from an arbitrary distribution, $p \in \Delta(\xi)$.  Each interaction with a task, or distribution of tasks, is considered a trial, which itself consists of $N$ episodes, represented in Figure \ref{fig:meta}. During a trial, the agent receives observations, rewards, and termination signals from the environment, where episode termination signals represent the episode boundaries. These variables are used to update the agent's internal memory $h$, which is typically represented by the hidden state of an RNN \citep{Hochreiter1997LongSM} or the context of a transformer network \citep{Vaswani2017AttentionIA}. Importantly, the agent continuously updates $h$ across episodes within the same trial; the memory is only reset at the end of each trial. The overall goal is to maximize the total reward accumulated over an entire trial, 
\begin{align}
    \max_{\pi} \mathbb{E}_{\xi \sim p }\Biggr[ \sum_{\text{episode}=1}^N \mathbb{E}_{\pi}\Big[ \sum_t r^{\xi}(s_t,a_t)\Big] \Biggr],
\end{align}
where $r^{\xi}$ is the reward associated with task $\xi$. This objective incentivizes the agent to learn how to adapt its policy based on the experience gathered so far during a trial, effectively forcing it to implicitly learn, through the updates to its policy's memory $h$, a reinforcement learning rule capable of efficient adaptation to new tasks. When further conditioning the policy $\pi$ on the task's goal $g$, as done by \citet{bauer2023ada}, this approach can lead to human-timescale adaptation.

\subsubsection{Relationship to the Algorithmic Template}
We do not provide a generic template related to Algorithm~\ref{alg:hrl_generic} for this section, as different algorithms optimize different components in the inner and outer loops. For example,
\citet{frans2018meta} optimize a task-specific high-level policy in the inner loop while treating the option policies as shared primitives, and update these shared option policies in the outer loop. Both levels optimize the same task reward, so the method does not instantiate \textsc{OptionSpecification}; the outer loop instantiates \textsc{OptionLearning}, while the inner loop corresponds to \textsc{PolicyImprovement} of the high-level policy.
\citet{veeriah2021discovery} instead learns option policies and the high-level policy using RL on sampled tasks in the inner loop, which corresponds to \textsc{OptionLearning}, and meta-learns option reward (subgoal) parameters and termination functions via meta-gradients in the outer loop, which corresponds to \textsc{OptionSpecification}. 
For black-box meta-RL, if the hidden state $h$ is regarded as an option identifier, it can be viewed as an instantiation of Algorithm~\ref{alg:hrl_generic} with an implicit option space. \textsc{OptionSpecification} is trivial, as no explicit objectives are defined beyond the task reward $r^{\xi}$. \textsc{OptionLearning} optimizes the option policy $\pi(a \mid s, h)$, while \textsc{PolicyImprovement} is instead realized through the update rule of $h$ (e.g., via an RNN), which governs how the high-level behaviour evolves over time.

\subsubsection{Benefits and Opportunities}

\paragraph{Transfer.} 

As discussed earlier, a major benefit of learning options is that of reuse: options learned in one part of the state-space could speed up learning in another \citep{Konidaris2007BuildingPO,taylor2009transfer}. Some methods have tried to discover transferable options using meta-learning. For example, MLSH \citep{frans2018meta} discovers a set of policies and trains a high-level policy to select among them. The meta-objective trains these components such that the high-level policy can quickly learn to solve new tasks from a distribution by reusing the learned skills, making them reusable across a pre-specified task distribution \citep{nam2022metaskills,gupta2024meta,fu2023meta}.  MODAC \citep{veeriah2021discovery} uses meta-gradients \citep{xu2018meta,oh2020discovering} to do the same: an outer loop learns option reward functions and termination functions that an inner loop maximizes using policy gradients. The outer loop of the optimization learns from the reward coming from the environment.


\paragraph{Exploration.} Meta-learning approaches have also sought to address the exploration question in non-stationary and multi-task settings. When the agent finds itself in a new environment, how can it systematically leverage prior experience to guide targeted exploration of the new environment? This problem is called \textit{meta-exploration} by \citet{beck2025tutorial}. For example, when someone is in a new house and they have to look for utensils, they begin their search from the kitchen; similarly, we would like to create RL agents that can direct their exploration for quick adaptation in new environments \citep{gupta2024meta}. This is one of the motivations for the Adaptive Agent (AdA) \citep{bauer2023ada} that uses meta-learning to train a policy capable of human-timescale adaptation in a massive, combinatorial task space \citep{oel2021xland}. Specifically, they use black-box meta-RL: the policy is implemented as a Transformer-XL model \citep{Dai2019TransformerXLAL}. This model $\pi_\theta$ takes the history $h$ of interactions within the current episode (past states, actions, rewards) and goal description, $g$, as input to determine the next action. The adaptation happens implicitly within the recurrent state of the model. The combinatorial complexity of the environments allows for careful selection of tasks that are at the appropriate difficulty given the current agent capabilities, generating an effective meta-learning curriculum. As such, AdA mixes ideas from meta-learning as well as curriculum learning, which we cover in the next section.\looseness-1

\paragraph{Opportunities for Research.} 
\begin{itemize}
    \item \textit{Relaxing the multi-task formulation.} Meta-learning approaches have demonstrated abilities of transfer, adaptation, and meta-exploration---abilities that have been challenging to scalably acquire using other techniques. Furthermore, meta-learning via in-context learning \citep{dong2022survey,bauer2023ada,raparthy2023generalization} provides a scalable, and potentially simpler way, to acquire these crucial capabilities. Existing meta-learning approaches rely on a specialized multi-task formulation, with clear task boundaries and episodes. Methods that lift these assumptions will be able to bring these capabilities to a wider variety of settings. For an in-depth review of meta learning approach, please refer to \citet{beck2025tutorial}.
\end{itemize}

\subsection{Intrinsic Motivation}\label{sec:other_im_od}

\begin{figure}[h]
    \centering
    \includegraphics[width=0.9\linewidth]{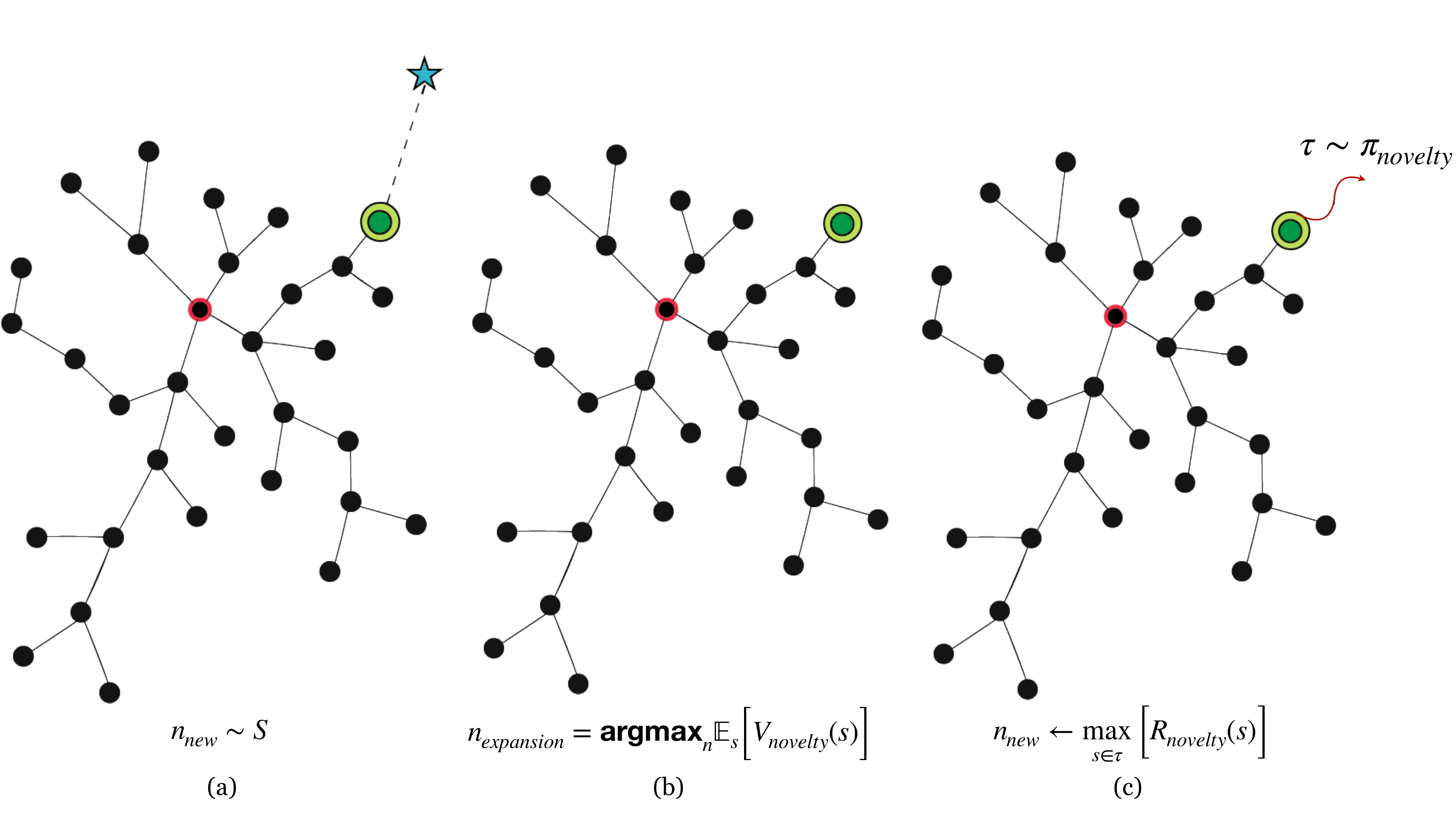}
    \caption[IM-DSG Illustration]{(a) In the original DSG algorithm, a state is sampled uniformly at random (blue star) from the state-space $\mathscr{S}$ and the graph is pulled towards it via its nearest neighbor node (green). (b) The IM-DSG agent uses intrinsic motivation to identify a node to expand using an exploration value function $v_\text{novelty}$. (c) Once the agent reaches the expansion node, it executes an exploration policy $\pi_\text{novelty}$, and the most novel state in the resulting trajectory is identified as a target for a new skill.}
    \label{fig:im-dsg}
\end{figure}

Intrinsic Motivations (IM) drive actions for their own sake, meaning that they are not in service of achieving an obvious, externally specified goal; instead, they are in service of augmenting knowledge and learning skills whose utility only becomes apparent later on \citep{oudeyer2009im,barto2005intrinsic,berlyne1965arousal,harlow1950}.
Computationally, this can be formalized using notions of information gain \citep{Bellemare2016pseudocounts}---an agent may take actions that result in new information about its environment, even if it requires forsaking extrinsic reward in the short term. IM underpins a developmental approach where an agent learns reusable skills autonomously, preparing it for various future challenges. For example, children, as intrinsically motivated biological agents, develop skills by engaging in activities that yield interesting, memorable outcomes \citep{gopnik2009scientist}; these skills improve in efficiency with repetition and can be strategically reproduced for specific goals \citep{barto2024intrinsically}. Such behaviours are well represented by options, with the intended outcomes encapsulated in the options' subgoals \citep{singh2004im}. While we have discussed IM-based option discovery approaches like empowerment maximization, spectral methods, and bottleneck discovery, this section explores additional methods not covered by these categories.

\subsubsection*{Novelty-Based Discovery}

One common intrinsic motivation signal is novelty, which decreases with repeated state visitations. For example, upon visiting state $s$, an agent might receive
\begin{equation}\label{eq:count-bonus}
r^{\text{int}}(s) = N(s)^{-1/2}  ,  
\end{equation}
where $N(s)$ is the visit count. This count-based bonus is one of several common choices for encouraging exploration of infrequently seen states by making familiar states less rewarding \citep{auer2008ucrl2,strehl2008mbieeb,Bellemare2016pseudocounts,lobel2023flipping}. An example of using this for option discovery is provided by the Relative Novelty algorithm \citep{Simsek2004UsingRN}. Here, a state $s$ is deemed a good subgoal if it leads to experience that is significantly more novel than the experience preceding it. Let $n(s)$ be a novelty score (e.g., inverse visitation count from Equation~\ref{eq:count-bonus}), and let $w^+$ and $w^-$ be fixed-size windows of future and past states, respectively. The relative novelty at time $t$ is then computed as
\begin{equation}\label{eq:relative-novelty} 
\mathrm{RN}(s_t) = \frac{1}{|w^+|} \sum_{s \in w^+} n(s) \Big/ \left( \frac{1}{|w^-|} \sum_{s \in w^-} n(s) \right).   
\end{equation}
States with high relative novelty are likely to be gateways to unexplored regions (e.g., doorways in Figure~\ref{fig:both-pdfs}), and can be automatically selected as subgoals. An option is then created to reach each such subgoal from a broader initiation set, often by learning a dedicated policy using intrinsic reward. In this way, the agent transforms spikes in novelty into reusable skills without knowledge of external task rewards.

A related example is First Return, then Explore (FRTE) \citep{Ecoffet2020FirstRT}, which formalizes intrinsic motivation using count-based novelty over discretized ``cells'', which are pre-specified state abstractions that serve as option subgoals (an example state abstraction is spatially downsampling image-based observations). Every time a new cell is encountered, the agent logs it in an archive. The policy is then conditioned to return to these cells to deepen exploration. FRTE selects target cells based on their inverse visitation count, returning to underexplored frontiers before taking random, exploratory actions. This loop results in a growing archive of reachable states, each effectively defining an option subgoal.
This return-then-explore strategy also has a model-based counterpart: \citet{bagaria2025imdsg} extend the Deep Skill Graphs algorithm discussed in Section~\ref{sec:skillchaining} to image-based observation spaces where a meaningful distance metric is not readily apparent. Their algorithm, \textit{Intrinsically Motivated Deep Skill Graphs} (IM-DSG), learns a graph-based model of the world---nodes of the graph represent option subgoal regions (abstract states) and edges represent option policies (abstract actions). Figure~\ref{fig:im-dsg} illustrates the main steps of the algorithm: first, the agent picks an existing node based on how much that node is expected to contribute to exploration (Figure~\ref{fig:im-dsg}(b)), then the agent plans with its abstract model using dynamic programming to determine the options it needs to execute to reach the sampled node, from where it executes a novelty-seeking exploration policy (Figure~\ref{fig:im-dsg}(c)). States visited by the novelty-seeking policy are candidates for creating a new node in the graph, similar to the Relative Novelty algorithm \citep{Simsek2004UsingRN} discussed earlier.

\subsubsection*{Factored and Causal Discovery}

The novelty-based methods above identify \textit{where} in the state space to create option subgoals. A complementary approach is to identify \textit{what aspect of the state} an option should control, using the causal structure of factored MDPs. A factored MDP \citep{boutilier2000factored} assumes the state can be described by a vector of state variables and that the transition dynamics factorize according to a dynamic Bayesian network (DBN). HEXQ \citep{hengst2002discovering} was an early algorithm in this category. HEXQ automatically decomposes a factored MDP into subtasks by detecting \textit{exits}---states where the transition dynamics of a lower-level state variable can no longer be predicted independently of higher-level variables. More formally, if $X$ and $Y$ are state variables ordered by their rate of change (faster-changing variables are lower-level), an exit is identified where the conditional entropy $\mathcal{H}(Y' \mid X, a)$ increases sharply, indicating that a transition in $X$ cannot be explained without accounting for $Y$. Each such transition is marked as an exit and treated as a subgoal. HEXQ then learns a hierarchy of options, each option driving the agent toward one of these subgoals---lower-level options correspond to frequently-changing variables, whereas higher-level options handle more slowly-changing aspects. In this way, HEXQ yields an option hierarchy spanning multiple levels of temporal abstraction.

A similar approach is proposed by \citet{jonsson2006factored}, who analyze the structure of a learned DBN to extract a causal abstraction hierarchy. For each action, they examine the DBN's parent-child dependencies: if variable $X_i$ influences the next-state value of $X_j$, i.e., $X_i \in \text{Parent}(X_j' \mid a)$ for some action $a$, then $X_i$ is said to causally affect $X_j$. These dependencies define a directed graph over state variables, which is decomposed into strongly connected components (SCCs). The SCCs are then topologically ordered to yield levels of abstraction---variables in earlier components are controlled first, while later components are conditioned on them. This is because earlier variables in the topological ordering tend to be those that causally influence, but are not influenced by, variables in later components. While their algorithm assumes access to the transition model, \citet{Vigorito2008AutonomousHS} propose a model-free algorithm that incrementally builds DBNs online through exploration. When new dependencies are detected---e.g., when variable $X_i$ begins to influence $X_j$---a new option is instantiated to induce that dependency reliably. This learning process can be guided by structure learning techniques (e.g., maximizing Bayesian Information Criterion), or by identifying transitions that lead to salient changes in state abstractions. More recently, \citet{nayyar2024autonomous} cluster states based on the temporal-difference (TD) error:
\begin{equation}
    \delta_t = r_t + \gamma v(s_{t+1}) - v(s_t),
\end{equation}
which serves as a proxy for learning progress. Regions with high variance in $\delta_t$ are recursively split, producing a symbolic abstraction over state variables and spawning new options targeted to regions where prediction error remains high. This method resembles HEXQ, but instead of focusing on the frequency of variable changes, it uses the TD-error incurred by candidate state abstractions.

One downside of these factored approaches is that they assume the agent observes factored state variables, which requires significant domain knowledge. \citet{bagaria2025going} address this limitation by developing an agent that learns to identify relevant features directly from image observations. When their agent encounters a particularly novel state, it uses counterfactual analysis (specifically, they use the DeepSHAP algorithm; \citealp{shapley1953value,Lundberg2017AUA,shrikumar2017learning}) to isolate which visual features are responsible for the novelty of that state. Then, the agent learns a classifier that focuses only on these salient features \citep{singh2004im}, ignoring other aspects of the image. This feature-specific classifier serves as an abstract subgoal \citep{bagaria23scaling} for option learning, enabling factored skill discovery without requiring pre-specified state decompositions.

\subsubsection{Relationship to the Algorithmic Template}

\begin{algorithm}[t]
\caption{Novelty-Based Option Discovery: Instantiation of Algorithm~\ref{alg:hrl_generic}.}
\label{alg:novelty_inst}
\begin{algorithmic}[1]
\Procedure{OptionSpecification}{$\mathscr{O}, \mathscr{D}$}
    \State Update novelty scores $n(s)$ from visitations in $\mathscr{D}$
        \Comment{e.g., Equation~\ref{eq:count-bonus}}
    \State $B \leftarrow \{s : \mathrm{RN}(s) > c_{\mathrm{RN}}
            \text{ and } s \text{ is not the subgoal of any } o \in \mathscr{O}\}$
    \Statex \Comment{e.g., relative novelty spikes (Equation~\ref{eq:relative-novelty})}
    \State $\Omega \leftarrow \varnothing$
    \For{$b \in B$}
        \State $r^o_b(s) \leftarrow \mathbbm{1}(s = b)$
            \Comment{Example subgoal reward}
        \State $\Omega \leftarrow \Omega \cup \{(b, r^o_b)\}$
    \EndFor
    \State \Return $\Omega$
\EndProcedure
\Statex
\Procedure{OptionLearning}{$\mathscr{O}, \mathscr{D}, \Omega$}
    \For{$\omega = (b, r^o_b) \in \Omega$}
        \State $\beta_b(s) \leftarrow \mathbbm{1}(s = b)$;\;
               $\mathscr{I}_b(s) \leftarrow 1$
            \Comment{Example termination and initiation}
        \State $\mathscr{O} \leftarrow \mathscr{O} \cup \{(\mathscr{I}_b, \pi_b, \beta_b)\}$
    \EndFor
    \State Train each option policy $\pi(\cdot|s,o)$, $o \in \mathscr{O}$,
           to maximize $r^o$.
    \State \Return $\mathscr{O}$
\EndProcedure
\end{algorithmic}
\end{algorithm}

As in previous sections, the primary variation lies in how
\textsc{OptionSpecification} selects subgoals; these methods do it through internally  generated signals such as novelty scores, relative novelty spikes, or structural features like causal exits.
Algorithm~\ref{alg:novelty_inst} presents the template for novelty-based discovery, where \textsc{OptionSpecification} thresholds a novelty criterion (here, relative novelty with threshold $c_{\mathrm{RN}}$) to select subgoals.
Option construction follows the same defaults as bottleneck discovery (Algorithm~\ref{alg:bottleneck}), but with two differences that reflect the interleaved setting: \textsc{OptionSpecification} only creates options for subgoals that are not already targeted by an existing option, and \textsc{OptionLearning} continues to train all option policies as new experience arrives.
As in the other instantiations, $\Omega$ denotes the newly specified objectives; the objectives of existing options persist within $\mathscr{O}$.
A distinctive feature of novelty-based methods is that these criteria are often 
\textit{non-stationary}: novelty scores decrease as the agent 
accumulates experience, so \textsc{OptionSpecification} must be 
periodically re-invoked as new data arrives from \textsc{Interact}. 
This tight interleaving between \textsc{Interact} and 
\textsc{OptionSpecification} resembles the incremental pattern seen in 
algorithms such as L-Cut (Section~\ref{sec:od/graph-based}) and skill chaining (Section~\ref{sec:skillchaining}). Factored approaches like 
HEXQ~\citep{hengst2002discovering} and the causal abstraction method of 
\citet{jonsson2006factored}, by contrast, solve 
\textsc{OptionSpecification} as a one-time analysis of the DBN 
structure, similar in phasing to bottleneck preprocessing. The online 
variant of \citet{Vigorito2008AutonomousHS}, which incrementally builds 
the DBN through exploration, is closer to the interleaved paradigm.

\subsubsection{Benefits and Opportunities}
\paragraph{Exploration.} Intrinsic motivation is perhaps most often used to facilitate exploration, especially when extrinsic rewards are sparse, delayed, or misleading---although early methods were equally motivated by the acquisition of reusable skills \citep{singh2004im,barto2005intrinsic}. By rewarding novelty, surprise, or learning progress, IM helps the agent identify and prioritize skills that could result in mastery of the environment \citep{Veeriah2018ManyGoalsRL}. The resulting behaviours are often more structured and directed than undirected exploration strategies like $\epsilon$-greedy or softmax sampling.

\paragraph{Transfer.} Options discovered through intrinsically motivated factor-based methods are transferable because they target changes in individual state variables or abstract subspaces. For example, in HEXQ, an option that changes a frequently occurring variable---like an agent's position---can be reused across many contexts where that variable matters, regardless of the values of other variables. Similarly, in methods based on causal abstraction \citep{jonsson2006factored,Vigorito2008AutonomousHS}, options are constructed to affect only a specific part of the environment while assuming other parts are stable or independently controllable. This modularity reduces interference between options because each option specializes in a different part of the environment, which additionally encourages compositionality in behaviour space. As a result, once such an option is learned, it can be reused across multiple tasks or contexts without retraining, greatly improving sample efficiency.\looseness=-1

\paragraph{Opportunities for Research.} 
\begin{itemize}
    \item \textit{Generalize factored approaches to large observation spaces.}
    Factored approaches like HEXQ assume access to explicit state variables, limiting their applicability to high-dimensional observation spaces such as images or sensor data. Recent works \citep{bagaria2025going,RodriguezSanchez2025FromPT,Kumar2025QuestioningRO,higgins2016beta,Kim2018DisentanglingBF} demonstrate some promising initial results, but significant challenges remain in automatically discovering meaningful factors without domain knowledge. Future research could focus on developing methods that identify relevant features and their causal dependencies in continuous, high-dimensional spaces while maintaining the transferability and composability benefits of factored approaches.
    
    \item \textit{Improved estimates of novelty.}
    While count-based novelty measures work well in discrete spaces, they struggle in continuous environments where exact state revisitation is unlikely. Recent advances \citep{Bellemare2016pseudocounts,burda2018exploration,Machado20,lobel2023flipping,Guo2022BYOLExploreEB} provide neural network-based alternatives, but fundamental challenges persist in distinguishing meaningful novelty from environmental stochasticity (sometimes colloquially called the ``noisy TV'' problem). Future work should develop more robust novelty estimation methods that can maintain exploration incentives across different timescales, handle function approximation errors gracefully, and integrate structural priors about environment dynamics to focus on causally relevant state changes.
    
    \item \textit{Connections to other discovery techniques.}
    Intrinsic motivation approaches have developed largely independently from other option discovery methods like graph clustering, empowerment, and spectral decomposition techniques. However, recent theoretical work suggests that some of these approaches may be unified under information-theoretic frameworks \citep{achiam2018variational}. Establishing formal connections between intrinsic motivation and other discovery paradigms (for example, the ROD cycle; \citealp{machado2023temporal}) could enable hybrid approaches that leverage the complementary strengths of different methods.\looseness=-1
\end{itemize}

\subsection{Directly Optimizing for the Benefits of Hierarchical Reinforcement Learning}
\label{sec:formalguarantee}

Many of the option discovery methods that we have discussed so far rely on \textit{proxy} objectives; these objectives include finding bottleneck states, empowerment maximization, more reliable composability, and so on. The intuition is that if the agent had options that maximized these proxy objectives, it would unlock agent-level capabilities such as effective exploration, credit assignment, or transfer. Indeed, these methods often show empirical success in some scenarios, but the formal connection between these proxy objectives and the overall objectives of the agent is unclear \citep{Solway14Optimal}. For example, options that target bottleneck states are empirically useful in some tasks, but what kind of performance can we expect from the same technique in an entirely different problem? In fact, several papers have shown that not all skills are created equal---that is, options that are perfectly suited for a particular task might severely hurt agent-level objectives in other tasks \citep{AAMAS08-jong,Solway14Optimal}. To address this gap, a class of methods---initiated by \citet{Solway14Optimal}---has sought to discover options with precise guarantees on agent-level objectives. These methods explicitly state the performance criterion of the agent and then derive an algorithm that discovers options with bounded loss on that criterion.

In the language of Algorithm~\ref{alg:hrl_generic}, these methods share a common structure: \textsc{OptionSpecification} is formulated as an explicit optimization problem over a precisely defined agent-level objective---such as planning time, cover time, or sample complexity during transfer. The option objectives~$\Omega$ are the solution to this optimization, and \textsc{OptionLearning} then trains option policies to achieve them, typically using standard methods. What distinguishes this family is that the specification objective is not a proxy: it is a formally defined quantity that is either the agent-level objective itself or provably related to it. Most methods in this section solve \textsc{OptionSpecification} as a one-time combinatorial optimization before any learning or deployment, similar in phasing to some of the bottleneck methods of Section~\ref{sec:od/graph-based}. However, whereas bottleneck methods use heuristic criteria to identify subgoals, these methods derive their subgoals from a well-defined performance criterion---often proving NP-hardness of the resulting optimization and providing approximation algorithms with bounded suboptimality. We organize the discussion by the agent-level objective being optimized: planning efficiency, exploration, credit assignment, and transfer.

\subsubsection{Benefits and Opportunities}

\paragraph{Planning.} In the planning context, option discovery can be framed as the search for a set of options that minimizes the \textit{planning time}---defined as the number of iterations a planning algorithm (e.g., value iteration) takes to approximate the optimal value function $v^{*}$ within some accuracy $\epsilon$ \citep{silver2012compositional, jinnai2019planning}. Here, \textsc{OptionSpecification} amounts to solving a combinatorial optimization: find at most~$k$ options that minimize the number of value iteration steps needed to reach~$\epsilon$-accuracy. Formally, given a maximum allowable value error $\max_{s\in\mathscr{S}}|v^{*}(s)-\hat{v}(s)| \leq \epsilon$, the goal is to find a set of at most $k$ options $\mathscr{O}$ that minimizes $L_{\epsilon}$, the number of iterations needed to reach this accuracy:
\begin{equation}
\min_{\mathscr{O}} \; L_{\epsilon} \quad \text{s.t.} \; |\mathscr{O}| \leq k.
\end{equation}

\citet{jinnai2019planning} prove that this problem is NP-hard, even in deterministic tabular MDPs. They introduce approximation algorithms with provable guarantees, but their results are limited to \textit{point options}---options that initiate and terminate in a single state \citep{jinnai2019planning}.

While the above method minimizes worst-case planning time, Average Options \citep{ivanov2024discovering} focuses instead on minimizing the expected planning time across a distribution of tasks. These tasks share the same transition dynamics but differ in their start and goal states. The idea is to discover options that reduce the expected cost of reaching any state from any other:
\begin{align}
\argmin_{\mathscr{O}} \quad d_\mathscr{O}(G) &= \argmin_{\mathscr{O}} \sum_{s\in\mathscr{S}}\sum_{s'\in\mathscr{S}} d_\mathscr{O}(s, s'),
\end{align}
where $d_\mathscr{O}(s, s')$ is a non-symmetric distance metric (e.g., shortest path length) in the MDP graph augmented with options $\mathscr{O}$; such an augmentation adds edges to the graph while leaving nodes unchanged. Like the worst-case version, this problem is also NP-hard. However, by reducing it to the well-studied $k$-medians with penalties problem in graph theory \citep{meyerson2009minimizing}, \citet{ivanov2024discovering} derive efficient approximation algorithms with bounded suboptimality. Planning can also be sped up using options in the single-task setting: \citet{yiwan2022fasterplanning} present an option discovery algorithm that seeks options that maximize reward---similar to option-critic \citep{Harb2017WhenWI}---but reduces the number of options available at different states to reduce planning time.

\paragraph{Exploration.} In the context of exploration, \citet{pmlr-v97-jinnai19b} formalize the performance criterion of the agent as minimizing the number of steps needed for a policy to visit every state (as a proxy for discovering some unknown reward). Here, \textsc{OptionSpecification} seeks options that minimize the expected cover time of a random walk on the augmented MDP graph. They show that this performance criterion is related to the 
graph-theoretic property of \textit{cover time}, which measures how 
many steps a random walk needs to visit every state in a graph. To 
define cover time, we first need the \textit{hitting time}. Given a 
Markov chain $\{X_t\}$ over states $V$, the hitting time from state 
$i$ to state $j$ is the first time the chain visits $j$ when started 
from $i$:
\begin{equation}
    H_{ij} = \inf\{t \geq 1 : X_t = j \mid X_0 = i\}.
\end{equation}
The cover time from state $i$ is the time needed to visit every 
state at least once:
\begin{equation}
    C_i = \max_{j \in V} H_{ij}.
\end{equation}
Both $H_{ij}$ and $C_i$ are random variables; the quantity of interest 
is the expected cover time $\mathbb{E}[C(G)]$, which bounds how long a random walk takes, in the worst case over starting states, to visit every state in the graph and hence to discover an unknown 
rewarding state~\citep{pmlr-v97-jinnai19b}. \citet{pmlr-v97-jinnai19b} show that the expected cover time $\mathbb{E}[C_i]$ can be most effectively reduced by creating an option that connects the two states that are furthest apart according to the second eigenvector of the graph Laplacian (see Equation~\ref{eqn:laplacian}); this result provides a formal justification for the eigenoption heuristic discussed in Section~\ref{sec:od/spectral}. \citet{pmlr-v97-jinnai19b} also show that finding options that minimize cover time in a graph is NP-hard; but, they provide an approximation algorithm that minimizes an upper bound on the expected cover time. This method was later extended to continuous environments using deep learning-based approximations of the graph Laplacian \citep{jinnai2020Exploration,wu2019laplacian}, further strengthening the connections to the eigenoptions literature \citep{machado2017laplacian,machado2023temporal,Klissarov2023DeepLO} (c.f.\ Section~\ref{sec:od/spectral}).

\paragraph{Credit Assignment.}
As discussed earlier, options can accelerate policy evaluation by enabling value updates that span multiple steps, rather than progressing one step at a time. \citet{bacon16matrix} formalize this intuition using the lens of \emph{matrix splitting}, a technique from numerical linear algebra that speeds up the solution of linear systems. In this case, \textsc{OptionSpecification} seeks a set of options whose induced dynamics minimize the spectral radius of an iteration matrix, thereby maximizing the rate at which iterative policy evaluation converges. In their view, each set of options defines a modified Bellman operator that can be interpreted as a \emph{preconditioned} version of the original policy evaluation problem. Recall that the Bellman expectation equation for a fixed policy \( \pi \) is:
\begin{equation}
    \mathbf{v} = \mathbf{r}_\pi + \gamma \mathbf{P}_\pi \mathbf{v},
\end{equation}
where \( \mathbf{v} \in \mathbb{R}^{|\mathscr{S}|} \) is the value function, \( \mathbf{r}_\pi \) is the expected reward vector, and \( \mathbf{P}_\pi \) is the transition matrix under policy \( \pi \). This is a linear system of the form \( \mathbf{A}\mathbf{v} = \mathbf{b} \), with \( \mathbf{A} = \mathbf{I} - \gamma \mathbf{P}_\pi \) and \( \mathbf{b} = \mathbf{r}_\pi \). Planning with options induces a \emph{matrix splitting} \( \mathbf{A} = \mathbf{M} - \mathbf{N} \) \citep{varga2000matrix}, leading to an iterative update of the form:
\begin{equation}\label{eqn:matrix-splitting}
\mathbf{v}_{k+1} = \mathbf{M}^{-1}\mathbf{N} \mathbf{v}_k + \mathbf{M}^{-1} \mathbf{b}.    
\end{equation}
In this formulation, the matrix \( \mathbf{M} \) reflects the dynamics induced by the options and is chosen to be easy to apply and invert; the remaining part \( \mathbf{N} \) captures what is not directly handled by the options. The matrix \( \mathbf{M}^{-1}\mathbf{N} \) is the \emph{iteration matrix}, as it governs how the current value estimate \( \mathbf{v}_k \) influences the next one \( \mathbf{v}_{k+1} \). This kind of transformation is known as \emph{preconditioning}: a way of rewriting the problem so that the resulting iterative updates converge more quickly. The speed of convergence is governed by the \emph{spectral radius} of the iteration matrix~$\mathbf{M}^{-1}\mathbf{N}$---its largest eigenvalue in absolute value. A smaller spectral radius means that errors shrink faster with each iteration. From this perspective, a good set of options is one whose induced splitting minimizes the spectral radius of~$\mathbf{M}^{-1}\mathbf{N}$, enabling value information to propagate more efficiently. While \citet{bacon16matrix} do not introduce a concrete option discovery algorithm, they offer a powerful design principle: discover options that act as preconditioners for value propagation. This opens the door to leveraging ideas from numerical linear algebra in option discovery.

\paragraph{Transfer.} In the context of transfer, \citet{Solway14Optimal} define the optimal set of options as those that maximize the efficiency with which an agent can learn the optimal policy for other, possibly unseen, sets of tasks. Here, \textsc{OptionSpecification} is an optimization over option sets that maximizes Bayesian model evidence under the distribution of tasks the agent expects to encounter. They show that a hierarchy maximizing model evidence also provably minimizes the geometric mean of the number of samples needed to find the optimal policy for any task in the given task distribution. \citet{brunskill2014pac} consider a similar formulation of the option transfer problem: given interaction data from a set of tasks, how can an agent learn options that minimize the sample complexity of learning in a future stream of tasks? They find that this problem is at least as hard as the set cover problem in operations research, and is thus also NP-hard. They use a greedy approximation algorithm for option discovery and evaluate it empirically in a tabular MDP.

\paragraph{Opportunities for Research.}
\begin{itemize}
    \item \textit{Guarantees in more general settings.} The papers discussed in this section emphasize the importance of formally stating the objective of option discovery and relating that to the overall objectives of the agent. However, this research is still nascent, and more work exploring this subject is needed. For instance, can we develop formal algorithms that bound planning time without needing the assumption of point options~\citep{jinnai2019planning}? Can we bound planning time or cover time when using function approximation? Although \citet{brunskill2014pac} derive an algorithm to minimize sample complexity during transfer, the greedy approximation algorithm they present does not bound sample complexity; future work could extend their theoretical results to bound the performance of the greedy approximation algorithm.
    
    \item \textit{Online and incremental option discovery with guarantees.} Most algorithms discussed in this section solve a combinatorial optimization offline, assuming access to a known MDP graph or to enough interaction data to construct one~\citep{jinnai2019planning,ivanov2024discovering}. A few methods do construct options incrementally---for instance, \citet{pmlr-v97-jinnai19b} add options one at a time, each chosen to further reduce an upper bound on the expected cover time---but this incremental construction still operates on a fully known state-space graph. Extensions to larger or continuous state spaces do exist: \citet{jinnai2020Exploration} scale covering options using deep learning-based approximations of the graph Laplacian, and \citet{ivanov2024discovering} adapt their planning-time objective to continuous domains by first discretizing the state space offline. In both cases, however, the formal guarantees of the tabular setting either do not carry over or apply only to the discretized proxy of the underlying problem rather than to the true MDP. An important open question is therefore whether any of the objectives considered in this section---planning time, cover time, or sample complexity during transfer---admit algorithms that discover and refine options as the agent collects data \textit{while preserving} formal guarantees on the corresponding agent-level objective. 
    
    \item \textit{Stronger and more practically relevant guarantees.} The guarantees currently available in this literature are typically worst-case over all MDPs of a given size, and the algorithms producing them usually scale with the the size of the state space \citep{ivanov2024discovering}. Can we obtain \textit{instance-dependent} approximation ratios that adapt to structural properties of a given MDP---such as its diameter, bottleneck structure, or spectral gap---rather than worst-case ratios over all MDPs? \citet{brunskill2014pac} provide PAC-style sample-complexity bounds for option discovery in the transfer setting, but analogous guarantees are missing for exploration. \citet{fruit2017exploration} derive regret bounds for exploring with a \textit{fixed} option set; can analogous guarantees be obtained when the options themselves are being discovered online, as they are for flat exploration algorithms~\citep{kearns2002near,Amortila24}? And on the scaling side, can we characterize how the quality of discovered options degrades with the expressivity of a function approximator, the size of the option set, or the number of tasks in a transfer distribution?
\end{itemize}

\subsubsection{Relationship to the Algorithmic Template}

In the language of Algorithm~\ref{alg:hrl_generic}, the methods in this 
section all formalize \textsc{OptionSpecification} as an explicit 
optimization problem over a well-defined agent-level objective, rather 
than relying on a proxy. The output~$\Omega$ is a set of subgoal 
states and corresponding option reward functions, just as in bottleneck 
discovery (Algorithm~\ref{alg:bottleneck}), but the subgoals are selected by optimizing an agent-level 
criterion rather than a graph-theoretic heuristic. 
\textsc{OptionLearning} and \textsc{PolicyImprovement} are typically 
standard and receive less attention in this literature.

Two structural features distinguish these methods from those discussed 
earlier. First, the optimization that determines~$\Omega$ considers 
interactions between options: the value of including any single option 
objective in~$\Omega$ depends on what other objectives are already 
present. In bottleneck methods (Section~\ref{sec:od/graph-based}), 
each element of~$\Omega$ is typically scored independently---for example, the betweenness 
score of a state does not depend on what other options exist. Here, by 
contrast, adding an edge to the graph that connects two distant states 
is valuable only if no existing option already provides a similar 
shortcut. This set-level dependence is precisely what makes these 
optimization problems NP-hard and necessitates approximation 
algorithms. Second, many approaches in this section adopt a separated 
pipeline: the optimization over~$\Omega$ is solved once given a model 
or dataset, and the resulting subgoals are fixed before deployment. In 
such cases, \textsc{OptionSpecification} runs once, which may limit 
adaptability when new regions of the state space are discovered. 
However, the cover-time minimization algorithm of \citet{pmlr-v97-jinnai19b} has also been applied in an online setting, where \textsc{OptionSpecification} is re-invoked periodically as the agent gathers more experience and refines its estimate of the transition graph, interleaving \textsc{OptionSpecification} with \textsc{OptionLearning} and \textsc{Interact}.

\section{Discovery through Offline Datasets}
\label{offline}
Offline RL (or batch RL) aims to learn policies from pre-collected datasets, avoiding active data collection. This enables scalable deployment in domains such as robotics, autonomous driving, education, and healthcare~\citep{levine2020offline}, where interaction data can be difficult to obtain. Similarly, \textit{offline skill discovery} leverages these datasets to extract temporally abstract behaviours that can later serve as high-level primitives (in either offline or online RL) to accelerate learning.



In this section, to facilitate the discovery of skills, we assume access to an offline dataset 
$\mathscr{D}=\{\tau_i\}^N_{i=1}$,  where $\tau_i$ is a trajectory generated through interaction with the environment.
When all components are observed, a trajectory can be written as  
$\tau_i=(s_t^i,a_t^i,r_t^i)_{t=1}^{T}$. Depending on the method and dataset, however, a trajectory may contain only a subset of these components. The trajectories may be expert demonstrations or data collected by one or more behaviour policies. For example, many methods in Section~\ref{sec:maxlike} do not require rewards, and some do not require actions~\citep{kim2019variational}.
A related line of work learns reusable representations of environment dynamics, often from offline data, without explicitly defining temporally extended skills. One important family is based on successor representations and successor features. These representations separate expected future state occupancies or feature accumulations from the task-specific reward, allowing previously learned knowledge to be reused when the reward changes~\citep{dayan93successor,barreto2017successor,Barreto2020FastRL}. Universal successor features~\citep{ma2020universal}, forward-backward representations~\citep{touati2023does,tirinzoni2025zeroshot}, and successor-measure-based auxiliary tasks~\citep{farebrother2023protovalue} develop this principle in different ways. Other approaches capture reusable structure using random features~\citep{chen2023selfsupervised}, Hilbert-space representations~\citep{hilbert}, or Laplacian representations~\citep{shehmar2026alps,chandrasekar2026lk}. Because these methods do not necessarily define temporally extended skills, we do not discuss them in detail.\looseness=-1

\subsection{Variational Inference of Skill Latents}
\label{sec:maxlike}
Offline datasets often contain recurring, temporally extended behaviours without labels for either the behaviours or their boundaries. For example, a trajectory from a robotic manipulation task may include reaching for an object, grasping it, and transporting it to a target location without indicating where one behaviour ends and another begins. 
To uncover this structure, variational-inference methods use an encoder to map trajectory segments to latent codes and a decoder to reconstruct the observed trajectories. Following common usage in the literature, we refer to each latent code as a \emph{skill} throughout this section.\footnote{Under the formalism of Definition~\ref{def:option_id} and Section~\ref{sec:related_formalism}, the latent code is an option identifier: it indexes the corresponding option rather than denoting the complete option itself.}


These methods usually rely on \textit{unlabeled experiences}, $\tau = (s_t, a_t)_{t=1}^T$---that is, data collected without explicit reward feedback---and in some cases, even excludes actions \citep{kim2019variational}.
This is often referred to as \textit{unsupervised skill discovery} \citep{eysenbach2019diversity}.
To accommodate different formulations in the literature, we associate each trajectory $\tau$ with a latent skill sequence $\zeta$:
\begin{equation}
\label{eq:skill_boundary}
  \zeta=(z_t,b_t)_{t=1}^{T},
  \qquad 
  z_t\in\mathbb{R}^{d},\;
  b_t\in\{0,1\},
\end{equation}
where $z_t$ represents the skill active at time $t$ and
$b_t$ is a boundary signal that indicates when a skill starts or ends, i.e.\ the analogue of an option-termination
signal.\footnote{Both $z$ and $b$ can be represented in different forms across methods. For example, \citet{kipf2019compile} uses $b_m\in\mathbb{Z}$ to denote the position of the $m$-th boundary, while some methods omit explicit boundary variables~\citep{ajay2020opal}.}

Let $p_\phi(\tau,\zeta)$ denote a generative model over a trajectory $\tau$ and its latent skill sequence $\zeta$, parameterized by $\phi$. The model is learned by maximizing the marginal log-likelihood of the trajectories in the offline dataset $\mathscr{D}$:
\begin{equation}\label{eq:marglik}
  J(\phi)=
  \mathbb{E}_{\tau\sim\mathscr{D}}
  \bigl[\log p_{\phi}(\tau)\bigr].
\end{equation}
Here the term $p_{\phi}(\tau)=\int p_{\phi}(\tau,\zeta)\,d\zeta$ has already
\emph{marginalized} the latent skill sequence $\zeta$, so maximizing
$J(\phi)$ encourages the model to explain the observed
trajectories without fixing any particular skills in advance.
Because marginalizing over all possible $\zeta$ is usually intractable, variational methods introduce an approximate posterior $q_\phi(\zeta\mid\tau)$ and apply Jensen's inequality to derive the \emph{evidence lower bound} (ELBO), a tractable lower bound on $\log p_\theta(\tau)$:

\begin{align}
\log p_{\phi}(\tau)
  &=\log\!\int q_{\phi}(\zeta\mid\tau)
              \frac{p_{\phi}(\tau,\zeta)}{q_{\phi}(\zeta\mid\tau)}
              d\zeta \nonumber \\
  &\ge \mathbb{E}_{q_{\phi}(\zeta\mid\tau)}
         \!\bigl[\log p_{\phi}(\tau,\zeta)-\log q_{\phi}(\zeta\mid\tau)\bigr] \nonumber \\
  &=\mathbb{E}_{q_{\phi}}
        \!\bigl[\log p_{\phi}(\tau\mid\zeta)\bigr]
    -D_{\mathrm{KL}}\!\bigl(q_{\phi}(\zeta\mid\tau)\,\|\,p_{\phi}(\zeta)\bigr).
\end{align}

Averaging over $\tau\!\sim\!\mathscr{D}$, and introducing the
$\beta$-weight\footnote{We slightly abuse the notation here, where usually in this paper $\beta$ refers to the termination function.}~\citep{higgins2016beta} which balances reconstruction and regularization terms yields the training objective:
\begin{equation}\label{eq:elbo}
J_{\mathrm{ELBO}}(\phi)=
\underbrace{\mathbb{E}_{\tau\sim\mathscr{D},
    \zeta\sim q_{\phi}(\zeta\mid\tau)}
      \bigl[\log p_{\phi}(\tau\mid\zeta)\bigr]
  }_{\text{reconstruction}}
\;-\;
\beta\,
\mathbb{E}_{\tau\sim\mathscr{D}}\,
  \underbrace{
    D_{\mathrm{KL}}\!\bigl(
       q_{\phi}(\zeta\mid\tau)\,\|\,p_{\phi}(\zeta)
    \bigr)
  }_{\text{regularization}},
\end{equation}
which is \emph{maximized} with respect to $\phi$. 
The first term encourages \(\zeta\) to capture the latent skill structure that makes \(\tau\) likely under
\(p_\phi(\tau\mid\zeta)\).
The second term {\em regularizes} those encodings toward the
prior, encouraging the emergence of a compact skill space.
In practice, three distinct
models are employed whose parameters are jointly denoted by
$\phi$:
\begin{itemize}
  \item \textbf{Prior} $p_{\phi}(\zeta)$: defines a prior distribution over latent skill sequences. A common choice is a fixed, factorized prior (e.g., unit Gaussian for $z_t$ and Bernoulli for $b_t$), but the prior can instead be endowed with learnable parameters and conditioned on the current state, yielding $p_{\phi}(\zeta\mid s)$~\citep{ajay2020opal,nam2022metaskills}. This can also serve as a prior on the policy over skills (i.e., a high-level policy), $\mu_\kappa(z\mid s)$.
\item \textbf{Decoder} $p_{\phi}(\tau \mid \zeta)$: models the probability of a trajectory conditioned on a given skill sequence. In many works, this likelihood is factorized over timesteps and parameterized as a skill-conditioned policy,
\begin{equation}
p(\tau \mid \zeta)
= p(s_1)\prod_{t=1}^T
p(s_{t+1} \mid s_t, a_t)\,
\pi_\theta(a_t \mid s_t, \zeta)
\;\propto\;
\prod_{t=1}^{T} \pi_{\theta}(a_t \mid s_t, \zeta),
\end{equation}
so that the decoder can reconstruct only the action sequence conditioned on states and skills~\citep{kipf2019compile,ajay2020opal,pertsch2021accelerating,nam2022metaskills}. Under this parameterization, Equation~\ref{eq:elbo} becomes:
\begin{equation}\label{eq:elbo-policy}
J_{\mathrm{ELBO}}(\phi,\theta)=
\underbrace{\mathbb{E}_{\tau\sim\mathscr{D},\;
    \zeta\sim q_{\phi}(\zeta\mid\tau)}
      \Bigl[\sum_{t=1}^{T}
        \log\pi_{\theta}\bigl(a_t \mid s_t,\zeta\bigr)\Bigr]
  }_{\text{action-reconstruction}}
\;-\;
\beta\,
\mathbb{E}_{\tau\sim\mathscr{D}}\,
  \underbrace{
    D_{\mathrm{KL}}\!\bigl(
       q_{\phi}(\zeta\mid\tau)\,\|\,p_{\phi}(\zeta)
    \bigr)
  }_{\text{regularization}}.
\end{equation}

\item \textbf{Encoder}
        $q_{\phi}(\zeta\mid\tau)$: given an
        observed trajectory, returns a distribution over the
        skill sequence that likely produced it.
\end{itemize}


Variational-autoencoder (VAE)-based methods~\citep{DBLP:journals/corr/KingmaW13} are the main focus of this subsection: they train an encoder \(q_\phi(\zeta\mid\tau)\) that predicts a distribution over \(\zeta\) from each trajectory and optimize the ELBO in Equation~\ref{eq:elbo}. Related latent-variable methods share the same high-level goal of inferring a latent sequence \(\zeta\) that explains the observed trajectories, but differ in how \(\zeta\) is inferred and which training objective is used.
Expectation-Gradient methods~\citep{fox2017multi,krishnan2017ddco} instead begin with unsegmented trajectories in which the active option and the option boundaries are unknown. At each timestep, they estimate a probability distribution over the active option, together with the probability that the previous option terminated and a new option was selected. They then use these probabilities to update the option policies, termination functions, and high-level policy so that the model assigns higher likelihood to the observed trajectories.
Adversarial approaches such as Directed-Info GAIL~\citep{sharma2018directed} train a discriminator to distinguish policy-generated state–action pairs from demonstration, while using an auxiliary objective to encourage different skills to induce distinguishable behaviours.

The learned skills naturally support a hierarchy. 
In such works, a low‑level controller, \(\pi_\theta(a_t \mid s_t, z_t)\), is typically trained offline via behavioural cloning to execute any given skill, while a high‑level policy, \(\mu_\kappa(z_t \mid s_t)\), is subsequently optimized, either online~\citep{pertsch2021accelerating,nam2022metaskills,salter2022mo2}, 
or with offline RL~\citep{ajay2020opal}.
Learned skills can also augment the primitive action space, expanding the agent’s control repertoire~\citep{fox2017multi, kipf2019compile, jiang2022learning}, or be used to construct pseudo-reward signals by measuring the distance between each transition's latent embedding and calibrated expert behaviour~\citep{liu2023clue}.

Beyond pure likelihood maximization, it is also common to add a \textbf{compression} regularizer grounded in the \emph{minimum description length} (MDL) principle~\citep{rissanen1978modeling,Thrun1994FindingSI}.  MDL prefers the model that can be transmitted with the fewest total bits of (i) the model parameters and (ii) the data encoded through that model.  Viewing the latent codes as skill identifiers and boundaries (Equation~\ref{eq:skill_boundary}), the model parameters are the decoder (and prior if parameterized), and the data are the offline trajectories; hence, a concise skill set shortens the overall description length.  \looseness=-1

The \emph{bits-back} coding argument~\citep{NIPS1993_9e3cfc48,1310354,zhang2021minimum} shows that maximizing the ELBO  (Equation~\ref{eq:elbo}) approximately minimizes the description length, but with an ill-chosen prior $p(\zeta)$, the optimum can collapse to a degenerate representation (e.g., a single skill encoding that simply mirrors the observations).  To avoid this, LOVE~\citep{jiang2022learning} augments the ELBO with an MDL-inspired information-cost term that explicitly penalizes skills increasing the expected code length of transmitting trajectories, yielding a representation that is both \emph{informative} (high ELBO) and \emph{economical} (low description length):

\begin{equation}
\label{eq:dl}
   \min_{\phi}\  L_{\mathrm{DL}}(\phi)
\quad
\text{s.t.}
\quad
J_{\mathrm{ELBO}}(\phi) \;\ge\; C, 
\end{equation}
\begin{equation}
L_{\mathrm{DL}}(\phi)
~=
\mathbb{E}_{\tau \,\sim\, \mathscr{D},\;\{b_t,z_t\}\,\sim\,q_\phi(\cdot\,\mid\,\tau)}
\Bigl[
-\;\sum_{t=1}^T
b_t\,\log\,p_{z}^{*}\bigl(z_t;\phi\bigr)
\Bigr],
\end{equation}
where \(b_t = 1\) indicates the start of a new skill at time \(t\), and $C$ is a constant. The optimal prior on $z$, \(p_{z}^{*}\), that minimizes the expected description length, is defined by:
\begin{equation}
p_{z}^{*}(z;\phi)
~=
\frac{
\mathbb{E}_{\tau \,\sim\, \mathscr{D},\;\{b_t,z_t\}\,\sim\,q_\phi(\cdot\,\mid\,\tau)}
\Bigl[
\sum_{t=1}^T
b_t\,\mathbbm{1}_{\{z_t=z\}}
\Bigr]
}{
\mathbb{E}_{\tau \,\sim\, \mathscr{D},\;\{b_t,z_t\}\,\sim\,q_\phi(\cdot\,\mid\,\tau)}
\Bigl[
\sum_{t=1}^T
b_t
\Bigr]
},
\end{equation}
with $\mathbbm{1}$
denoting the indicator function.
Intuitively, $\mathcal L_{\mathrm{DL}}$
penalizes having too many skill boundaries and distinct skill choices, driving the method toward a concise skill decomposition, and longer skills encompassing common structures are generally favored to avoid the degenerate solution where each skill represents a single action. 
\citet{salter2022mo2} instead leverages the concept of \textit{bottleneck} option by introducing a \textit{predictability} objective, encouraging option-level transitions to be predictable. The authors show that maximizing this predictability reduces the conditional entropy and thus the optimal code length, and this objective is equivalent to applying the MDL principle.

\citet{vlastelica2023diverse} cast offline skill discovery as \textbf{empowerment maximization}, presented in Section \ref{sec:empowerment_skill_discovery},  under an imitation constraint.  
To ensure that each skill remains faithful to the offline demonstrations, they constrain the divergence between the induced state occupancy and the state occupancy \(d_E\) from a skill-independent expert dataset, resulting in a constrained optimization problem:

\begin{equation}
\max_{\{\pi_z\},\,q_\phi}
\;
\mathbb{E}_{z\sim p(z),s\sim d_{\pi}}
[\log q_\phi(z\mid s)]
\quad
\text{s.t.}\;
D_{\mathrm{KL}}\bigl(d_{\pi}\,\|\,d_E\bigr)\le\varepsilon,
\;\forall z.
\label{eq:emp_offline}
\end{equation}
Here, \(d_{\pi}\) denotes the state-occupancy measure of the skill-conditioned policy, estimated in the offline setting via density-based learning.
The term \(q_\phi\) represents a skill discriminator that tightens a variational lower bound on the mutual information between skills and states.
It simultaneously \emph{diversifies} behaviours by maximizing the lower bound on the mutual information between states and skills, and \emph{regularizes} them by penalizing departures from the expert state distribution $d_E$.

\begin{algorithm}[t]
\caption{Variational Inference of Skill Latents: Instantiation of 
Algorithm~\ref{alg:hrl_generic}.}
\label{alg:offline_skill_latents}
\begin{algorithmic}[1]
\Require Offline dataset $\mathscr{D}$, prior $p_{\phi}(\zeta)$.
\State Initialize model parameters $\phi,\theta$.
\Statex

\Procedure{OptionSpecification}{$\mathscr{O}, \mathscr{D}$}
    \State Define (the structure of) the latent skill variables $\zeta = (z_t,b_t)_{t=1}^{T}$.
    \State Define the objective $\Omega = J_{\mathrm{ELBO}}(\phi,\theta)$. \Statex \Comment{Eq.~\ref{eq:elbo}, can also include MDL or other objectives, from Eq.~\ref{eq:dl}--Eq.~\ref{eq:emp_offline}.}
    \State \Return $\Omega$
    \Comment{Defines what latent skills should explain}
\EndProcedure
\Statex
\Procedure{OptionLearning}{$\mathscr{O}, \mathscr{D}, \Omega$}
    \State Sample $\tau \sim \mathscr{D}$ and infer $\zeta \sim q_{\phi}(\zeta\mid\tau)$.
    \State Update $\phi,\theta$ to maximize $\Omega$.
    \State Construct $\mathscr{O}$ implicitly from learned skills $z$ and boundary signals $b$.
    \Statex \Comment{Each $z$ induces an option via $\pi_\theta(a\mid s,z)$ with termination $\beta(s,z)=\mathbbm{1}(b=1)$}
    \State \Return $\mathscr{O}$
\EndProcedure
\end{algorithmic}
\end{algorithm}

\subsubsection{Relationship to the Algorithmic Template}
When instantiating Algorithm~\ref{alg:offline_skill_latents}, \textsc{OptionSpecification} typically runs only once and $\Omega$ is fixed thereafter. Design Principle~1 in Section~\ref{sec:option-discovery} does not typically hold here; instead, a single shared objective is used, and the method performs single-level optimization, simultaneously inferring $z$ and $b$ and learning the skill-conditioned policy within \textsc{OptionLearning}.

\subsubsection{Benefits and Opportunities}

\paragraph{Credit Assignment.}
By extracting hierarchical structure from offline datasets, agents can break complex trajectories into more manageable subgoals.
This segmentation makes it easier to understand why certain results occur, as it allows each outcome---such as achieving a subgoal---to be more directly linked to the specific actions and conditions that produced it. 
In other words, by focusing on a sequence of subgoals rather than the full sequence of primitive actions, the learning algorithm can more easily attribute success or failure to specific decisions or events. 
For example, \citet{kipf2019compile} tackle credit assignment in maze-like environments with delayed feedback. Their method infers segment boundaries \( q(b_i \mid s, a) \), with \( b_i \in [1, T+1] \) functioning similarly to option termination functions, and encodes each segment using \( q(z_i \mid  s, a) \) as latent skill (subgoal) descriptors. This segmentation captures subgoal structure, facilitating effective credit assignment across subgoals when applying the skills in sparse-reward settings.
Similarly, \citet{kim2019variational} improve credit assignment in goal-oriented navigation tasks by decomposing action-free trajectories into subsequences by inferring skill descriptors $z_t$ and binary termination signals $b_t\in\{0,1\}$. 

\paragraph{Transfer.} 
By discovering skills from offline datasets, agents develop a foundational set of versatile competencies. These competencies can then be transferred to new tasks with minimal adjustment.
\citet{pertsch2021accelerating} show promising results on transferring skills obtained in the offline dataset to more complex simulated robotic tasks unseen in the dataset (e.g., maze navigation with larger maps in evaluation). Similarly, \citet{jiang2022learning} and \citet{salter2022mo2} show that by optimizing a compression objective, in addition to the reconstruction one, the discovered skills help transfer across multiple tasks.
\citet{nam2022metaskills} demonstrate that by meta-training a high-level policy, $\pi_\theta(z_t \mid s_t, e)$, where $e$ is a task encoding, and executing a low-level policy, $\pi_\theta(a_t \mid s_t, z_t)$, which is learned via behavioural cloning, the agent can solve a wide range of new tasks in a meta-RL setting.

\paragraph{Exploration.} 
When reusing the offline discovered skills for online interaction, this can reduce the difficulty of exploration since the agent can quickly apply well-tested, pre-learned behaviours rather than learning them through trial-and-error in real-time.
\citet{fox2017multi} show promising exploration results in a simple four-room domain by augmenting the action space with discovered parameterized options.
\citet{salter2022mo2} show that the learned temporally compressed bottleneck options are beneficial for exploration in maze-like environments with delayed rewards.

\paragraph{Avoiding Distributional Shift.} 
In offline RL, a learned policy may select actions with little or no support in the fixed dataset, for which value estimates can be unreliable. 
\citet{ajay2020opal} leverage offline skill discovery to mitigate this issue. Their approach encodes short trajectories (e.g., every $K$ steps) from the dataset into a skill code $z$. By maximizing the log-likelihood of actions in trajectories $\tau$, conditioned on states $s_t$ and skill descriptors $z$, the method captures recurring behaviours present in the data. The offline dataset can then be enhanced using $z$, and a high-level policy, $\pi_\theta(z\mid s)$, can be derived by off-the-shelf offline RL algorithms. The authors show that such a temporal structure reduces compounding errors for extrapolating out-of-distribution actions in offline RL.

\paragraph{Opportunities for Research.}
\begin{itemize}
    \item \textit{Optimization challenges.} Evident in some studies~\citep{jiang2022learning}, optimization challenges can lead to degraded skill quality as poor local optima corresponding to degenerate skill decompositions can be empirically hard to escape.
Additionally, the under-utilization of reward signals in existing datasets creates an opportunity to further refine learned skills, and incorporating offline RL methods---rather than relying solely on reconstruction-based approaches---into HRL may unlock greater performance gains, as \citet{hu2023provably} provide  positive theoretical results on sample efficiency.
\item \textit{Broader scope of test environments.} Additionally, many methods in this field primarily validate their concepts on simulated robotic navigation tasks, which typically involve deterministic transitions and rewards~\citep{gao2024act}, and often favor specific inductive biases.
A natural extension would be scaling this paradigm to real-world, image-based tasks, or other practical applications with different properties.
\end{itemize}

\subsection{Hindsight Subgoal Relabeling}
\label{sec:subgoalgen}
In Section \ref{sec:maxlike}, we discussed the methods that infer the skill descriptors, usually characterized by latent codes or skill embeddings. 
In this section, we explore methods that identify and relabel subgoals within an offline dataset, effectively leveraging existing trajectories to learn how to achieve subgoals \citep{kaelbling1993learning}.
Offline experiences offer valuable insights into identifying subgoals, which can be viewed as milestones or waypoints for accomplishing a task \citep{gupta2019relay,park2024hiql}, or abstracting bottleneck states to make a good partition of the state space \citep{paul2019learning}.
This is conceptually similar to hindsight experience replay \citep{Andrychowicz2017HindsightER} we discussed in Section \ref{sec:curriculum}, which relabels the final or intermediate states reached in a trajectory as if they were the intended goals.



In an exemplar work by \citet{paul2019learning}, a reward-free trajectory dataset,
\(\mathscr{D}=\bigl\{(s^{i}_{1},a^{i}_{1},\dots,s^{i}_{T})\bigr\}_{i=1}^{N}\), is
segmented into an \emph{ordered} list of \(n_g\) disjoint partitions,
\(G=\{1,\dots,n_g\}\), that serve as subgoals. The label $g_t^i=j$ indicates that state $s_t^i$ is assigned to the $j$-th subgoal region.

\smallskip
\textit{Initial relabeling.}
Because the demonstrations do not provide these assignments, the method begins with a simple initial guess: each trajectory is divided into $n_g$ consecutive portions of approximately equal duration, and the $j$-th portion is assigned label $j$:
\begin{equation}
g_t^{i}=j
\quad\text{if and only if}\quad
\left\lfloor\frac{(j-1)T}{n_g}\right\rfloor < t
\le
\left\lfloor\frac{j T}{n_g}\right\rfloor,
\qquad j\in G.
\end{equation}

\smallskip
\textit{Iterative refinement.} Repeat the following two steps until the label change falls below a threshold. Alternating these steps until
convergence yields a classifier $\mu_\kappa$ that both partitions the state space and
respects the required ordering.
Such a $\mu_\kappa(g\mid s)$ can provide information on whether a state is a bottleneck.
\begin{itemize}
\item \emph{Learning step}: fit a classifier
\(\mu_\kappa(g\mid s)\) by cross-entropy on the current
labels:
\begin{equation}
L(\kappa)
\;=\;
\mathbb{E}_{(s,g)\sim \mathscr{D}}\bigl[-\log \mu_{\kappa}(g\mid s)\bigr].
\end{equation}
\item \emph{Inference step}: enforce the trajectory order
\(1\!\prec\!2\!\prec\!\dots\!\prec\!n_g\) with Dynamic Time Warping\footnote{An algorithm that aligns two sequences by allowing non-uniform time shifts, so that similar patterns are matched even if they occur at different timesteps or speeds.}~\citep{Müller2007}
over the sequence
\(\bigl(\mu_\kappa(g=1\mid s_t^i),\ldots,\mu_\kappa(g=n_g\mid s_t^i)\bigr)\).
\end{itemize}

\smallskip
\textit{Potential function and intrinsic reward.}
The most probable class defines a potential
\(\Phi_\kappa(s)=\arg\max_{g}\mu_\kappa(g\mid s)\), representing the most likely subgoal index associated with state $s$. Based on this potential, an intrinsic reward, $r'$, that measures \textit{progress} toward successive subgoals, is defined as:
\begin{equation}
r'(s,a,s')\;=\;\gamma\,\Phi_\kappa(s')-\Phi_\kappa(s),        \label{eq:paul_intrinsic}
\end{equation}
where $\gamma$ is the discount factor of the MDP.

\textit{Learning schedule.} The policy \(\pi_\theta\) is first initialized through behaviour cloning, after which online RL proceeds with the augmented reward \(r + r'\). This phase exploits subgoal guidance to supply dense progress signals without additional expert interaction and still preserves the original optimum \citep{paul2019learning}.


As another example, Relay Policy Learning (RPL)~\citep{gupta2019relay}
relabels all demonstration trajectories
$\{\tau^i\}_{i=1}^{N}$, where
$\tau^i=(s_1^i,a_1^i,\dots,s_{T}^i)$, to construct a low-level dataset
$\mathscr{D}_{\ell}$ and a high-level dataset $\mathscr{D}_{h}$
(represented jointly as $\mathscr{D}$ in
Algorithm~\ref{alg:hindsight_subgoal}):
\begin{align}
\mathscr{D}_{\ell}
&=\bigl\{\, (s_t,a_t,g_{\ell}) \;\bigm|\;
      g_{\ell}=s_{t+w},\;
      0\le t<T,\;
      1\le w\le W_{\ell},\;
      t+w\le T \,\bigr\},\\[2pt]
\mathscr{D}_{h}
&=\bigl\{\, (s_t,g_{\ell},g_{h}) \bigm|
      g_{\ell}=s_{t+\min(W_{\ell},w)},\;
      g_{h}=s_{t+w},\;
      0\le t<T,\;
      1\le w\le W_{h},\;
      t+w\le T \,\bigr\}.
\end{align}
Here, $w$ is the number of timesteps between the current state and a selected
future state, while $W_{\ell}$ and $W_h$ are the maximum look-ahead windows
used to construct low- and high-level training examples, respectively. RPL uses $W_{\ell}=30$ and $W_h=260$ in all experiments.
Thus, $\mathscr{D}_{\ell}$ offers \emph{short-horizon}
examples (subgoal horizon $\le W_{\ell}$) for training a
low-level controller $\pi_{\theta}(a\mid s,g_{\ell})$ to reach nearby
states, whereas $\mathscr{D}_{h}$ pairs each \textit{long-horizon} target
$g_{h}$ (up to $W_{h}$ steps away) with a feasible intermediate subgoal
$g_{\ell}$, enabling hierarchical planning by a high-level goal-setter
$\mu_{\kappa}(g_{\ell}\mid s,g_{h})$.
The imitation objective is:
\[
\max_{\kappa,\theta}\;
\mathbb{E}_{(s,a,g_\ell)\sim\mathscr{D}_{\ell}}
        \bigl[\log\pi_{\theta}(a\mid s,g_\ell)\bigr]
\;+\;
\mathbb{E}_{(s,g_{\ell}',g_{h})\sim\mathscr{D}_{h}}
        \bigl[\log\mu_{\kappa}(g_{\ell}'\mid s,g_{h})\bigr].
\]
During execution, every $W_{\ell}$ steps the
high-level policy samples a new subgoal
$g_{\ell}\sim\mu_{\kappa}(\cdot\mid s,g_{h})$ and the low-level
controller tracks it step-by-step until the next subgoal is issued.

\begin{algorithm}[t]
\caption{Hindsight Subgoal Relabeling: Instantiation of 
Algorithm~\ref{alg:hrl_generic}.}
\label{alg:hindsight_subgoal}
\begin{algorithmic}[1]
\Require Offline dataset $\mathscr{D}$, subgoal relabeling rule $\mathbb{S}$.
\State Initialize option set $\mathscr{O}$, 
low-level policy $\pi_{\theta}$.
\Statex
\Procedure{Interact}{$E, \mathscr{O}, \mu, \mathscr{D}$}
    \State Sample trajectory $\tau \sim \mathscr{D}$.
    \State Relabel $\tau$ with subgoals using $\mathbb{S}$.
    \State Construct relabeled dataset $\mathscr{D}'$.
    \State \Return $\mathscr{D}'$
    \Comment{Offline setting: no new environment interaction}
\EndProcedure
\Statex
\Procedure{OptionSpecification}{$\mathscr{O}, \mathscr{D}$}
    \State Define subgoals $g$ (from relabeled states in $\mathscr{D}$ or from a learned classifier).
    \State Define subgoal-conditioned objectives $\Omega = \{r^g\}$.
    \Statex\Comment{e.g., goal-reaching reward or intrinsic reward from subgoal progress}
    \State \Return $\Omega$
\EndProcedure
\Statex
\Procedure{OptionLearning}{$\mathscr{O}, \mathscr{D}, \Omega$}
    \State Update option policy $\pi_{\theta}(a\mid s,g)$ to maximize $r^g\in \Omega$.
    \State Construct options $\mathscr{O}$ from learned goal-conditioned behaviours.
    \Statex \Comment{$\forall s,\ \mathscr{I}(s)=1$; $\beta(s)=\mathbbm{1}\{s \text{ satisfies } g\}$}
    \State \Return $\mathscr{O}$
\EndProcedure
\end{algorithmic}
\end{algorithm}

\subsubsection{Relationship to the Algorithmic Template}
Algorithm~\ref{alg:hindsight_subgoal} has a similar algorithmic structure as Algorithm~\ref{alg:her_inst} for hindsight experience replay but differs in design details. Same as described in Section~\ref{sec:template-curriculum}, we choose to put the relabeling process in \textsc{Interact} but conceptually it can be regarded as \textsc{OptionSpecification}. Note that the work by \citet{paul2019learning} can be interpreted as an instance of option discovery where subgoal indices $g$, inferred via the classifier $\mu_{\kappa}(g\mid s)$, serve as option identifiers, inducing implicit options through the potential function $\Phi_{\kappa}(s)$ and the resulting intrinsic reward, rather than through explicit goal-conditioned policies as in~\citet{gupta2019relay}, which slightly differs from the description in Algorithm~\ref{alg:hindsight_subgoal}, and its \textsc{OptionSpecification} is similar to Algorithm~\ref{alg:empowerment}.

\subsubsection{Benefits and Opportunities}
\paragraph{Credit Assignment.}
Subgoal relabeling decomposes complex tasks into manageable intermediate objectives by highlighting which actions contribute to reaching the final goal.

\citet{paul2019learning} address credit assignment by constructing an intrinsic reward based on a subgoal policy $\mu_\kappa(g\mid s)$, which identifies a state's progress toward a final goal. This classifier is trained via an EM-style procedure that enforces an ordering constraint over subgoal indices, ensuring that states later in a demonstration receive higher labels. The resulting potential function, $\Phi_\kappa(s) = \arg\max_g \mu_\kappa(g\mid s)$, induces a shaped reward (Equation~\ref{eq:paul_intrinsic}) which provides dense feedback aligned with behavioural progress. This intrinsic signal facilitates temporal credit assignment by rewarding transitions that advance the agent through the learned subgoal structure, even when extrinsic rewards are sparse or delayed.

Alternatively, RPL~\citep{gupta2019relay} addresses the credit assignment challenge from sparse, delayed rewards by relabeling demonstrations with overlapping sliding-window subgoals. It trains a goal-conditioned low-level policy to reach short-horizon targets, while a high-level policy selects subgoals. This hierarchical structure ensures that each low-level episode ends with an intrinsic success signal. As a result, external rewards propagate after at most one window. This transforms the long-horizon problem into a sequence of locally supervised updates, enabling faster and more stable credit assignment than flat or single-level baselines.\looseness=-1
Similarly, \citet{park2024hiql} relabel subgoals as the state \(s_{t+k}\) that lies exactly \(k\) steps ahead of the current state \(s_t\).
A high-level policy, \(\mu_{\kappa}\bigl(s_{t+k}\mid s_t, g\bigr)\), proposes such waypoints conditioned on the ultimate goal \(g\),
while a low-level policy, \(\pi_{\theta}\bigl(a_t\mid s_t, s_{t+k}\bigr)\), outputs primitive actions that move the agent toward the subgoal.
Both policies are optimized via a shared goal-conditioned value function \(V_{\phi}\):
\(\mu_{\kappa}\) maximizes \(V_{\phi}(s_{t+k}, g)\), whereas \(\pi_{\theta}\) maximizes \(V_{\phi}(s_{t+1}, s_{t+k})\).
Specifically, \(\mu_{\kappa}\) is trained to choose subgoals \(s_{t+k}\) that maximize
\(V_{\phi}(s_{t+k}, g)\), while \(\pi_{\theta}\) is trained to select actions that make the next state \(s_{t+1}\) have high value
\(V_{\phi}(s_{t+1}, s_{t+k})\) relative to the current subgoal.
Because different subgoals induce much larger variations in \(V_{\phi}\) than individual actions, the high-level receives a more reliable learning signal, and since \(\pi_{\theta}\) queries \(V_{\phi}\) only for nearby states where estimates are more accurate, the entire hierarchy is less susceptible to noise and approximation errors in the value function, resulting in a more robust policy and credit assignment.

\paragraph{Interpretability.} Identifying subgoals within the offline dataset can provide insights into understanding the decision-making process. For example,  \citet{paul2019learning} present visualizations of the state space in robotic navigation tasks, such as AntMaze and AntTarget, demonstrating that the state space can be structurally partitioned using discovered subgoals. 
The structural decomposition is intuitively meaningful to humans, facilitating better understanding and verification.

\paragraph{Opportunities for Research.}
\begin{itemize}
    \item \textit{Enhancing interpretability of decision making with interpretable subgoals.} 
    Although \cite{paul2019learning} show positive empirical results, current offline relabeling schemes select subgoals with limited transparency into why particular states are chosen or how they steer the learned policy. Embedding interpretability or alignment objectives, such as attributing subgoal selection to human-understandable criteria, would not only clarify the decision rationale, but also foster trust and diagnosability. 
    \item \textit{Scaling to complex observations by identifying latent subgoals.} 
    Researchers can extend offline subgoal relabeling to environments with high-dimensional inputs, such as images, language, or tactile data, with the work by~\citet{park2024hiql} as an example. To do so, methods could identify subgoal representations in some latent space that pinpoint meaningful milestones within these spaces. By focusing on compact embeddings, relabeling can remain effective even when raw observations are noisy, partial, or multimodal.\looseness=-1
\end{itemize}



\section{Discovery with Foundation Models}
\label{sec:largemodels}
Agents that learn skills from scratch through environment interactions are directly exposed to the inherent complexities of the domains in which they operate.
Such agents must learn from their stream of experience how to organize the collected data into meaningful chunks in order to derive a useful set of skills. 
 To mitigate these challenges, we can instead build on prior knowledge contained in large pretrained models to guide the discovery of useful skills in complex environments. The starting assumption for such methods is that pretrained models contain knowledge about the environment of interest. Although this assumption may not always hold, it is likely applicable to many domains of interest and will grow as the training paradigm of LLMs expands in scope.

Simultaneously, an interesting feature of LLM-based methods is that, as these large models are based on human priors and are instantiated through natural language, the set of behaviours will generally be more interpretable. In fact, leveraging language, without using LLMs, has produced a prolific line of work \citep{Shu2017HierarchicalAI,Bahdanau2018LearningTU,fu2018from,Jiang2019LanguageAA,Colas2020LanguageConditionedGG,Colas2020LanguageAA,Akakzia2021GroundingLT}.\footnote{For an in-depth review of the use of language in RL, please refer to the work by \citet{Luketina2019ASO}.} These works underscore, amongst other features, the compositional nature of language.  This quality makes it a particularly useful space to represent a variety of goals.

It is therefore natural to consider LLMs for HRL as they provide both useful inductive biases from pre-training on human data and a meaningful abstraction space through natural language. This connection is reinforced by the fact that \textbf{LLMs, by their very nature, can represent goal-conditioned policies, where goals are specified linguistically}.

As such, many recent works leverage LLMs to decompose tasks into subtasks \citep{Pignatelli2024AssessingTZ,Yang2024SelfGoalYL,Wang2024AgentWM}, an operation done according to their pre-existing understanding of the task's underlying structure. Another perspective is to use LLMs as a measure of interestingness to propose a curriculum of goals and tasks in open-ended domains \citep{colas2023augmentingautotelicagentslarge,Zhang2023OMNIOV, faldor2025omniepic, robogen, zala2024envgen}.
The key to converting an LLM's latent knowledge into a functional agent lies in efficiently learning the options required to execute the decomposed goals.

In this section, we investigate four families of methods that propose solutions to this problem. The first family consists of methods using embeddings from large pretrained models as representations from which option rewards are defined. Next, we present methods that use large pretrained models to provide feedback, in the form of rewards or preferences, for learning different skills. Building on the code generation capabilities of large models, we present two families of methods that write code to either craft reward functions to learn specific behaviours. Finally, as LLMs can be seen as goal-conditioned policies, we cover methods that use them directly to specify subgoals and achieve them.

Not all the presented methods in this section are hierarchical by nature, where hierarchical is defined as containing a high level policy that sequentially selects goals. In some cases, papers focus on defining reward for learning task-conditioned or skill-conditioned policies, without including the high level policy. Given our focus on methods for discovering temporal structure, we believe that learning from these methods is very valuable. 


\subsection{Embedding Similarity}
\label{sec:embsim}

 As foundation models are trained on Internet-scale datasets, their embeddings contain useful structure for a variety of tasks.
 Such embeddings can be the result of contrastive pretraining on image and text pairs, for instance, the Contrastive Language-Image Pretraining (CLIP) encoder \citep{Radford2021LearningTV}. Methods in this section leverage these embedding in pixel-based tasks to learn different behaviours specified through language. Let $\mathbf{w}_i \in \mathbb{R}^d$ represent the normalized feature vector (embedding) generated by the image encoder for the $i$-th image, $I_i$. Similarly, let $\mathbf{u}_j \in \mathbb{R}^d$ be the normalized feature vector generated by the text encoder for the $j$-th text, $T_j$. The similarity between image $i$ and text $j$ is computed using the cosine similarity (which simplifies to the dot product for normalized vectors):
\begin{equation}
    C_{ij} = \mathbf{w}_i^\top \mathbf{u}_j.
\end{equation}
 These embeddings can then be used to represent image or language goals and define reward functions by taking the cosine similarity between the embeddings of the goal and the observation in which the agent is currently situated,
 \begin{equation}r^g(s) = \mathbf{w}(s)^\top \mathbf{u}(g).\end{equation}
 
 This reward function is then maximized by a goal-conditioned policy interacting with an environment to learn behaviours that achieve the specified goals.

\begin{algorithm}[t]
\caption{MineCLIP: Instantiation of 
Algorithm~\ref{alg:hrl_generic} through embedding similarity.}
\label{alg:embedding_sim}
\begin{algorithmic}[1]
\Require Offline dataset $\mathscr{D}$. 
\State Initialize embedding models, $\mathbf{w}$ and $\mathbf{u}$, with the CLIP encoder.
\State Initialize policy parameters $\theta$.
\Statex

\Procedure{OptionSpecification}{$\mathscr{O}, \mathscr{D}$}
    \State Fine-tune multi-modal embeddings, $\mathbf{w}$ and $\mathbf{u}$, through Eq.~\ref{eq:im_loss} and Eq.~\ref{eq:text_loss} on  $\mathscr{D}$.
    \State Define option objectives as  $\Omega=\{r^{g}\}$ where $r^g(s_{t-16:t}) = \mathbf{w}(s_{t-16:t})^\top \mathbf{u}(g)$.
    \State \Return $\Omega$
\EndProcedure
\Statex
\Procedure{OptionLearning}{$\mathscr{O}, \mathscr{D}, \Omega$}
    \State Construct options $\mathscr{O}$:
    \State 
    \hspace{\algorithmicindent} Update $\pi_\theta(\cdot|s,g)$ using transitions sampled from $\mathscr{D}$ to maximize $r^g\in \Omega$.
    \State \hspace{\algorithmicindent} Time-induced termination $\beta$ after fixed $H_o$ timesteps.
    \State \hspace{\algorithmicindent} $\forall s,\ \mathscr{I}(s)=1 $
    \State \Return $\mathscr{O}$
\EndProcedure
\end{algorithmic}
\end{algorithm}

To obtain these vectors, the objective is formulated as minimizing a cross-entropy loss, applied symmetrically for both image-to-text and text-to-image prediction tasks. The loss for predicting the correct text caption for a given image $i$ (considering all $N$ text captions in the batch) is defined as:


\begin{align}
    L_{\text{image}_i} = -\log \frac{\exp(c_{ii} / e)}{\sum_{j=1}^N \exp(c_{ij} / e)},
    \label{eq:im_loss}
\end{align}
where $e$ is the temperature hyperparameter. The loss for predicting the correct image for a given text caption $i$ (considering all $N$ images in the batch) is:
\begin{align}
L_{\text{text}_i} = -\log \frac{\exp(c_{ii} / e)}{\sum_{j=1}^N \exp(c_{ji} / e)}.
    \label{eq:text_loss}
\end{align}


\begin{figure}[t]
    \centering
    \includegraphics[width=0.95\linewidth]{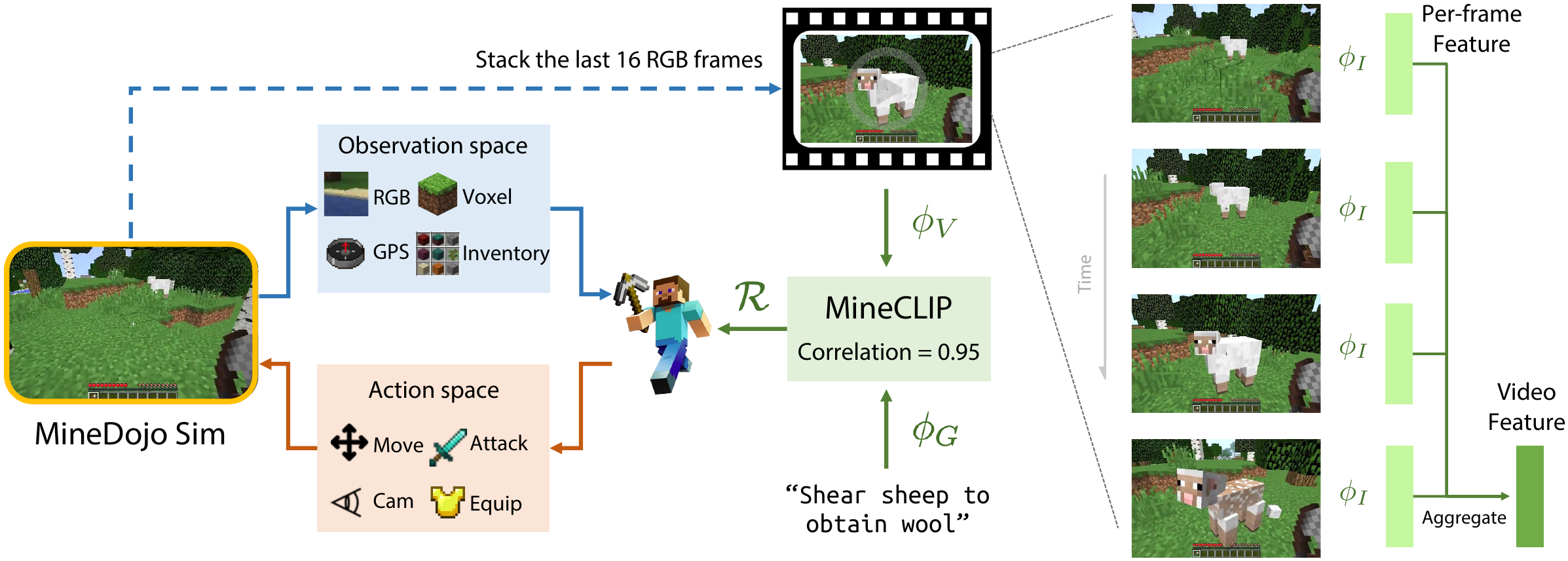}
    \caption{Illustration of the method of embedding similarity for defining option reward functions. Visual observations and language goal descriptions are converted into embeddings, and their similarity (e.g., via MineCLIP) is used to define reward functions for goal-conditioned policies. In this example, the agent is rewarded for successfully performing sheep shearing. Figure taken from \cite{Fan2022MineDojoBO}.}
    \label{fig:emb_sim}
\end{figure}

\citet{Fan2022MineDojoBO}  instantiate this idea in the open-ended Minecraft game \citep{Johnson2016TheMP,Kanervisto2022MineRLD2}. To do so, they introduce the MineDojo framework shown in Figure~\ref{fig:emb_sim}. 
The authors collect a large dataset of Minecraft gameplay for training a reward function that would map textual goals and a sequence of observations to a scalar value indicating their similarity. We present the  algorithm in Algorithm~\ref{alg:embedding_sim}.
The language goal is encoded through the pretrained CLIP encoder \citep{Radford2021LearningTV}  whereas the video encoder is composed of an image encoder and a temporal aggregator that accumulates 16 consecutive frames from the video. This leads to the following non-Markovian reward,
\begin{equation}
r^g(s_{t-16:t}) = \mathbf{w}(s_{t-16:t})^\top \mathbf{u}(g).    
\end{equation}
The authors train their reward model, called MineCLIP,  on the aforementioned dataset using the same losses as in Equation \ref{eq:im_loss} and \ref{eq:text_loss}.
This resulting reward function excels at capturing correct behaviour on a wide collection of tasks, such as ``Combat zombie''.
\citet{Lifshitz2023STEVE1AG} build on this work to obtain an instruction-following agent in Minecraft, where language instructions represent goals.



CLIP-based methods have also been applied to robotics. \citet{Xiao2022RoboticSA} fine-tune the CLIP model on a small dataset of robotic tasks and then utilize the model to label, using a set of predefined annotations, a much larger dataset of unlabeled observations. Using this larger dataset, the authors then train language-conditioned policies to achieve goals through imitation learning. Further improving the sample efficiency of embedding-based methods, \citet{Palo2023TowardsAU} show the possibility of efficiently fine-tuning the same CLIP model on as little as 1000 data points.  
Avoiding the costly operation of fine-tuning large pretrained models, \citet{Cui2022CanFM} investigate the prospect of using the CLIP model in a zero-shot fashion for defining goal-conditioned policies, obtaining good results on robotics tasks. Similarly, \citet{Rocamonde2023VisionLanguageMA} leverage a fixed pretrained CLIP model and study the scaling effect of such models on the resulting RL performance.

To tackle long-horizon tasks, methods like JARVIS-1 \citep{wang2023jarvis1} propose hierarchical frameworks that employ LLMs as high-level planners to sequence options that were obtained by the methods described in this section, that is, embedding-based options. The planner selects options by querying a multimodal memory of past successful plans, generates short-horizon subgoals, and validates them via a "self-check" simulation before dispatching them to a low-level controller for execution. Execution terminates upon goal completion, when a task-specific step limit expires, or prematurely if environment feedback signals an error. Such execution failures immediately prompt the high-level planner to diagnose the underlying issue and generate a newly sequenced plan.

\subsubsection{Relationship to the Algorithmic Template}
The main question raised by the embedding similarity-based method in Algorithm \ref{alg:embedding_sim} is where the embedding models come from. As we have covered in the previous paragraph, this can be achieved either through large datasets from scratch or fine-tuning existing models on smaller datasets. Termination is typically independent of these vectors and defined through a time limit (see Design Principle 2, Section~\ref{sec:option-discovery}).

\subsubsection{Benefits and Opportunities}
\paragraph{Exploration.} Methods building on embedding-based rewards empirically show improved exploration in complex tasks. In particular, in open-ended environments such as Minecraft, the dense nature of the reward functions obtained from embedding similarity significantly helps with exploration, leading to sophisticated behaviour \citep{Fan2022MineDojoBO,Lifshitz2023STEVE1AG}.  The dense nature of such reward functions is also particularly useful for approaches studying the challenge of robotics \citep{Xiao2022RoboticSA,fu2024} and web navigation \citep{Baumli2023VisionLanguageMA}. \citet{Du2023GuidingPI} investigate how guiding exploration with an LLM during a pretraining phase can help an agent's downstream performance. To do so, the authors introduce the idea of restricting the reward function through a similar threshold,
\begin{align}
r^{g,T}(s_t,a_t,s_{t+1}) = 
\begin{cases} 
      r^g(s_t,a_t,s_{t+1}) & \text{if }  > T, \\
      0 & \text{otherwise },
\end{cases}
\end{align}
where $T$, the threshold, is a hyperparameter. This reduces the noise of possibly imperfect embeddings used to define the reward function, further improving exploration.

\paragraph{Transfer.} Another important benefit from LLM-based approaches to skill discovery stems from the compositional nature of language, which easily allows for specifying a variety of goals.  For example, \citet{Du2023VisionLanguageMA} study how pretraining the agent on self-generated goals, where good behaviour is rewarded by the embeddings of an LLM, can lead to improved downstream performance on a set of complex goals. To encourage reaching a diversity of goals that will transfer well, the authors additionally prompt the LLM to generate $k$ goals and reward the agent on the goal with the greatest reward,
\begin{equation}
r^{g_{max}} = \max_{i=1...k} r^{g_i,T}(s_t,a_t,s_{t+1}).
\end{equation}
Similar generalization to different language-conditioned goals is reported by \citet{Lifshitz2023STEVE1AG}. Instead of directly training with a goal-conditioned model, \citet{Mahmoudieh2022ZeroShotRS} efficiently train a discrete set of smaller policies, used as a basis of behaviour. This is then distilled into a single language-conditioned neural network, which can better generalize on a larger spectrum of behaviours than the basis.

\paragraph{Opportunities for Research.} 
\begin{itemize}
    \item \textit{Understanding the trade-offs of different embeddings.} An important question when working with embedding similarity measures is with respect to the origin of the embeddings themselves. Most of the presented papers rely on CLIP, but other embeddings have been used, such as the Bidirectional Encoder Representations (BERT) embeddings \citep{Devlin2019BERTPO} and the Reusable Representations for Robot Manipulation (R3M)   \citep{Adeniji2023LanguageRM}, which is pretrained on the  Ego4D dataset \citep{Grauman2021Ego4DAT} through a combination of contrastive and video-language alignment losses. When considering a wide range of tasks, it is not clear which model shows greater performance, or is more amenable to fine-tuning.
    \item \textit{Expanding beyond text-image similarity.} Most works compute the similarity between a language goal instruction and the current observation. \citet{Sontakke2023RoboCLIPOD} instead compute the similarity between an agent attempting to reach a goal and a demonstration of such a successful behaviour. Moreover, contrary to most works, the authors compute the reward at the trajectory level, that is, the reward is only given at the end of an interaction. The authors show that their approach can be applied even in the case where the demonstration is done by a human physically completing the task, rather than teleoperating a robot, which presents greater opportunities for generalization. 
\end{itemize}

\subsection{Providing Feedback}
\label{sec:provfeedback}
Leveraging the embeddings of foundation models to measure the similarity between a desired goal and the current state places significant emphasis on the quality of the embeddings themselves.
One way to avoid this is by considering the auto-regressive nature of LLMs, which allows for chain-of-thought  \citep{Wei2022ChainOT} and in-context learning \citep{Brown2020LanguageMA}. 
Such capabilities can be particularly useful to define option reward functions. This can be done by taking as input a state, or a trajectory, as well as a goal description, and using LLMs to output a scalar feedback, representing the degree of success with respect to the goal. Alternatively, preferences can be elicited from an LLM over pairs of states that are then converted into a reward model through preference-based learning \citep{wirth}, which can then be used to learn goal-conditioned low-level policies.

\subsubsection*{Direct Reward}
One of the most straightforward way to construct a reward function using foundation models is to employ them as direct evaluators of task success. This approach leverages the broad semantic understanding of large language models (LLMs) and vision-language models (VLMs) to autonomously verify whether an agent has completed a specified goal, bypassing the need to learn reward functions.

Formally, this paradigm models the reward as a direct mapping from an environment state $s \in \mathscr{S}$ and a natural language goal $g \in \mathscr{G}$ to a scalar success metric $y \in \mathscr{Y}$. For instance, to evaluate a binary success measure, \citet{Du2023VisionLanguageMA} pass a sequence of visual observations alongside a query such as ``Did the agent successfully place the cactus left of the sofa?'' to a multimodal model \citep{Alayrac2022FlamingoAV}. The subgoal reward is then directly defined by this binary output $y \in \{0,1\}$:
\begin{equation}
 r^g(s) = y = \texttt{LLM}(s,g).
\end{equation}
Such reward functions are evaluated on a diversity of domains: embodied simulations \citep{Abramson2021CreatingMI}, robotic manipulation with a 6DoF device, and human interactions in the Ego4D dataset \citep{Grauman2021Ego4DAT}. To obtain accurate success measures, the authors have to initially fine-tune the model on a large dataset of expert interactions. Instead of costly model updates,
\citet{kwon2023reward} propose to replace weight updates with few-shot in-context examples, building on improved learning capabilities in the employed LLM.
\citet{pan2024autonomous} show that measures of success obtained from a multimodal LLM have high agreement (up to 92.9\%) with oracle evaluators. Such results are reported on WebArena \citep{zhou2023webarena} and Android-in-the-Wild \citep{rawles2023androidinthewild} benchmarks. Leveraging the strong performance of LLMs as direct reward modelers, \citet{bai2024digirl} successfully train robust RL policies on a variety of goals derived from changing web interfaces.

\subsubsection*{Eliciting Preferences}
\begin{figure}[!t]
    \centering
    \includegraphics[width=\linewidth]{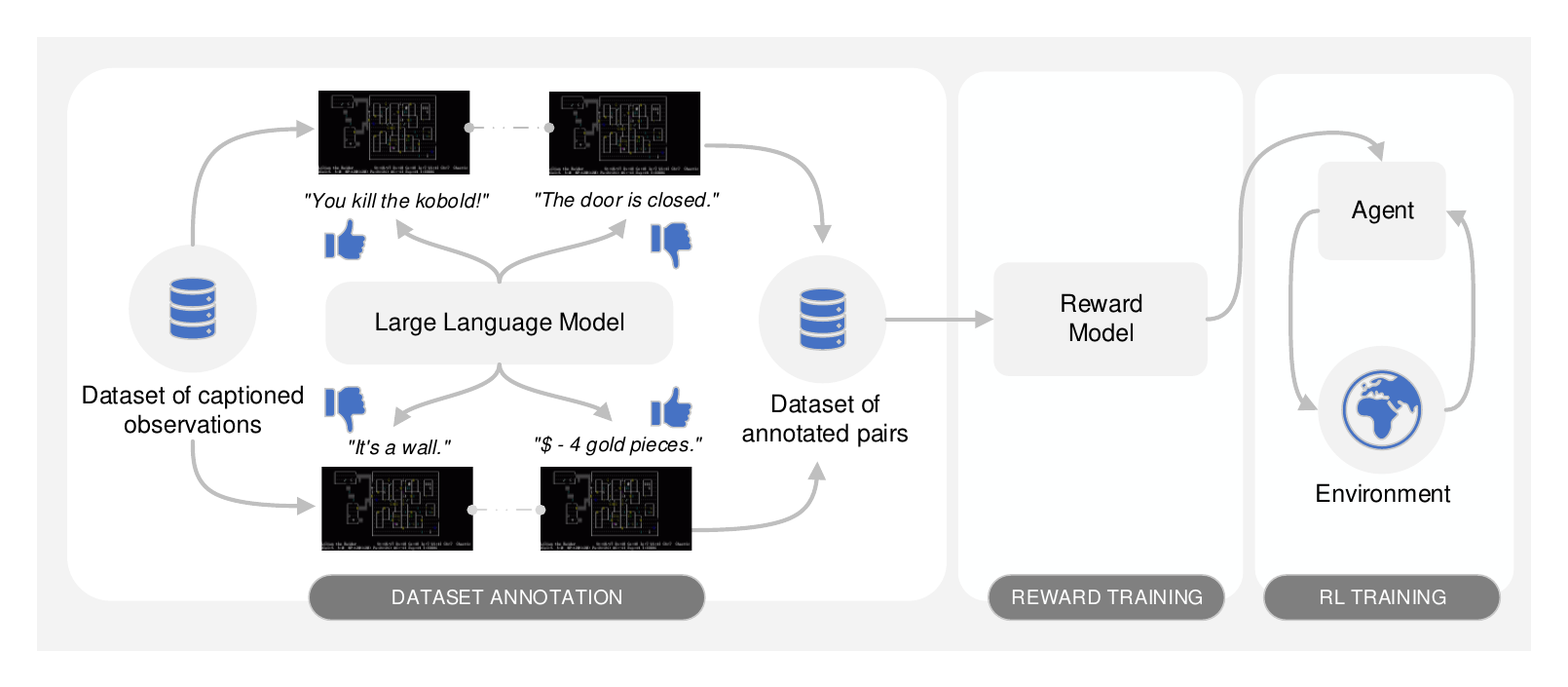}
    \caption{Learning option rewards from AI feedback proceeds in three phases. In the first phase, an LLM is conditioned on a behaviour description and queried for preferences over pairs of observations, which are stored with their preference labels within a dataset.  
    In the second phase, the preferences are distilled into an observation-based scalar reward function. Finally, an agent is trained interactively with RL, receiving a scalar signal at every step through the reward function extracted from the preferences.}
    \label{fig:diagram}
\end{figure}
When an LLM's direct output serves as the reward signal, it may reliably detect success, but it often lacks the nuance required to properly gauge a specific state's value compared to the full spectrum of alternatives. Instead, we can leverage reinforcement learning from AI feedback \citep{Bai2022ConstitutionalAH}. By relying on preference-based learning \citep{wirth,thomaz2006reinforcement}, we can obtain more refined valuations for each state that accurately reflect its relative merit within the much larger state space. Building on this idea, \cite{Klissarov2023MotifIM} introduce the Motif algorithm, which leverages an LLM's feedback to guide an agent acting in the open-ended NetHack environment \citep{kuettler2020nethack}. 
Observations from the environment are presented to an LLM before querying, using chain-of-thought prompting, to provide a preference over which observation is more desirable for a certain goal. This process is shown in Figure~\ref{fig:diagram}.

Formally, the annotation function is given by $\texttt{LLM}:\mathscr{S} \times \mathscr{S} \times \mathscr{G} \to \mathscr{F}$, where $\mathscr{S}$ is the space of states, where $\mathscr{G}$ is the space of goals defined through natural language, and ${\mathscr F = \{ 1, 2, \varnothing \}}$ is a space of preferences for either the first, the second, or none of the captions.
These preferences are then distilled into a reward function through the Bradley-Terry model \citep{Bradley1952RankAO} and given to an RL agent interacting with the environment, \begin{equation}
\label{eq:rewloss}
\begin{aligned}
        \mathcal L (\nu) = - \mathbb{E}_{(s_1, s_2, g, y) \sim \mathscr{D}_\text{pref}}   \Bigg[        & \mathbbm{1}[ y = 1 ] \log  P_\nu[s_1 \succ s_2 | g] + \mathbbm{1}[ y = 2 ] \log  P_\nu[s_2 \succ s_1 | g] \\
        & + \mathbbm{1}[ y = \varnothing ]   \log \left (\sqrt{P_\nu[s_1 \succ s_2 | g] \cdot P_\nu[s_2 \succ s_1 | g]} \right) \Bigg],
\end{aligned}
\end{equation}
where $P_\nu[s_a \succ s_b | g] =  \frac{e^{r_\nu^g(s_a)}}{e^{r_\nu^g(s_a)} + e^{r_\nu^g(s_b)}}$ is the probability of preferring a state $s_a$ to another $s_b$ given a goal $g$; $r_\nu^g$ is the reward defined with respect to the goal specified in the LLM's prompt. \looseness=-1

\begin{algorithm}[t]
\caption{MaestroMotif: Instantiation of 
Algorithm~\ref{alg:hrl_generic} by providing feedback.}
\label{alg:feedback}
\begin{algorithmic}[1]
\Require Offline dataset $\mathscr{D}$ of interactions. 
\State Initialize policy parameters $\theta$ and option reward parameters $\nu$.
\Statex

\Procedure{OptionSpecification}{$\mathscr{O}, \mathscr{D}$}
    \State Prompt the LLM to describe each option’s intended behaviour textually.
    \State Optimize Eq. \ref{eq:rewloss} on  $\mathscr{D}$ through binary cross-entropy, where preferences are queried w.r.t. the option's language description.
    \State Define option objectives as  $\Omega=\{r^{o}_{\nu}\}$.
    \State \Return $\Omega$
\EndProcedure
\Statex
\Procedure{OptionLearning}{$\mathscr{O}, \mathscr{D}, \Omega$}
    \State Construct options $\mathscr{O}$:
    \State\hspace{\algorithmicindent} Update $\pi_\theta(\cdot|s,o)$ through online interactions to maximize $r^o_\nu\in \Omega$.
    \State \hspace{\algorithmicindent} $\mathscr{I}(s)$ and $\beta(s)$ defined symbolically through LLM-generated code.
    \State \Return $\mathscr{O}$
\EndProcedure
\end{algorithmic}
\end{algorithm}

Through the process of comparing states to alternatives, eliciting LLM preferences, or receiving AI feedback, nuanced and fine-grained reward functions can be provided. Such reward functions can also be understood as process-based rewards \citep{uesato2023solving,Lightman2023LetsVS}. \citet{Klissarov2023MotifIM} leverage this characteristic to learn a set of policies that exhibit a certain behaviour across time, such as preferring generally more cautious strategies when exploring. This is in contrast to the work on LLM as direct reward modelers, which typically define rewards for reaching goal states as binary success detectors \citep{Du2023VisionLanguageMA}. 
As illustrated in the MaestroMotif algorithm \citep{Klissarov2024MMotifIM}, the flexibility offered by AI feedback is key in designing HRL agents capable of subtle behaviours and fast adaptation.
Adding to the generality of AI feedback, \citet{wang2024} investigate the resulting policies across a range of continuous control domains using pixel observations and a multimodal LLM.  Their findings show that reward functions generated through AI feedback yield more performant policies compared to embedding similarity approaches or methods that directly query the LLM for scalar rewards. 


\subsubsection{Relationship to the Algorithmic Template}
Algorithm \ref{alg:feedback} details the MaestroMotif method presented in \citet{Klissarov2024MMotifIM}. By leveraging preferences to define a reward function, the option objective can capture nuanced relationships between different states in the open-ended domain of NetHack. However, the authors did not use the same preference-based mechanism to define the initiation and termination functions. Instead, they opted for a symbolic representation through code, which is possibly simpler and more compute-efficient.

\subsubsection{Benefits and Opportunities}

\paragraph{Exploration.} \citet{Klissarov2023MotifIM} illustrate the potential of AI feedback-based rewards to significantly improve exploration on the complex open-ended world of NetHack. 
The obtained reward function is shown to be naturally dense and encodes a variety of important milestones, such as unlocking doors or picking up items. 
The authors hypothesize that, by querying the model on thousands of pairs of observations from the environment, the LLM's common sense reasoning and domain knowledge are distilled into a useful reward function.


\paragraph{Credit Assignment.} 
\citet{wang2024} report that the reward obtained from preferences monotonically increases as the agent advances towards the goal, naturally assigning credit to states in between the starting state and the goal. 
\citet{klissarov2024modelingcapabilitieslargelanguage} further study the dense nature of such reward functions and reveal a strong correlation with value functions obtained at the end of training. As such, value functions have been trained to propagate information through temporal difference learning \citep{Sutton1988LearningTP}, the authors argue that this high correlation is another indication that the reward functions based on LLM feedback are useful for credit assignment. An equivalent perspective is that the resulting dense reward can be seen as a form of reward redistribution \citep{ArjonaMedina2018RUDDERRD,transporting_value,Klissarov2020RewardPU,ni2023when}, which is an established method for improving credit assignment.


\paragraph{Transfer.} In MaestroMotif, \citet{Klissarov2024MMotifIM} show how a set of semantically meaningful skills can be easily re-composed zero-shot to adapt to complex new tasks. 
Leveraging the code generation abilities of LLMs, they propose a neuro-symbolic approach where skill policies are neural networks trained by reinforcement learning, and the high-level policy is defined through code. The authors then use the in-context learning abilities of LLMs to re-compose the skills, significantly outperforming baselines that are trained specifically on each of the tasks.
Their approach highlights how the compositional nature of language can be particularly helpful when combined with a set of linguistically-defined skills, leading to an easily promptable agent.

\paragraph{Opportunities for Research.} 
\begin{itemize}
    \item \textit{Simplifying the reward learning process.} Despite the strength of preference-based methods for crafting rewards through LLMs, they are more complex than directly querying for a reward signal. Is there a way to leverage the improved exploration and credit assignment without the additional complexity? Is an existing dataset of observations needed for eliciting useful preferences? \citet{zheng2024onlineintrinsicrewardsdecision} provide an initial answer to these questions by comparing different ways in which the LLM feedback is leveraged, for example, by using it as a label for a classification loss. Their results show surprisingly strong performance of some of these simpler baselines, even when querying the LLM with online interactions. 
\end{itemize}

\subsection{Reward as Code}
\label{sec:rewardascode}

\begin{figure}[t]
    \centering
    \includegraphics[width=0.95\linewidth]{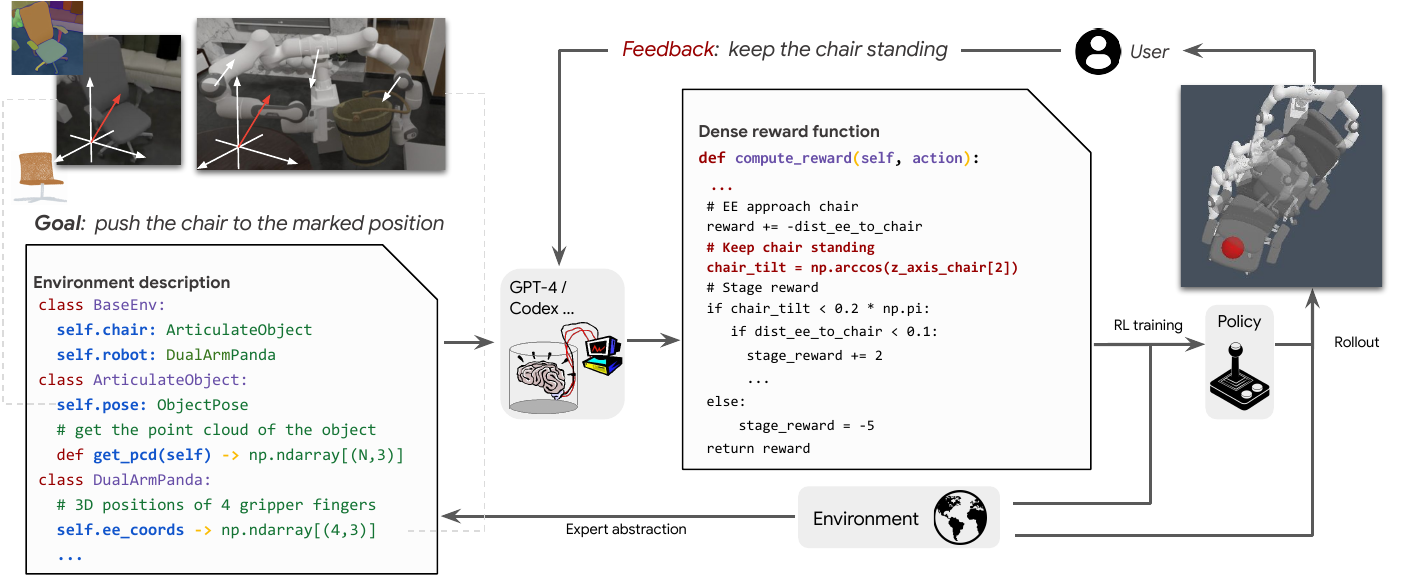}
    \caption{Defining reward functions as code requires access to a symbolic representation of the environment. This is done through an \textit{expert abstraction} function that represents the environment as a hierarchy of Pythonic classes. The \textit{user instruction}
     describes, in natural language, the goal to be achieved. The agent then interacts with the environment to maximize this symbolic reward function. It is also possible to include \textit{user feedback} that summarizes the failure modes of the current reward code. Figure taken from \citet{Xie2023Text2RewardAD}.\looseness=-1}
    \label{fig:reward_code}
\end{figure}

Instead of relying on LLMs to evaluate good and bad behaviour from observations, it is possible to rely on their code generation abilities to write reward functions which can be used immediately to learn low-level goal-conditioned policies. In this line of work, a goal description is given to the LLM as input, as well as additional information from the environment,
\begin{align}
     \texttt{code}^g &\sim \texttt{LLM}(g,\text{info}), \\
     r^g(s) &= \texttt{code}^g(s).
\end{align}
This additional information often constitutes important symbolic information, such as low-level features, that is used to define the code. This code is then executed alongside the environment simulation to generate a reward for every state $s$.
\citet{Xie2023Text2RewardAD} explore the possibility of leveraging an LLM's capacity to code reward functions for robotics tasks.  The authors provide the LLM with additional information in the form of a symbolic representation of the environment (e.g., Python classes describing each object and methods to access specific information about it). Furthermore, the authors provide the LLM with helpful functions from different packages (such as quaternion computation in NumPy) to be used for reward generation. Finally, their algorithms also allow for integrating human feedback. Such a framework is shown in Figure~\ref{fig:reward_code}.
\citet{Yu2023LanguageTR} similarly investigate how LLMs can generate reward functions for learning robotics skills. 
In their approach, an LLM takes as input a detailed language description of a goal and instantiates a set of reward functions. 
 

\begin{algorithm}[t]
\caption{Instantiation of 
Algorithm~\ref{alg:hrl_generic} through reward as code.}
\label{alg:reward_code}
\begin{algorithmic}[1]
\State Initialize policy parameters $\theta$.
\Statex

\Procedure{OptionSpecification}{$\mathscr{O}, \mathscr{D}$}
    \State Prompt an LLM to define the code implementing the reward function.
    \State Define option objectives as  $\Omega=\{r^{g}\}$ where $r^{g} = \texttt{code}^g(s).$
    \State \Return $\Omega$
\EndProcedure
\Statex
\Procedure{OptionLearning}{$\mathscr{O}, \mathscr{D}, \Omega$}
    \State Construct options $\mathscr{O}$:
\State \hspace{\algorithmicindent} Update $\pi_\theta(\cdot|s,g)$ through online interactions to maximize $r^g\in \Omega$.
\State \hspace{\algorithmicindent} Time-induced termination $\beta$ after fixed $H_o$ timesteps.
    \State \hspace{\algorithmicindent} $\forall s,\ \mathscr{I}(s)=1 $
    \State \Return $\mathscr{O}$
\EndProcedure
\end{algorithmic}
\end{algorithm}

Another notable work is that of \cite{Ma2023EurekaHR}, which presents Evolution-driven Universal REward Kit for Agent (EUREKA). They provide a task description to the LLM, such as ``make the pen spin to a target orientation'', and proceed to do an evolutionary search on the reward function space. This process is supported through additional context given to the LLM in the form of selected parts of the source code of the environment. For each candidate reward function generated by the LLM, a full training run is executed via massively distributed RL experiments within Isaac Gym~\citep{Makoviychuk2021IsaacGH}. The most promising reward function candidates are then retained and given to the LLM together with the learning statistics, such that the model performs in-context learning and suggests a new batch of candidates. 

\subsubsection{Relationship to the Algorithmic Template}
Algorithm \ref{alg:reward_code} relies entirely on the coding ability of LLMs. Given that code that represents rewards for simulated robotics is present on the internet, it is a reasonable assumption to expect LLMs to be able to output bespoke variations for different behaviours.

\subsubsection{Benefits and Opportunities}

\paragraph{Transfer.} The ability to efficiently generate reward functions, without human supervision, is particularly important for transfer. For example, \citet{Ma2023EurekaHR} achieve super-human level reward design for complex robotics skills across a variety of embodiments. In the domain of Minecraft, \citet{Li2023AutoMA} show how reward functions as code can be used to solve a variety of long-horizon goals given access to the symbolic features from the environment.

\paragraph{Opportunities for Research.} 
\begin{itemize}
    \item \textit{Going beyond symbolic representations.} Generating a reward function as code is a powerful paradigm: it avoids the need to query the LLM during the RL phase and does not require learning a parametric reward model. However, by definition, such an approach requires access to symbolic features from the domain of interest, which can be limiting if this involves real-world interactions with humans. \citet{venuto2024code} propose to query the LLM to craft its own symbolic representation from high-dimensional observations, similar to the work by \citet{Palo2024KeypointAT}. These representations are then used to define reward functions in code. However, their approach requires access to expert demonstrations, which future work could alleviate.
\end{itemize}

\subsection{Directly Modeling the Policy}
\label{sec:directmodelingpolicy}
So far, we have covered methods that leverage foundation models to define goal reward functions through a variety of strategies, such that goal-conditioned policies can be obtained by maximizing the reward functions. Alternatively, there exists a line of work that uses LLMs to directly model the policy itself, where goals are defined through prompts and conditioning the LLM on them, effectively serving as goal-conditioned policies. In this setting, the LLM is oftentimes updated through in-context learning \citep{Wei2022ChainOT} to obtain policy improvements, bypassing the need for performing parameter updates, which can be costly and time-consuming.
\begin{figure}
    \centering
    \includegraphics[width=1.\linewidth]{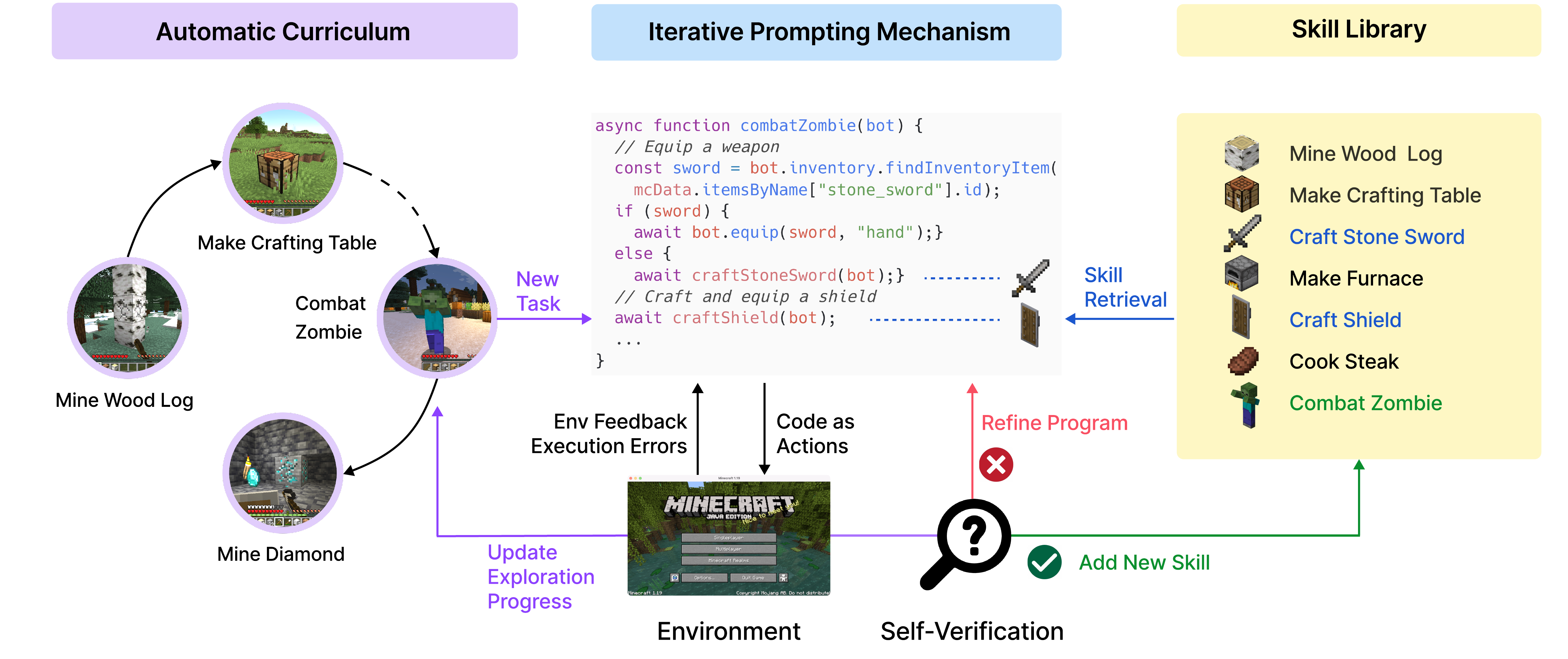}
    \caption{An LLM is conditioned on a goal description and generates snippets of code which instantiate skill policies. When interacting in multimodal environments, such as Minecraft, a bridge between this symbolic skill policy representation and the high-dimensional nature of the environment has to be present. Under such a setting, an LLM can act as an HRL agent, efficiently achieving complex goals. Figure taken from \cite{Wang2023VoyagerAO}.}
    \label{fig:direct_policy}
\end{figure}
Building on the code generation capabilities of LLMs, \citet{Liang2022CodeAP} propose to define robotic skills as policies in the form of Python code,
\begin{align}
     \texttt{code}^g &\sim \texttt{LLM}(g), \\
     a &\sim \texttt{code}^g(s),
\end{align}
where the goal-conditioned $\texttt{code}^g$ acts as a the goal-conditioned policy $\pi(a|s,g)$. They show that the LLM can re-compose calls to an API such that a new code policy achieves a specific goal. In particular, they propose hierarchical code-generation that recursively defines undefined functions from existing functions, leading to strong performance on robotics tasks. \citet{kwon2023language} extend this work, removing assumptions such as providing in-context examples or requiring the LLM to predict end-effector poses.
In the complex open-ended environment of Minecraft, \citet{Wang2023VoyagerAO} propose Voyager (Figure~\ref{fig:direct_policy}), a method leveraging an LLM to continually expand a library of skills. Such skills are crafted by prompting the LLM to define specific behaviours in code, building on an existing JavaScript API \citep{PrismarineJS2013} that allows for grounding the generated code in the multimodality of Minecraft. Voyager further uses ideas of auto-curriculum and self-reflection to update the set of skills, learn new ones, or define their composition for a given task.

Another line of work directly queries the LLM for actions by giving as context the natural language description of a goal and the current state, 
\begin{equation}
 a \sim \texttt{LLM}(s,g).
\end{equation}
In these settings, LLMs output the low-level actions in an environment, effectively as a goal-conditioned policy. This particular instantiation highlights that LLMs are already particularly effective HRL agents that can be conditioned on goal descriptions. The current focus of such models is on computer-based tasks \citep{anthropic2024computeruse,openai2025computerusing}. Despite the appeal of generalizing zero-shot to new language instructions, current LLMs are still quite limited in successfully performing long-horizon tasks by directly selecting low-level actions \citep{openai2024systemcard,zhou2023webarena}.

\begin{algorithm}[t]
\caption{Instantiation of 
Algorithm~\ref{alg:hrl_generic} by directly querying the LLM.}
\label{alg:direct_pol}
\begin{algorithmic}[1]
\Procedure{OptionSpecification}{$\mathscr{O}, \mathscr{D}$}
    \State Prompt an LLM to specify the goal description $g$ in natural language or code.
    \State Define option objectives as  $\Omega=\{g\}$.
    \State \Return $\Omega$
\EndProcedure
\Statex
\Procedure{OptionLearning}{$\mathscr{O}, \mathscr{D}, \Omega$}
    \State Construct options $\mathscr{O}$:
    \State \hspace{\algorithmicindent} $\pi_o(a\mid s)= \texttt{LLM}(a|s,g)$
    \State \hspace{\algorithmicindent} $\forall s,\ \mathscr{I}(s)=1$
    \State \hspace{\algorithmicindent} Terminate when the LLM outputs the end of generation token.
    \State \Return $\mathscr{O}$
\EndProcedure
\end{algorithmic}
\end{algorithm}

\subsubsection{Relationship to the Algorithmic Template}
In Algorithm \ref{alg:direct_pol} the option objectives take an abstract form: language descriptions. Given the algorithmic simplicity and the modular nature of language, it is currently the most widespread method for defining skills in LLM-based AI agents.

\subsubsection{Benefits and Opportunities}
\paragraph{Exploration.} By relying on an LLM's common sense, prior knowledge, and possible API libraries, researchers have shown that agents explore their environment significantly better. By directly modeling the policy, it is possible to condition the LLM on a wider variety of goals and find well-performing policies for a subset of easier goals. This allows for making progress on very hard exploration problems by breaking the task into achievable milestones. Examples include collecting diamonds in Minecraft \citep{Wang2023VoyagerAO} or intricate web navigation tasks \citep{zhou2023webarena}. 
By conditioning an LLM on language and directly outputting a sequence of actions to achieve tasks, agents achieve tasks that would be extremely difficult, or even impossible, to learn by maximizing a reward function.


\paragraph{Transfer.} Directly acting with an LLM greatly simplifies how users can leverage the compositional nature of language. For example, the same LLM can be directly conditioned on a variety of computer interaction tasks and achieve them zero-shot \citep{anthropic2024computeruse}. Alternatively, a library of skills can be re-composed through in-context learning to craft new skills \citep{Wang2023VoyagerAO}. 

\paragraph{Opportunities for Research.} 
\begin{itemize}
    \item \textit{Lifting restrictions on the action space and action frequency.} The prospect of directly generating a wide spectrum of behaviours simply by querying a large pretrained model is particularly appealing. It essentially encompasses the fundamental promise of HRL for fast adaptation thanks to the compositional nature of language. However, it also poses interesting challenges. For example, would such a model be restricted to a certain action space, or is there a way to efficiently adapt to a variety of embodiments? Are there limitations in terms of action frequencies? The domain of computer navigation is especially promising as grounding an LLM in the action space of computers would allow a model to achieve many economically useful tasks. However, the same model could not be used to control an embodied robot, unless fine-tuning is performed, which for large models is costly. A varying action space also raises the necessity to co-fine-tune the model to avoid catastrophic forgetting \citep{rt22023arxiv}. \looseness=-1
\end{itemize}

\section{Using Discovered Options}
\label{sec:5}
\addtocontents{toc}{\protect\setcounter{tocdepth}{3}}

\begin{wrapfigure}[23]{Hr}{0.5\textwidth}
    \centering
    \includegraphics[width=.95\linewidth]{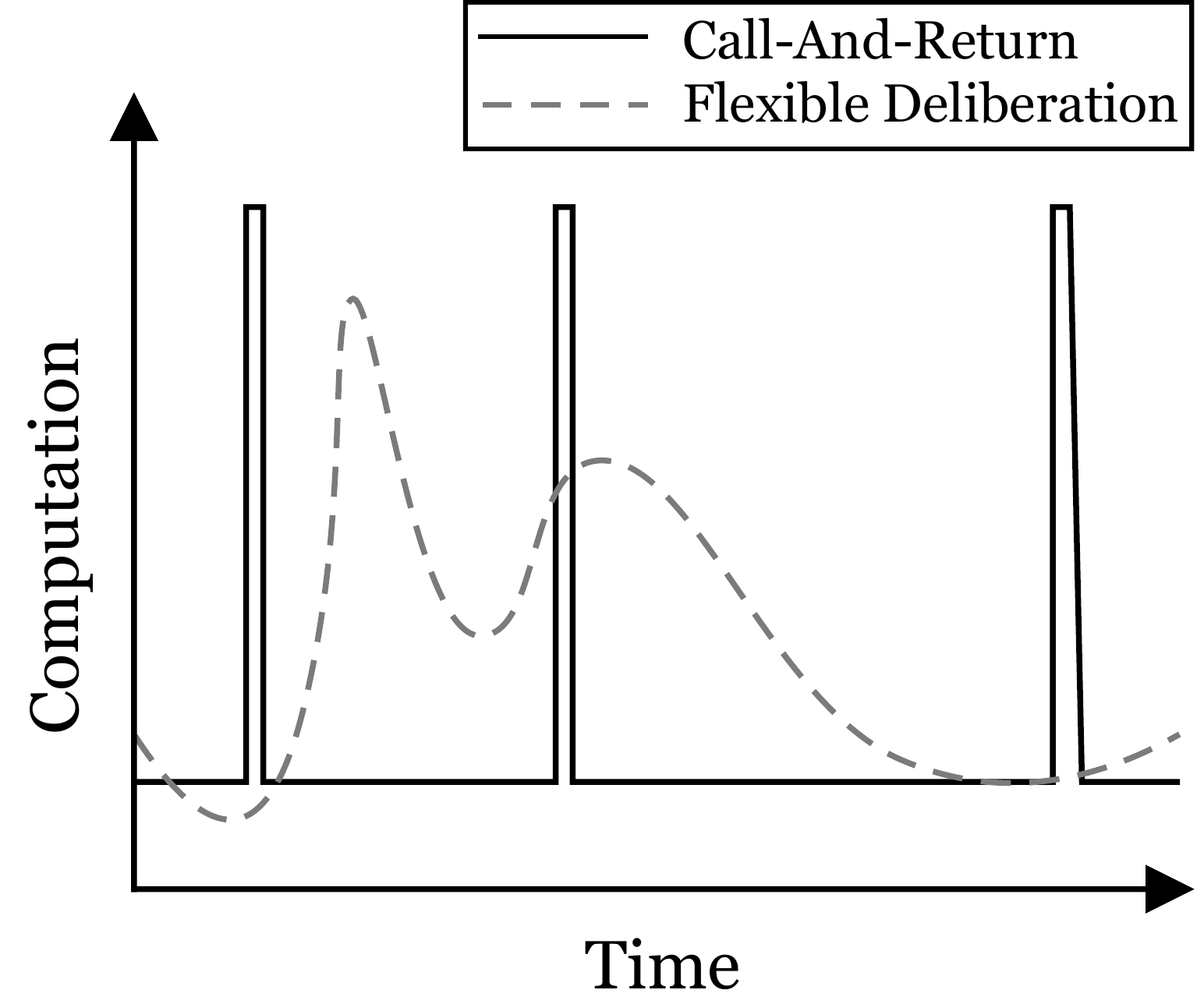}
    \caption{Depiction of the distribution of computation over time for the standard call-and-return model of execution. High-level decisions incur a greater computation cost than low-level ones, producing the spikes that characterize call-and-return. We also present a hypothetical model that distributes computation more flexibly across timesteps.} 
\label{fig:deliberation}
\end{wrapfigure}

In the previous sections, we presented a variety of approaches addressing the option discovery problem. This naturally leads to the question: once an agent has discovered a set of options, how does it use them to select primitive actions in the environment? We refer to the mechanism by which options translate into action selection as the \textit{execution model}. In all execution models, the agent pays the computational cost of selecting a primitive action at every timestep---the same cost incurred in flat RL.  HRL adds  the computational cost of deliberating over options, which might involve a forward pass through a large network, a planning computation using option models, or chain-of-thought reasoning in an LLM.
This added cost can pay for itself; for example, planning with options can require far fewer value-iteration backups to reach a near-optimal solution \citep{mann2015approximate}.
An execution model is fully specified by two choices: \textit{when} the agent deliberates over options, and \textit{how} it uses options to select primitive actions.

\subsection{When to Deliberate}

At one extreme, deliberation occurs only at sparse decision points (e.g., upon option termination under the call-and-return execution model); at the other, the agent reconsiders its choice at every timestep. These endpoints trade off deliberation cost against responsiveness; we illustrate this spectrum in Figure~\ref{fig:deliberation}. One way to think about this tradeoff is through the lens of dual-process theories of cognition \citep{kahneman2011thinking}: selecting among options resembles slow, deliberate ``System 2'' reasoning---for example, model-based planning using option models---while acting according to the current option's policy resembles fast, habitual ``System 1'' action selection. The less frequently the agent engages System 2, the lower the per-step computation overhead, but the greater the risk of persisting with a suboptimal option.

\subsubsection{Call-and-return}

In the call-and-return model, the agent's high-level policy selects a single option at each \textit{decision point}---a timestep at which no option is currently executing, i.e., the start of the episode or the termination of the previous option. The selected option then governs action selection until it terminates, at which point the high-level policy selects again. This is the most widely used execution model and was used to describe HRL in Figure~\ref{fig:hrl_diagram}. Call-and-return concentrates the computational cost of deliberation at option boundaries: the agent amortizes the cost of a single deliberation over the option's duration, so the longer the option persists, the lower the per-step overhead \citep{wen2020efficiency}.

Because options execute for variable durations, the time between consecutive high-level decisions is a random variable. The decision process at the option level is therefore a semi-Markov decision process (SMDP, see Section~\ref{def:option} for details) \citep{bradtke94smdp}. In practice, this requires only minor modifications to standard algorithms. In the model-free case, the experience tuple for updating the high-level policy becomes $(s_t, o_t, \sum_{k=0}^{\tau} \gamma^k r_{t+k}, s_{t+\tau})$, where $\tau$ is the duration of option $o_t$. In the model-based case, the reward model predicts expected discounted cumulative reward during execution, and the transition model predicts the discounted probability of terminating in $s_{t+\tau}$ \citep{Sutton1999BetweenMA}.

A potential limitation of call-and-return is \textbf{commitment}: the agent cannot change course mid-option, even if the option's behaviour becomes suboptimal given newly visited states. The agent must wait for the option to terminate before it can deliberate again.

\subsubsection{Interruptions}

The commitment problem motivates a natural extension: allowing the agent to interrupt a running option before its natural termination and select a new one \citep{Sutton1999BetweenMA}. Rather than waiting for termination, the agent periodically checks whether the current option should be replaced, introducing a parameter $k$ that controls how often this check occurs. When $k$ equals the full option duration, this recovers call-and-return; when $k = 1$, the agent reconsiders at every timestep. The parameter $k$ thus provides a knob that interpolates between the two extremes of the computation cost spectrum.

One concrete interruption criterion analyzed by \citet{Sutton1999BetweenMA} is to interrupt option $o$ in state $s$ whenever $Q(s, o) < V(s)$---that is, whenever the advantage of the current option is negative and the agent could do better by switching. Under this criterion, and when option policies and termination conditions are fixed, the interrupting execution model yields a value function at least as great as call-and-return for any policy over options.

The value of $k$ also determines the Markov properties of the resulting process. At $k = 1$, the process over the augmented state $(s, o)$ is Markov: the action and switching probability depend only on the current state and option, with no dependence on how long the option has been executing. For $k > 1$, the process becomes $k$-Markov in $(s, o)$---the agent must also know how many steps remain until the next interruption check, which can be captured by augmenting the state with a counter.

When options are being \textit{learned} rather than held fixed, unrestricted interruption can cause the hierarchy to degenerate: the agent may learn to terminate every option after a single timestep, collapsing to flat control \citep{Bacon2016TheOA}.

\subsection{How to Select Actions}

Given a deliberation schedule, the agent must also decide how to use its options to produce a primitive action at each timestep. The simplest approach is to follow the policy of a single active option. An alternative is to combine the policies or value functions of multiple options, selecting each action by drawing on the collective knowledge of the option set.

\subsubsection{Single Option}

The most straightforward approach is to follow the policy of the currently active option: at each timestep, the agent executes whatever action that option prescribes. This is the standard assumption in call-and-return and interruptions as described above.

\subsubsection{Option Composition}
\label{sec:option_composition_gpi}

An alternative is to combine the value functions of \textit{multiple} options to select each action. Generalized policy improvement (GPI) \citep{barreto2017successor, barreto2019transfer, Barreto2020FastRL} is a concrete instantiation. Given a set of policies with associated action-value functions $Q_1, \ldots, Q_n$, GPI selects $a = \arg\max_a \max_i Q_i(s, a)$. This provides a guarantee: the resulting policy performs at least as well as the best individual policy in every state. Note that the $Q_i$ here are standard action-value functions for flat policies---there is no initiation set, termination condition, or notion of commitment; GPI queries each policy's value function and cherry-picks the best action at each timestep. Other approaches combine option value functions directly \citep{Todorov2009CompositionalityOO, Hunt2018EntropicPC, Haarnoja2018ComposableDR}, for instance by taking convex combinations.

\paragraph{Composing the two choices.} The choice of action selection mechanism is orthogonal to the choice of deliberation frequency, and interesting execution models arise from combining them. The \textit{option keyboard} \citep{barreto2021optionkeyboard} is a clear illustration. It defines composite options whose internal policy uses GPI-style maximization over multiple base-option value functions at every timestep, while a high-level call-and-return policy selects \textit{which} composition to pursue. Concretely, suppose the agent has $n$ base options, each with a policy $\pi_i$, an action-value function $Q_i$, and a \textit{cumulant} $e_i$---an option-specific reward signal capturing what the option is trying to achieve \citep{sutton2011horde}. A new composite option is specified by a weight vector $w \in \mathbb{R}^n$ that linearly combines the base cumulants into $e = \sum_i w_i e_i$. Let $Q_i^e(s, a)$ denote the value of taking action $a$ in state $s$ and thereafter following base policy $\pi_i$, evaluated under the combined cumulant $e$. At each timestep, the composite option selects
\begin{equation}
    a = \arg\max_{a} \max_i \, Q_i^e(s, a),
\end{equation}
continuing until a termination condition is met. The high-level policy then chooses a new weight vector and the process repeats. Within the composite option, multiple value functions are queried at every timestep (composition); between options, the high-level policy deliberates over weight vectors (call-and-return). At the other extreme, combining composition with interruptions at $k = 1$ recovers standard GPI---the agent composes value functions at every timestep, thereby precluding the need for a high-level policy, with no temporal commitment and a fully Markov process. The option keyboard has recently been instantiated with options derived from the graph Laplacian \citep{chandrasekar2026lk}, illustrating how discovery methods from Section~\ref{sec:od/spectral} supply the base options that composition mechanisms then combine; we return to this in Section~\ref{sec:multiple-behaviours}.

The execution models above define how an agent converts its options into actual decisions. They leave open the question of how the agent learns which options to invoke; we turn to this question next.

\subsection{Learning High-level Policies}
\label{sec:learn_highpolicy}

The agent's high-level policy, $\mu(o|s)$, is responsible for selecting an option. We present different approaches to learning this quantity by separating methods into three categories: model-free approaches, model-based approaches, and approaches that rely on in-context learning using foundation models.

\subsubsection{Model-free Approaches}

Usual model-free RL methods (like Q-learning) can, with slight modifications, be used to learn a policy that selects options $\mu(o|s)$. These modifications simply involve discounting rewards obtained during option execution appropriately and using the state at the end of option execution as the next environment state, i.e., the experience tuple used to update the agent is $(s_t, o_t, \sum_{k=0}^{\tau}\gamma^k r_{t+k}, s_{t+\tau})$, where $\tau$ is the duration of execution of option $o_t$ \citep{bradtke94smdp}. This approach, while simple, treats option execution as a black-box. When the chosen option is Markov, meaning that its duration $\tau$ can be written purely as a function of state (and not time), then intra-option learning can be used for improved sample-efficiency. As long as states observed during option execution are inside the option's initiation set, then the corresponding transitions can be used to update $\mu(o|s)$ \citep{sutton1998intra}. Specifically, an SMDP transition $(s_t, o_t, \sum_{k=0}^{\tau}\gamma^k r_{t+k}, s_{t+\tau})$ can be decomposed into up to $\tau$ transitions of the kind $(s_i, o_t, \sum_{k=0}^\tau \gamma^k r_{i+k}, s_{i+\tau})$ for all $i$ such that $s_i \in \mathscr{I}_o$. \citet{Bacon2018Temporal} later generalize these insights to policy gradient methods by proposing the option gradient theorem. 

\paragraph{Bandits that maximize learning progress.} A popular model-free approach is to treat the high-level policy as a contextual bandit (which can be thought of as an MDP with $\gamma=0$). The reward function for the bandit is designed to carefully trade off various objectives. For example, when the extrinsic reward is dense and informative, the bandit simply chooses the option expected to maximize the reward \citep{schaul2020adapting}. When the reward function is sparse or deceptive, then a measure of \textit{learning progress} (LP) is often added to the extrinsic reward; the idea is that the agent should pick options that (in addition to greedily maximizing reward) would also improve its knowledge of the environment and its own competence in the environment \citep{colas2022autotelic}. Although measuring LP itself is intractable, proxies are used in practice. Competence progress \citep{oudeyer2009im,stout2010competence} prioritizes skills whose capabilities change the most with time---these skills represent subgoals of intermediate difficulty \citep{pmlr-v80-florensa18a}. Count-based bonuses prioritize options that lead to high novelty \citep{bagaria23scaling,badia2020ngu,badia2020agent57}, and density-based approaches \citep{pong2020skewfit} attempt to maintain a high entropy distribution for option selection from different states \citep{pitis2020mega}.

\subsubsection{Model-based Approaches} 

In the model-free case, the high-level policy $\mu(o|s)$ is a learned mapping from states to options. In the model-based case, this mapping is computed on the fly: the agent learns transition and reward models for its options, uses those models to plan, and the plan itself determines which option to execute next. In some model-based approaches, options are used only inside the planner---to propagate value over long horizons---and are never executed in the environment at all. The appeal of this approach is that options afford temporally-extended models of the world, which enable longer-horizon planning. When the agent instead learns single-timestep models, it must roll them out over a long horizon, and model-prediction errors compound over time \citep{talvitie2017self,janner2019mbpo}, leading to potentially massive errors in value approximation \citep{kearns2002near}. Options reduce this compounding by requiring fewer model applications to cover the same horizon \citep{mann2015approximate}---but they do not eliminate it entirely, since option models are themselves learned via bootstrapping and can accumulate bias at long timescales \citep{farebrothertemporal,farebrother2026compositional}.

Regardless of the type of option model, several shared concerns arise during learning. The agent's interaction data can be used to learn option models either on-policy, updating the model for an option after it is executed \citep{Sutton1999BetweenMA}, or off-policy, using intra-option learning \citep{sutton1998intra} to simultaneously learn about many options from the data collected at every timestep. Some methods learn in the agent's raw observation space, which must contend with high-dimensional inputs and outputs~\citep{nair20hierarchical}; focusing on local, subgoal-conditioned models can ease this burden~\citep{Lo2024Goal}. Others operate in an abstract state space, which introduces the challenge of \textit{drift}: option models must rapidly adapt to changes in the evolving abstract representation.

The most basic use of an option model is to evaluate a single option: predict what will happen if the agent executes option $o$ from state $s$, and use that prediction to estimate the option's value. But the real power of option models emerges when the agent plans over a \textit{sequence} of options---evaluating the expected return of executing option 1, then option 2 from wherever option 1 terminates, and so on. To plan over such a sequence, the agent must represent the \textit{handoff} between consecutive options: what does the world look like when one option terminates and the next begins? The choice of how to represent this handoff determines both what kind of sequential composition is possible and what planning algorithms can be used. We distinguish three families of approaches based on the information they retain about the boundary state.

\textit{(a) Full distribution over boundary states.} The most expressive representation is to model the complete distribution over states that an option might terminate in. \citet{silver2012compositional} showed that option models have a natural algebraic structure: an option model can be represented as a matrix bundling transition probabilities and rewards, and composing two options sequentially---executing one, then the other---corresponds to simple algebraic manipulations of their matrices. Option models are therefore closed under composition: chaining two yields another option model of the same form, enabling recursive multi-level planning. However, representing full transition matrices is only tractable in finite state spaces. To extend compositional planning to continuous state spaces, recent work learns full generative models of the \textit{successor measure}---the discounted distribution over future states induced by following a policy \citep{dayan93successor}. The discount factor $\gamma$ can be interpreted as a continuation probability, making the option's lifetime geometrically distributed; because the geometric distribution is both memoryless and closed under composition, chaining successor measures at different discount factors produces another successor measure of the same form. This observation underlies the \textit{geometric horizon model} (GHM), a generative model of the successor measure conditioned on a policy and discount factor \citep{Janner20GammaModels}. \citet{Thakoor2022GeneralisedPI} showed how the successor measures of individual policies compose---via sampling---into the successor measure of a \textit{geometric switching policy}, a sequence of policies where each executes for a geometrically distributed duration. \citet{farebrother2026compositional} generalized this to variable switching probabilities, allowing each policy in the sequence to persist for a different duration, and introduced a horizon-consistency training objective that anchors long-horizon predictions to shorter-horizon ones, reducing bias at long timescales. The strength of this family is unbiased composition via sampling; the limitation is that sampling-based planning (e.g., random shooting) does not scale as gracefully to long combinatorial plans as the symbolic methods discussed in (c).

\textit{(b) Expected boundary state.} A simpler alternative is to predict only the expected terminal state and reward---an \textit{expectation model} \citep{wan2019expectation}. When the value function is linear in the state representation, the expected state is sufficient: $\mathbb{E}[V(s')] = V(\mathbb{E}[s'])$, so no information is lost \citep{schoknecht2002optimality,parr2008analysis,sutton2008dyna,sorg2010linear}. Expectation models are also efficient to learn, requiring only the solution of a system of linear equations \citep{sutton2023reward}, which is especially useful when state representations evolve over time. However, when the agent chains multiple options, it must feed the predicted terminal state of one option into the model of the next. Under nonlinearity, this computes $V(\mathbb{E}[s'])$ rather than $\mathbb{E}[V(s')]$; by Jensen's inequality, the resulting bias compounds with each link in the chain. Expectation models are therefore well-suited to evaluating individual options or short sequences under linearity, but less so to planning over long compositions in general settings.

\textit{(c) Abstract boundary state.} A third approach, developed under the \textit{skills-to-symbols} framework \citep{Konidaris2018FromST}, is to compress the boundary representation through state abstraction. The resulting option models predict transitions between \textit{abstract} states rather than in the original observation space \citep{li2006towards}. When options drive all relevant state variables to a small range of values, the abstraction takes the form of a \textit{graph}: nodes correspond to abstract states and edges to options, with an edge between two nodes when one option terminates in a state where another can successfully begin \citep{Konidaris2018FromST,bagaria2021skill,bagaria2025imdsg}. Options that do not control all state variables---setting some while leaving others unchanged---unlock a more structured representation: a type of factored MDP \citep{boutilier2000factored} described using Planning Domain Definition Language (PDDL) \citep{McDermott98}. The advantage of generating a PDDL description is that it can be solved by off-the-shelf classical planners, even for long-horizon, combinatorial problems that would be intractable for sampling-based methods. The cost is that the abstraction is lossy: the agent discards information about where exactly within an abstract state it will end up, which can lead to suboptimal plans \citep{abel2020value}. \citet{rodriguez-sanchez2024learning} propose continuous abstract state representations that relax the discrete assumption while providing formal bounds on the resulting value loss, quantifying exactly how much reward the agent foregoes by planning in the abstract space rather than the original one. \citet{james2020portable} address a complementary challenge---transferability---by learning abstract representations expressed from the agent's own perspective, which reduce sample requirements across sequences of related tasks \citep{Konidaris2007BuildingPO}.

In summary, the three families of boundary representations---distributional, expected, and abstract---trade off the richness of the handoff information against the tractability of the resulting planner. Distributional models enable unbiased composition via sampling, expectation models offer efficiency under linearity but degrade under chaining, and abstract models sacrifice distributional detail for access to powerful combinatorial planning algorithms. The choice depends on the structure of the problem: whether the state space is continuous or discrete, whether options have tight or diffuse effects, and whether the planning horizon demands combinatorial reasoning.

\subsubsection{In-context Learning}
If the high-level policy over options is implemented using a foundation model, e.g., an LLM,
in-context learning can be used to learn the high-level policy, $\mu(o|s)$. This can be done by having the LLM output Python code that implements skill-selection logic \citep{Wang2024AgentWM,Klissarov2023MotifIM}, or to output formal plans described using PDDL \citep{silver2024generalized}. Such policies can then be updated by providing execution traces as context to the LLMs and asking for code refinements. It is also possible to directly deploy the LLM in the environment to select skills at every high-level decision point \citep{saycan2022arxiv}. Since such approaches do not require gradient updates, they potentially offer faster adaptation. However, the nature of in-context learning is currently not well understood, for example, in terms of generalization and robustness, and is an active area of research.

\section{Challenges of Discovery}
\label{sec:6}
Arguably, one of the biggest challenges in discovering temporal abstractions arises from the \textbf{rich diversity of objectives} that the community has shown to yield meaningful options. 
As discussed throughout this paper, there are many distinct use cases for HRL, meaning methods are often aiming to achieve entirely different things. Because of this, it is frequently difficult to directly compare different methods across domains. This variety can be observed in the wide array of approaches presented in Sections~\ref{tabularasa}, \ref{offline}, and \ref{sec:largemodels}, begging the question whether there exists a single option discovery algorithm that could be generally applied.

Additionally, the \textbf{complexity overhead} that option discovery sometimes introduces can make it less appealing from a practical perspective \citep{jinnai2019skill}. The time invested by a practitioner in setting up an HRL algorithm is valuable. If this time investment does not lead to significantly improved performance on a particular task, or is not generally applicable across tasks, the practitioner will likely choose a simpler approach. 

The two aforementioned points indicate that there is much potential for research in HRL  to find \textbf{reliable and general} solutions as well as understanding where to apply them  (see Section \ref{sec:8}). We now highlight prominent technical challenges that arise when attempting to discover temporal abstractions.

\subsection{Non-Stationarity}
\label{sec:non-sta}

One of the main difficulties in learning a hierarchy of behaviours stems from its modular nature. A hierarchical agent must learn, potentially simultaneously, about the option policies, option reward functions, termination functions, initiation functions, and high-level policy. These components interact and are typically defined in terms of each other, so the non-stationarity induced by learning one will create corresponding non-stationarity in the others. 

A straightforward approach to deal with this non-stationarity is to learn the different components separately. For instance, this can be done by leveraging offline datasets (methods in Section \ref{offline}) to first learn a set of option rewards or a set of option policies, before fixing them. These components can then be provided to a high-level policy that will learn to achieve a certain task. Similarly, we can leverage an LLM's prior knowledge to define an option reward function or  a policy directly  (methods in Section \ref{sec:largemodels}). These would create stationary targets for the remainder of the components. LLMs can also be used to model the high-level policy itself, either by directly querying them or by leveraging their coding abilities to define the skill execution logic. The in-context learning abilities of LLMs could further allow for fast, gradient-free adaptation with respect to a changing option set.

When learning tabula rasa,  non-stationarity between all the randomly initialized option components can be particularly challenging. 
It is common to first define an option learning phase, where the high-level policy acts according to a more exploratory behaviour, for example by uniformly choosing over the options \citep{machado2017laplacian,eysenbach2019diversity}. Such a phase is meant to provide experience in learning the option reward functions and option policies.
\citet{nachum2018data} emphasize the difficulty of non-stationarity in HRL when learning from
past experiences that are stored in a dataset, called an experience replay buffer \citep{LIN1991323}. 
An option that was previously sampled and stored within a replay buffer, together with the experience it generated, would not produce the same data distribution if we were to sample it now. 
To alleviate this, \citet{nachum2018data} relabel which option was used for a stored datapoint with the option that is currently most likely to generate the actions seen in this datapoint. 

The non-stationarity challenge also affects the initiation function. Any method that learns initiation sets from option outcomes---for example, by classifying states according to whether option execution succeeded from them (Section~\ref{sec:skillchaining})---implicitly evaluates the option policy as it stood when the data was collected; such binary classification (or, equivalently, Monte Carlo value estimation) is only sound when the option policy is fixed. When the policy is still improving, a vicious cycle emerges: when an option fails from a state, its initiation probability there decreases, making the agent less likely to execute the option from that state, which in turn deprives the option policy of the very data it needs to improve in that region. Left unchecked, initiation sets shrink over the agent's lifetime, becoming overly conservative. \citet{bagaria2023effectively} diagnose this pathology and address it with tools from off-policy evaluation, adding exploration bonuses that increase the initiation probability of states from which policy improvement is most likely.

Non-stationarity, to a certain degree, has always been present in RL---for example, in actor-critic algorithms \citep{konda2000ActorCritic}. However, the large number of moving parts in a hierarchical agent makes it a fundamental challenge in HRL, and our understanding of it remains limited, both theoretically and empirically. Finding reliable option discovery methods will require both better ways to measure such instabilities and learning algorithms that remain stable in their presence.

\subsection{Simultaneously Learning About Multiple Behaviours}
\label{sec:multiple-behaviours}

If an agent has access to many options, it may solve a variety of tasks by re-composing option policies. However, such a large library comes at the cost of first learning the options themselves, highlighting some of the fundamental trade-offs presented in Section \ref{sec:tradeoffs}. 

To approach this problem, it is convenient to turn to off-policy algorithms \citep{precup_offpolicy}. Such algorithms allow for learning from data that was not generated by the current policy. \citet{Klissarov2021FlexibleOL} propose update rules to improve all options simultaneously by relying on a decomposition of the state-option distribution, introducing a minimal amount of off-policy corrections, and remaining compatible with any policy optimization algorithm. Their method can also be seen as an all-options policy optimization, similar to all-action updates in RL \citep{Sutton2001ComparingPA}. \citet{Daniel2016ProbabilisticIF} instead leverage the perspective in which options are seen as latent codes. The authors adopt an expectation-maximization approach, which assumes a linear structure of the option policies. \citet{smith2018inference} alleviate this assumption and derive a policy gradient objective that improves the data efficiency and interpretability of the learned options.  \citet{ho2} leverages dynamic programming to infer option and action probabilities in hindsight.

We have previously introduced the methods of hindsight relabeling \citep{Andrychowicz2017HindsightER} as part of the skill discovery methods. We can reframe their approach through this question: if you have a multitude of options, or even a continuous spectrum, which other option should you update for a given trajectory? Hindsight experience replay answers this question by relabeling the trajectory stored in the replay buffer with the final state that was reached. This is an instance of off-policy learning as the experience generated by one policy is used to update another policy. The importance of learning off-policy through re-labeling in a hierarchy is further emphasized by \citet{nachum2018data} and  \citet{levy2018learning}.

Is it possible to sample-efficiently learn about multiple options from a single stream of experience? \citet{Barreto2020FastRL}
 propose the Generalized Policy Improvement (GPI) update rules to answer this question (see also Section~\ref{sec:option_composition_gpi}). The authors extend the concept of improvement from a single policy to multiple policies simultaneously. 
 Specifically, their theorem states that, for a given set of policies, $\pi_1, \pi_2,...,\pi_n$, and their associated approximate $Q$ values,  $Q^{\pi_1},Q^{\pi_2},...,Q^{\pi_n}$, 
\begin{align}
    \pi(s) \in \underset{a}{\text{argmax}} \max_{i} Q^{\pi_i}(s,a),
\end{align}
then $Q^{\pi}(s,a) \ge \max_{i} Q^{\pi_i}(s,a)$. This update rule is used by \citet{barreto2021optionkeyboard} to efficiently learn how to execute a combination of options. \citet{Thakoor2022GeneralisedPI} further generalize the results beyond Markov policies, in particular, to options whose execution duration follows a geometric distribution. Recently, \cite{chandrasekar2026lk} have shown how options obtained from the graph Laplacian (see Section~\ref{sec:od/spectral}) can be used as the set of policies (options) for GPI, also quantifying the benefits of learning state-dependent weights that combine such policies. The idea of learning efficiently about multiple policies is closely related to concepts such as the successor representations \citep{dayan93successor} and successor features \citep{barreto2017successor}, as well as other decompositions of the transition function \citep{touati2023does}.

\subsection{Combining Rewards}
When learning option policies through their option reward functions (Definition \ref{def:option_rew}), we are faced with another important question: how should we balance between the option reward and the environmental reward? \citet{dayan1993feudal} argue that the lower-level policies should be agnostic to the environmental reward and learned only through the intrinsic one.\footnote{Although this work predates the options framework~\citep{Sutton1999BetweenMA}, these lower-level policies can be viewed as option policies.} \citet{vezhnevets2017feudal} take a softer approach and provide both rewards, possibly as the environmental reward contains rich information in the environments that were considered. In other cases, there is no intrinsic reward at all \citep{Bacon2016TheOA}. \citet{sutton2023reward} investigate these questions from the perspective of planning and learning with options that do or do not respect the environmental reward. Reward-respecting options are much more effective when used for planning. \citet{Zahavy2022DiscoveringPW} propose a point of view of constrained optimization to balance these objectives and leverage Lagrange multipliers in practice. A thorough examination concerning the trade-offs of how hierarchical agents combine environmental reward and intrinsic reward has yet to be made.

\section{Related Fields}
We now discuss several fields adjacent to HRL. As noted earlier, HRL leverages temporal abstraction, and it is therefore important to understand how it relates to abstractions defined over other components of RL, such as state and action abstractions.
Continual RL has also received increasing attention. Since continual-learning agents typically operate under limited capacity, abstraction can provide a principled mechanism for retaining and reusing knowledge, and temporal abstraction in particular offers one way to achieve this. Moreover, as discussed in Section~\ref{sec:non-sta}, hierarchical agents can inherently introduce non-stationarity, suggesting that methods developed in continual RL may also benefit HRL.
As we motivate HRL using the analogy of programming languages in Section~\ref{sec:2}, we also discuss the conceptually adjacent field of programmatic RL.
Finally, HRL is closely connected to multi-agent RL. Many hierarchical systems can themselves be viewed as multi-agent systems, while hierarchical organization also naturally arises in multi-agent settings, particularly in cooperative multi-agent RL. These connections therefore warrant discussion.

\label{sec:7}
\subsection{State and Action Abstractions in Reinforcement Learning}
Scaling RL for real-world applications faces challenges in handling high-dimensional or noisy observations and large action spaces. Accordingly, the RL community has long explored \textit{abstractions}, a core computer science concept for suppressing low-level details so that reasoning can proceed at a higher conceptual level~\citep{DBLP:journals/mima/ColburnS07}. This can mitigate the curse of dimensionality and improve sample efficiency \citep{konidaris2009efficient,ho2019value,DBLP:journals/corr/abs-2203-00397,ahmetoglu2026skill}.
The temporal abstractions introduced above provide one way to achieve this by reasoning over temporally extended behaviours rather than primitive actions. More generally, RL admits several other forms of abstraction that simplify different components of the learning problem. In this section, we review these forms of abstraction and the equivalence relations they induce.



\paragraph{State abstraction.} It maps concrete states to compact representations while retaining information relevant to decision making. At the level of implementation, classical approaches construct explicit state aggregations~\citep{singh1994reinforcement,li2006towards,hutter2014extreme}, whereas deep RL approaches use neural networks to compress high-dimensional observations into learned representations~\citep{gelada2019deepmdp,DBLP:journals/corr/abs-2203-00397, allen2021markov_abstractions}. Recent methods learn such representations via gradient-based optimization, introducing objectives such as reward prediction, self-prediction, inverse dynamics prediction, or metric losses to obtain compact and informative embeddings~\citep{gelada2019deepmdp,allen2021markov_abstractions,ni2024bridging,luo2025understanding}.

Independently of how the representation is implemented, state-abstraction schemes differ in the behavioural information they preserve~\citep{li2006towards}. For example, some merge states only when they yield identical immediate rewards and transition dynamics under every action (which is called \textit{model-preserving}), whereas others require equality of their optimal action-value functions. Bisimulation metrics relax these binary equivalence criteria by defining a graded behavioural distance based on discrepancies in rewards and transition dynamics~\citep{ferns2004metrics,ferns2011bisimulation,castro2020scalable,zhang2020learning,luo2025understanding}. 
Those metrics can support either state aggregation, by grouping states whose distance falls below a chosen threshold, or representation learning, by embedding states so that distances in the learned representation space approximate their behavioural distances~\citep{luo2025understanding}.

\paragraph{State-action abstraction.} 
Another line of work considers \textit{state-action abstraction}, notably MDP homomorphisms, which map states and actions to an abstract MDP while preserving reward and transition dynamics \citep{ravindran2001symmetries,ravindran2004algebraic,ravindran2004approximate,narayanamurthy2008hardness}. Such abstractions can identify behaviourally equivalent state-action pairs, including those arising from symmetries in the underlying MDP.
Similarly, this idea has been explored in deep RL literature~\citep{rezaei2022continuous,fujimoto2025towards}.

\paragraph{Action abstraction.} 
Throughout this paper, action abstraction is discussed primarily as temporal abstraction, whereby behaviours, usually spanning multiple timesteps, are represented as abstract actions.
We devote less attention to
per-timestep action abstraction,  which reduces the computational burden of large action spaces through action elimination \citep{even2006action,zahavy2018learn}, action embeddings or transformations~\citep{van2009using,dulac2015deep,jiang2022efficient}, and affordance-based filtering~\citep{abel2014toward,fulda2017can,khetarpal2020can}. Affordances, for example, restrict the actions available in a state to those satisfying a specified intent or task-relevant criterion.
A related but distinct form of abstraction operates over policies rather than the actions available at an individual decision point. 
\textit{Policy abstraction} maps policies into a compact space to facilitate policy selection or search and need not introduce temporal abstraction~\citep{mutti2022reward,tang2022inputting,zhang2022towards,pmlr-v202-nabati23a}, although it overlaps with HRL when the policies are temporally extended skills~\citep{barreto2021optionkeyboard}.\looseness=-1



\paragraph{Relationship to HRL.}
In addition to temporal abstraction, HRL can integrate of various types of state and action abstractions.
In earlier work of HRL, two common forms of state abstraction are employed: first, state abstraction within the high-level controller, enabling learning or planning in a more tractable space.
Feudal RL \citep{dayan1993feudal}, as a prominent example, employs information hiding to abstract low-level details from the state observed by the manager.
Second, state abstraction within the low-level controller, which speeds up option policy learning by constructing representations that are only relevant for that option~\citep{konidaris2009efficient}.
State abstractions within MAXQ \citep{dietterich1998maxq} and option models are natural examples, as options can be defined exclusively for states where the option is applicable.

Several HRL methods combine state and action abstractions with temporal abstraction to facilitate transfer across tasks. Relativized options use partial MDP homomorphisms to represent a family of structurally related options through a single reduced option MDP~\citep{ravindran2002model,ravindran2003smdp,ravindran2003relativized}. When the option is invoked, the current state is mapped into this reduced model, and the selected abstract action is mapped back to an appropriate concrete action according to the relevant transformation. Portable options~\citep{Konidaris2007BuildingPO} pursue the same transfer objective but use a different mechanism: rather than selecting an explicit transformation at each invocation, they define the option policy, initiation set, and termination condition directly over agent-centered features whose meanings remain consistent across tasks. These options can therefore be reused across tasks.
\citet{castro2010using} pursue the same broad goal of transfer across MDPs, but use bisimulation distances to determine which actions from a source policy can be reused in a target task. Their analysis also extends to temporally extended actions, showing that options can provide a more effective unit of transfer than primitive actions.

Whereas these methods seek representations that allow skills to transfer across settings, another line of work asks how states should be abstracted once a set of skills is given. \citet{Konidaris2018FromST} construct a symbolic state abstraction tailored to the available skills, mapping each low-level state to symbols that determine which skills are applicable and how they change the abstract state. More generally, 
\citet{abel2020value} characterize conditions under which a state abstraction and associated options preserve the ability to represent near-optimal policies. Each such option can initiate in any original state mapped to a particular abstract state and terminates upon reaching a state mapped to a different abstract state.
\citet{ahmetoglu2026skill} start from a given set of skills and construct an abstract state space that records which skills are available and predicts their rewards and transition outcomes. Each abstract state represents a probability distribution over states in the original MDP, and the resulting abstract MDP is Markov and approximately model-preserving.

\subsection{Continual Reinforcement Learning}
\label{sec:continual-rl}

Continual reinforcement learning (CRL) is often introduced through environmental \emph{non-stationarity}: components such as the transition function, reward function, state space, or action space may change over time \citep{Khetarpal2020TowardsCR}. While this formulation captures many important settings, it places the emphasis on changes in the environment rather than on the learning process of the agent. An agent-centric view instead characterizes CRL as continual adaptation through interaction \citep{ring95continual,abel2023definition,abel2024three}. Whereas conventional RL typically frames learning as the search for a solution to a specified problem, a continual learner may never reach a point at which learning should cease. Streams of tasks with unannounced boundaries are one instance of this broader setting, but are not its defining case.

The need for continual adaptation follows partly from the limited capacity of the agent. Rather than conditioning its decisions on the complete interaction history, a practical agent maintains an \emph{agent state}: an incrementally updated internal summary that may include learned representations, value estimates, models, stored experiences, and reusable skills \citep{dong2022simple,kumar2025continual}. Because memory, computation, and representational capacity are bounded, the agent cannot preserve and process every potentially relevant aspect of its history.
This perspective is sharpened by the \emph{Big World Hypothesis}, according to which the world contains more useful information and behavioural opportunities than a bounded agent can represent \citep{javed2024big}. Consequently, non-stationarity can be relative to the agent: even when the interaction process is stationary when conditioned on the complete history, it may appear non-stationary relative to the agent's compressed internal state. The \textit{stability-plasticity dilemma} should therefore not be reduced to a choice between remembering old experiences and prioritizing recent ones \citep{Carpenter1988AMP}. Instead, stability concerns retaining information that remains useful for future behaviour, whereas plasticity concerns incorporating newly available information that is likely to be useful. Some forgetting is unavoidable, and may be beneficial; it becomes catastrophic when the discarded information continues to support effective behaviour.

HRL offers a principled agent-side mechanism for organizing this limited capacity. Temporally extended actions, such as options or skills, can represent recurring courses of action as reusable decision units and allow an agent to learn and act at multiple timescales \citep{Sutton1999BetweenMA}. Such temporal abstractions may support continual adaptation by preserving useful behavioural structure, accelerating the acquisition of related behaviours, and enabling existing skills to be recombined or specialized for new situations \citep{machado2023temporal,Klissarov2023DeepLO}. In restricted continual-learning settings, theoretical results also indicate that the absence of temporal abstraction can incur additional regret \citep{letourneau2023time}. 
To sum up, one of the fundamental reasons for the synergy between HRL and continual RL is that both fields rarely focus on optimally solving tasks. Instead, they are concerned about fast adaptation and transferability.

While promising, integrating HRL and continual RL presents open research challenges.
As mentioned in Section \ref{sec:6}, it is necessary to develop scalable skill discovery methods that can function in non-stationary settings, devise frameworks that jointly optimize for continual learning and HRL objectives, and design benchmarks and metrics for evaluating agents.

\subsection{Programmatic Reinforcement Learning}
\label{sec:prog}
As stated in Section \ref{sec:2}, HRL could be viewed through an analogy with programming languages and formal systems.
An example of this connection is \textit{Hoare Logic}~\citep{10.1145/363235.363259}, a formal system for assessing the correctness of imperative programs, which shares a similar structure with the option framework (see Section \ref{hrl_formalism}) including initiation sets (pre-conditions), policies (commands), and termination conditions (post-conditions). Both frameworks facilitate reasoning about action sequences, thereby enhancing the structuring of complex decision-making processes.
Research efforts have been undertaken to bridge the gap between HRL, programming languages, logic, and formal methods.

\paragraph{Programs as high-level policy.} 
Early approaches, HAM~\citep{parr1997reinforcement} and PHAM~\citep{andre2000programmable} used hierarchies of partially specified finite-state machines (FSM) to structure policies.
There are four types of states in HAMs: \textit{Action} states execute actions, \textit{Call} states execute subroutines, \textit{Choice} states select subsequent states non-deterministically, and \textit{Stop} states halt execution and return control to prior call states.
This provides a prototype for early HRL methods, allowing for better compositionality, transferability~\citep{andre2000programmable}, and state abstraction~\citep{andre2002state}.
More recent approaches use programs, specifically in domain-specific languages (DSLs), as high-level policies to guide lower-level RL agents. These are often called programmatic policies. Such an approach allows the system designer to inject biases that could, for example, improve sample efficiency over neural representations~\citep{moraes2025innatecoder}. 

Programs convey structured, interpretable, and unambiguous information, and their incorporation into the policy space can reduce the search space for the overall solution and offer a natural method for integrating prior knowledge symbolically. The structured representation of these programs allows one to decompose policies into options that can also be used to induce spaces that are more conducive to search~\citep{moraes2024searching,moraes2025innatecoder}.
In general, the programs can be either hand-crafted~\citep{andreas2017modular,sun2020program}, or automatically synthesized over a predefined syntactic~\citep{carvalho2024reclaiming} or semantic (latent) space~\citep{yang2021program,hasanbeig2021deepsynth,moraes2023choosing,moraes2024searching}. Such synthesis can be performed by parameterizing the program space, also known as \textit{neuro-symbolic}~\citep{sheth2023neurosymbolic} approaches~\citep{denil2017programmable,sohn2018hierarchical,trivedi2021learning,zhao2021proto,qiu2022programmatic,liu2023hierarchical,linhierarchical24}, or by leveraging foundation models~\citep{Wang2023VoyagerAO,Klissarov2024MMotifIM,moraes2025innatecoder}. Learning search guidance for these spaces is also an active area of research~\citep{medeiros2022what,aleixo2023show}. The idea of decomposing policies into subprograms has also been explored even when the underlying policy is a neural network~\citep{alikhasi2024unveiling}.

\paragraph{Programs to option rewards.}
Recent studies~\citep{jothimurugan2019composable,icarte2022reward,furelos2023hierarchies,venuto2024code} demonstrate the feasibility of translating the formal languages (e.g., programs or FSMs) into pseudo-reward signals that facilitate RL training, conceptually connecting to the intrinsic rewards discussed in Section~\ref{sec:other_im_od}..

\paragraph{Distilling the neural policies to interpretable programs.} 
A series of studies focuses on condensing an agent's policy into more hierarchical, interpretable, and verifiable formats such as programs~\citep{verma2018programmatically,verma2019imitation} or Decision Trees~\citep{bastani2018verifiable}, enhancing both lightness and clarity.


\subsection{Cooperative Multi-Agent Reinforcement Learning}
Cooperative multi-agent RL (Cooperative MARL) and HRL can be seen as conceptually connected: managing problem complexity using the structure of the problem. By breaking down large-scale tasks into more manageable subtasks, both approaches improve tractability and facilitate learning.
In cooperative MARL, decomposition is achieved by distributing the decision-making process among multiple agents, whereas in HRL, it is accomplished through temporal abstraction.
As an example, Feudal RL \citep{dayan1993feudal} can be viewed as a multi-agent system comprising managers and workers. This framework naturally extends to cooperative MARL settings \citep{ahilan2019feudal}. Extensive research has explored the integration of HRL with cooperative MARL; 
\citet{hrl_2021} provide further details in Section 3.5 of their work.

\section{Promising Domains for Hierarchical Reinforcement Learning}
\label{sec:8}

In this manuscript, we have examined a wide diversity of HRL approaches, each time highlighting the important ways in which they help decision-making through the benefits we laid out in Section \ref{sec:benefits}.
The vast body of research in HRL encompasses a wide spectrum of methods spanning multiple environments and domains.
Under this diversity of approaches, a key question emerges: \textbf{in what domains should we expect HRL to be most effective?}
An obvious criterion is that the domain comprises temporally extended tasks, such as those found in lifelong learning. As the task horizon decreases, the agent's opportunities to exploit temporal abstractions become increasingly constrained. For example, decomposing short-horizon tasks into subtasks is likely to be less fruitful than long-horizon ones.
However, can we go beyond this obvious criterion to predict the suitability of HRL methods?

We argue that task complexity alone is not a sufficient justification for employing HRL; rather, \textbf{HRL methods should be employed to exploit specific forms of structure}, namely modularity and compositionality. While Section \ref{sec:tradeoffs} illustrates that HRL can efficiently yield good solutions under sample and computation budgets, making it attractive for complex environments where optimality is impractical, it should not be viewed merely as a fallback option when non-hierarchical RL fails. Positioning HRL simply as a last resort for "hard" tasks ignores the underlying mechanics of why these methods work. In reality, a complex environment lacking identifiable and repeated structure may see no benefit from HRL. Similarly, if the goal is only to solve a narrow task, the inherent overhead of learning a hierarchy would likely limit HRL's advantages


HRL appears best suited for long-horizon environments that allow for a diversity of goals that share a structural overlap (whether these goals are defined by the environment or the agent itself).
From this perspective, \textbf{open-ended systems are particularly promising domains for HRL methods}. \citet{Hughes2024OpenEndednessIE} define an open-ended system as one that presents a constant flow of novel and possibly learnable goals. It is common in such systems that these goals share, to a degree, a common underlying structure, which makes HRL particularly appealing. Below, we showcase specific domains exemplifying these characteristics. 
This list is not exhaustive but rather serves to illustrate settings where HRL might excel.


\paragraph{Web Agents.} The World Wide Web, a dynamic and ever-changing environment, presents a unique challenge for AI agents. The recent surge in interest has led to a variety of implementations of challenging domains, such as Android-in-the-Wild \citep{rawles2023androidinthewild} or WebArena \citep{zhou2023webarena}.  Its near-infinite tasks and constantly evolving goals demand adaptability and the ability to decompose complex objectives into manageable subgoals. As mentioned in Section \ref{sec:3}, even if the resulting agent is not hierarchical (i.e., does not explicitly carry a set of skills),  learning to navigate the web through HRL methods, such as curriculum-based ones, can help address the sheer complexity of the web. Indeed, Web Agents must learn to navigate a constantly shifting landscape of information and services, adapting to new data, evolving user preferences, and the emergence of novel websites and services. Another important characteristic is that many tasks of interest share a lot of underlying structure, a key point of HRL. Overall, this complex and open-ended domain requires agents capable of learning, adapting, and generalizing across multiple timescales, ultimately revolutionizing how we interact with the online world.

\paragraph{Robotics.} Robotics, with its emphasis on embodied intelligence and real-world interaction, presents a compelling domain for exploring the potential of HRL methods. The tasks robots face, from navigating complex environments to manipulating objects with dexterity, involve long horizons where, at each step, a low-level action is sampled from a continuous action space. HRL offers a natural framework for decomposing these complex tasks into manageable sub-policies, allowing robots to learn and refine abstract skills while also developing higher-level strategies for sequencing and coordinating them. Practical implementations of interest include AI2-THOR \citep{Kolve2017AI2THORAI}, Habitat \citep{Szot2021Habitat2T,puig2024habitat}, CALVIN \citep{mees2022calvin} and OGBench \citep{park2024ogbench}.

The ability to recompose learned skills into novel combinations is crucial for robots operating in unstructured and dynamic environments, where adaptability and generalization are key. For instance, a robot learning to grasp objects might develop sub-policies for reaching, orienting its gripper, and applying the appropriate force. Ideally, these individual skills could then be recombined and adapted to grasp a wide variety of objects in different contexts, without requiring retraining from scratch. The long horizons inherent in many robotic tasks, coupled with the need for flexible and adaptable skill acquisition, make HRL a promising approach for developing robots capable of performing complex, real-world tasks with increasing autonomy and efficiency. \citet{konidaris2011autonomous} used the methods discussed in Section~\ref{sec:skillchaining} to provide the first demonstration of option discovery on a real robot; scaling such methods remains an important open problem.

\paragraph{Open-ended games.} Training AI agents on games has a long history of striking successes in domains like Go \citep{Silver2017MasteringTG} or Atari 2600 games \citep{Mnih2015HumanlevelCT}\footnote{Note that this applies less to the case of hard exploration games such Montezuma's Revenge, Private Eye, or Pitfall!, among others.}. However, for HRL to provide significant value over non-hierarchical RL, the domain would likely be be complex, long-horizon, and open-ended. We have seen in Section \ref{sec:curriculum} such an example, where a goal-conditioned policy trained on a large diversity of tasks led to human-timescale adaptation. Key to this success was the fact that data was readily available through fast simulation, allowing for quicker research iteration. This makes it particularly interesting to study open-ended games in order to better understand HRL methods. We now provide two such examples. \textbf{NetHack} is a complex roguelike game, and it is a good environment for exploring the benefits of HRL. It has been brought to the RL community through the NetHack Learning Environment \citep{kuettler2020nethack}. Its open-ended nature, procedurally generated dungeons, and long-horizon gameplay require exploration, planning, and adaptation across multiple timescales. Success requires both immediate tactical decisions and  strategizing towards long-term goals, demanding credit assignment across extended temporal spans. The vast diversity of situations encountered also requires generalization, making HRL's ability to learn reusable sub-policies and higher-level strategies particularly valuable. 
\textbf{Minecraft} \citep{Johnson2016TheMP,Kanervisto2022MineRLD2}, with its expansive, procedurally generated world and open-ended gameplay, presents a compelling testbed for HRL algorithms. The game requires navigating across diverse biomes, gathering resources, crafting tools, and structures, and ultimately, surviving and thriving. This requires planning and execution across multiple timescales. For instance, while the immediate goal might be chopping down a tree for wood, this action serves the higher-level objective of building a shelter for protection against nocturnal mobs. Furthermore, Minecraft's crafting system inherently embodies a hierarchical structure. Creating complex items like diamond tools requires a chain of prerequisite crafting steps, each with its own subgoals and resource requirements. HRL agents could learn to decompose these complex tasks into manageable sub-policies, mirroring the hierarchical nature of crafting itself. 


\section{Conclusion}

The ability to write, store, and reuse subroutines is one of the great ideas in computer science \citep{GreatIdeas}. A subroutine, once written, can be called without regard to its internal operation; that is modularity. A subroutine can also be combined with others to generate new programs its author never anticipated; that is the power of compositionality. These two powerful ideas are the organizing principles that have enabled human engineers to write and maintain the extraordinarily complex programs that have powered the computing revolution. A generally intelligent AI agent will likely require similar organization, and HRL provides the foundations on which such agents can be built: procedural abstractions that improve exploration, credit assignment, and knowledge transfer. In this text, we looked at three broad families of approaches to acquiring such abstractions---discovery via online experience, via offline data, and via foundation models---organizing each around benefits and the central price it exacts.

No inductive bias is universally beneficial, and HRL's bias toward decomposable structure can hurt as readily as help when the structure is misimagined. Hierarchies enter a Pareto frontier between performance, sample efficiency, and computational cost, often trading a little of the first and third for a great deal of the second.

The core open question before the field is that of discovery: how can we build agents that acquire these abstractions themselves via interaction? 
This problem of option discovery, in its grandest form, is one of the biggest meta-scientific questions of our time. An agent that can ask useful questions (option objectives) and efficiently find answers to them (option learning), will eventually acquire abstractions that unlock sample efficient reward maximization in many tasks.

This is an opportune moment to study this problem of option discovery. Advances in deep learning have made the problems on which hierarchy plausibly matters---high-dimensional and long-horizon---tractable to study at scale. Modern foundation models and large offline datasets offer means of incorporating useful priors into an agent's learning, letting discovery begin from somewhere informed rather than from nothing. Together, these developments make the HRL problem feel within reach and well-poised to scale RL to larger, more complicated, and high-impact application areas.

\newpage
\section*{Notation}
\begin{table}[ht]
\centering
\caption{\small Glossary of notations used in RL and HRL (see Section~\ref{sec:3}).}
\renewcommand{\arraystretch}{1.3}
\begin{tabularx}{\linewidth}{>{\hsize=0.25\hsize}X >{\hsize=0.75\hsize}X}
\toprule
\multicolumn{2}{l}{\textbf{Reinforcement Learning (RL)}} \\
\midrule
$S_t \in \mathscr{S}$ & State at time step $t$ \\
$A_t \in \mathscr{A}$ & Action at time step $t$ \\
$R_{t+1}$ & Real-valued random variable representing the reward at time $t+1$ \\
$p(s' \mid s, a)$ & Transition probability \\
$\rho_0$ & Initial state distribution \\
$\gamma \in [0,1)$ & Discount factor \\
$\pi(a \mid s)$ & Policy over actions \\
$d_\pi^\gamma(s)$ & Discounted state occupancy under policy $\pi$ \\
$q_\pi(s, a)$ & Action-value function under policy $\pi$ \\
$v_\pi(s)$ & State-value function under policy $\pi$ \\
$V_\phi(s)$ & Estimated state-value function, parameterized by $\phi$ \\
$Q_\phi(s, a)$ & Estimated action-value function, parameterized by $\phi$ \\
\midrule
\multicolumn{2}{l}{\textbf{Hierarchical Reinforcement Learning (HRL)}} \\
\midrule
$o \in \mathscr{O}$ & Option \\
$\pi_\theta(a \mid s, o)$ & Option policy, parameterized by $\theta$ \\
$\beta_\psi(s, o)$ & Option termination function, parameterized by $\psi$ \\
$\mathscr{I}_\chi(s, o)$ & Option initiation function, parameterized by $\chi$ \\
$\mu_\kappa(o \mid s)$ & High-level policy, parameterized by $\kappa$ \\
$P_{\mathscr{O}}(s, o, s')$ & Option transition model \\
$R_{\mathscr{O}}(s, o)$ & Option reward model \\
$q_\pi(s, o)$ & Option-value function  under policy $\pi$  \\
$q_u(s, o, a)$ & Action-value within the context of option $o$  \\
$u_\beta(o, s')$ & Option value upon arrival \\
$v_\mu(s)$ & Value function under high-level policy $\mu$ \\
$\omega_o \in \Omega$ & Option objective associated with option $o$ \\
$r^o_\nu(s, a)$ & Option reward function for option $o$, parameterized by $\nu$ \\
$z \in \mathscr{Z}$ & Option identifier \\
$g \in \mathscr{G}$ & Goal or subgoal \\
\bottomrule
\end{tabularx}
\label{tab:notation_rl_hrl}
\end{table}

\newpage

\section*{Acknowledgements} We would like to thank Xujie Si, Khimya Khetarpal, Seohong Park, Bartłomiej Cupiał, Isabeau Prémont-Schwarz, Levi Lelis, Andrew Levy, Alex Ivanov, Nishanth Anand, and Jonathan Colaço Carr for their valuable feedback. This research is supported in part by NSF 1955361 and NSF CAREER 1844960.
The research is supported in part by the Natural Sciences and
Engineering Research Council of Canada (NSERC), and the Canada CIFAR AI Chair Program. 

\bibliography{refs_new}
\bibliographystyle{apalike}

\end{document}